\documentclass{article}

\usepackage{microtype}
\usepackage{graphicx}
\usepackage{subfigure}
\usepackage{booktabs} 

\usepackage{hyperref}

\newcommand{\theHalgorithm}{\arabic{algorithm}}

\usepackage[accepted]{icml2024}

\usepackage{amsmath}
\usepackage{amssymb}
\usepackage{mathtools}
\usepackage{amsthm}
\usepackage{bm}
\usepackage[algo2e]{algorithm2e}

\usepackage[capitalize,noabbrev]{cleveref}

\theoremstyle{plain}

\theoremstyle{definition}

\theoremstyle{remark}

\usepackage[textsize=tiny]{todonotes}

\icmltitlerunning{Optimal Recurrent Network Topologies for Dynamical Systems Reconstruction}

\begin{document}

\twocolumn[
\icmltitle{Optimal Recurrent Network Topologies for Dynamical Systems Reconstruction}


\icmlsetsymbol{equal}{*}

\begin{icmlauthorlist}
\icmlauthor{Christoph Jürgen Hemmer}{zi,hd}
\icmlauthor{Manuel Brenner}{zi,hd}
\icmlauthor{Florian Hess}{zi,hd}
\icmlauthor{Daniel Durstewitz}{zi,hd,iwr}
\end{icmlauthorlist}

\icmlaffiliation{zi}{Department of Theoretical Neuroscience, Central Institute of Mental Health, Medical Faculty Mannheim, Heidelberg University, Mannheim, Germany}
\icmlaffiliation{hd}{Faculty of Physics and Astronomy, Heidelberg University, Heidelberg, Germany}
\icmlaffiliation{iwr}{Interdisciplinary Center for Scientific Computing, Heidelberg University, Heidelberg, Germany}

\icmlcorrespondingauthor{Christoph Hemmer}{Christoph.Hemmer@zi-mannheim.de}
\icmlcorrespondingauthor{Daniel Durstewitz}{Daniel.Durstewitz@zi-mannheim.de}

\icmlkeywords{Machine Learning, ICML}

\vskip 0.3in
]



\printAffiliationsAndNotice{}  

\begin{abstract}
In dynamical systems reconstruction (DSR) we seek to infer from time series measurements a generative model of the underlying dynamical process. This is a prime objective in any scientific discipline, where we are particularly interested in parsimonious models with a low parameter load. A common strategy here is parameter pruning, removing all parameters with small weights. However, here we find this strategy does not work for DSR, where even low magnitude parameters can contribute considerably to the system dynamics. On the other hand, it is well known that many natural systems which generate complex dynamics, like the brain or ecological networks, have a sparse topology with comparatively few links. Inspired by this, we show that \textit{geometric pruning}, where in contrast to magnitude-based pruning weights with a low contribution to an attractor's geometrical structure are removed, indeed manages to reduce parameter load substantially without significantly hampering DSR quality. We further find that the networks resulting from geometric pruning have a specific type of topology, and that this topology, and not the magnitude of weights, is what is most crucial to performance. We provide an algorithm that automatically generates such topologies which can be used as priors for generative modeling of dynamical systems by RNNs, and compare it to other well studied topologies like small-world or scale-free networks. 

\end{abstract}

\section{Introduction}
\label{introduction}
In scientific settings we are commonly interested in the dynamical rules that govern the temporal evolution of an observed system. Recent data-driven deep learning approaches aim to infer (approximate) these from time series measurements of the system under study, often based on recurrent neural networks (RNNs) like LSTMs \cite{vlachas2018data,lstm}, reservoir computers \cite{pathak2018model,platt2021robust}, piecewise linear RNNs \cite{10.1371/journal.pcbi.1005542,brenner2022tractable,hess2023generalized}, or neural ordinary differential equations (Neural ODEs; \citet{chen2018neural, ko_homotopy-based_2023}). Producing a generative model of the dynamical process that underlies the observed time series, including its geometrical structure in state space and its invariant (long-term) statistics, we call dynamical systems (DS) reconstruction (DSR). 

In science we usually prefer small models, with as few relevant parameters as possible, to reduce training and simulation times, and to ease subsequent model analysis (interpretability). Magnitude-based parameter pruning is a well established strategy in deep learning to carve out such a low-dimensional (in parameter space) network structure \cite{blalock2020state}. It is based on the insight that successfully trained models often contain a `winning ticket', a small subnetwork with performance almost equal to that of the full large network \cite{frankle2019lottery}. Here, however, we demonstrate that magnitude-based parameter tuning does not work well for DSR. Instead, we find that parameters even small in relative size could substantially influence the dynamics of the trained model (Fig. \ref{fig:geometry-based_pruning}). This might be related to the fact that in dynamical systems neighborhood relations (topological structure) are often more important. Natural systems, like the brain \cite{bullmore2009complex,pajevic2012organization}, climate systems \cite{tziperman-97}, or ecological and social networks \cite{watts_collective_1998}, rarely follow an ``all-to-all'' connectivity, but have a well defined topology owing to physical and spatial constraints on the system. This suggests that deep learning models trained on time series from natural systems may inherit similar organizing principles, i.e. a low-dimensional parameter representation is conceivable, but it may be less related to the relative size of parameters.\newline

Here we demonstrate that this is indeed the case. In contrast to pruning by size, selecting parameters based on their relevance for the invariant geometrical structure of the dynamical system enabled to profoundly sparsify recurrent neural networks (RNNs) without considerably affecting performance. Such \textit{geometry-pruned} RNNs turned out to bear a specific topology, which \textit{in itself} was largely sufficient for performance-preserving sparsification. We further find that while more `traditional' models of network topology, like the Watts-Strogatz \cite{watts_collective_1998} or Albert-Barabási \cite{Barabasi99emergenceScaling} model, can also produce efficiency gains beyond magnitude-based or random pruning, these are not quite as profound as those obtained through our procedure.

\section{Related Work}\label{related work}

\paragraph{Dynamical Systems Reconstruction (DSR)}
In DSR we aim to obtain from time series data a generative model that is at least topologically conjugate to the flow of the underlying true system on the domain observed \citep{durstewitz2023reconstructing}, with the same invariant long-term properties, including attractor geometry and temporal characteristics (as assessed, e.g., through power spectra or auto-correlation functions; \citet{wood_statistical_2010,platt2021robust, brenner2022tractable,mikhaeil2022difficulty,platt2023constraining}). The forecasting ability of a model on its own is usually not considered a viable performance criterion for DSR models, because in chaotic systems nearby trajectories diverge exponentially fast and hence there is only a limited prediction horizon \cite{wood_statistical_2010, koppe_fmri_2019}. Various architectures and training algorithms have been proposed for DSR, based on RNNs like LSTMs \cite{vlachas2018data,lstm}, piecewise-linear RNNs (PLRNNs; \citet{koppe_fmri_2019, brenner2022tractable,hess2023generalized}), or reservoir computers \cite{pathak2018model,platt2021robust}, based on library methods \cite{brunton2016discovering, champion_data-driven_2019, messenger_weak_2021}, or based on neural ordinary differential equations (Neural ODEs; \citet{chen2018neural, karlsson_modelling_2019, alvarez_dynode_2020, ko_homotopy-based_2023}). A variety of different training strategies has been suggested for DSR models, e.g. based on Expectation-Maximization \cite{voss_nonlinear_2004, koppe_fmri_2019} or on variational inference \cite{kramer2021reconstructing}. However, the to date most effective training algorithms rely on Backpropagation Through Time (BPTT) \cite{rumelhart1986learning} combined with control-theoretic forms of teacher forcing (TF; \citet{williams1989learning}), like sparse \cite{mikhaeil2022difficulty, brenner2022tractable} or generalized \cite{hess2023generalized} TF, that guarantee optimal trajectory and gradient flows whilst training, and avoid exploding gradients even on chaotic systems. The role of RNN topology in DSR has, however, not been studied yet, providing one novel contribution of the present work.

\paragraph{Pruning and Lottery Ticket Hypothesis}
Linked to the double descent phenomenon \cite{Belkin_2019}, modern deep learning models are often extensively over-parameterized, surpassing the interpolation threshold, which leads to improved network trainability and expressiveness. While smaller models with fewer parameters are generally preferable for computational and memory reasons, and to enhance interpretability, this becomes almost imperative for these strongly over-parameterized systems. \citet{NIPS1989_6c9882bb} is one of the first studies that explored pruning networks in order to remove redundancies in parameters. Since then it has been shown numerous times that in deep learning architectures a non-trivial number of parameters remains essentially unexploited \cite{NIPS2015_ae0eb3ee,han2015deep}, and a variety of different pruning techniques have been developed \cite{blalock2020state}, for instance, Hessian based approaches \cite{NIPS1992_303ed4c6} or structured pruning \cite{He_2023}. The most common and straightforward procedure, however, remains pruning parameters based on their absolute value \cite{blalock2020state}. This can be done in different ways, for instance, pruning in a single step \cite{liu2018rethinking} or iteratively \cite{han2015learning, zhang_why_2021}. Hope that this may work more generally is based on the so-called lottery ticket hypothesis (LTH), which states \cite{frankle2019lottery}: \textit{``A randomly-initialized, dense neural network contains a subnetwork that is initialized such that -- when trained in isolation -- it can match the test accuracy of the original network after training for at most the same number of iterations.''} Empirically this was verified for feed-forward neural networks in image classification using iterative magnitude pruning \cite{frankle2019lottery, girish_lottery_2021}, and has spawned a range of theoretical and empirical follow-up studies \cite{malach_proving_2020, orseau_logarithmic_2020, zhang_why_2021, sreenivasan_rare_2022, burkholz_existence_2022}. \citet{burkholz_existence_2022} even suggests the existence of universal LTs that are winning tickets across tasks. However, so far there is comparatively little work on the LTH and pruning of RNNs \cite{yu2019playing,liu2021selfish,CHATZIKONSTANTINOU2021475}. 

\paragraph{Network Topology of Real World Systems}
Many complex real-world systems have a characteristic network topology \cite{watts_collective_1998, albert2002statistical}, for instance, a small-world topology shared by many biological and social systems \cite{kleinberg2000small,amaral2000classes, bassett2006small, rubinov2009small,muldoon2016small}, or a scale-free topology observed in brain anatomy and dynamics \cite{beggs_neuronal_2003,eguiluz2005scale,van2008small, rubinov2009small}. Scale-free networks are characterized by the existence of central nodes or `hubs', possibly linked to a hierarchical organization in the system \cite{ravasz_hierarchical_2003}. While sometimes the specific network topology may simply be a result of the underlying physics (e.g., spatially restricted interaction patterns or physical barriers; \citet{jiang2004topological}), in other instances they may bear specific functional or biological advantages like minimizing wiring costs while optimizing information transfer \cite{bullmore2012economy}, or increasing efficiency \cite{zhang_trapping_2009}. 

\paragraph{Network Topology in RNNs}
Given the importance of topological structure in real-world networks, inferring topology directly from observed data is of particular interest \cite{shandilya_inferring_2011, wang_topology_2016}, although not our focus here. Regarding the reverse direction, \citet{emmert-streib_influence_2006} were among the first to point out that RNN topology also influences learning dynamics. This has been particularly well studied in the context of reservoir computers (RCs; \citet{jaeger_harnessing_2004}). \citet{dutoit_pruning_2009} observed that pruning connections in the reservoir's output layer leads to improved generalization. \citet{yin_self-organizing_2012} adapt the structure of the dynamical reservoir to mirror that of cortical networks, while \citet{carroll_network_2019} study the influence of directedness of edges in RCs more generally. Other authors \cite{li_echo_2020, junior_clustered_2020, dale_reservoir_2021}  directly studied the impact of specific topologies, like hub structure, directed acyclic graphs, or Erdős–Rényi graphs, on RC performance. \citet{han_tighter_2022} study generalization bounds for RCs initialized with different graph structures. In contrast to our work here, however, all these studies impose a specific, previously defined topology to begin with, and do not examine structure or properties that arise through training or pruning (in fact, recall that in RCs the recurrent connectivity is \textit{fixed} and not altered throughout training). This question thus remains largely unexplored, especially in the context of DSR. Small-world topology has, however, been observed in task-optimized \textit{feedforward} networks like MLPs or CNNs \cite{you2020graph}, associated with superior performance. 

\section{Methodological Setting}
\label{methods}

\subsection{DSR Model and Training} \label{DSR_model_training}
For our numerical studies we focus on a well established SOTA model and training algorithm for DSR, the piecewise linear recurrent neural network (PLRNN) first introduced in \cite{10.1371/journal.pcbi.1005542}, but also checked LSTMs and vanilla RNNs to highlight that our results are more general. The PLRNN is defined by 
\begin{equation} \label{eq:PLRNN}
    \bm{z}_t=\bm{A}\bm{z}_{t-1}+\bm{W}\phi(\mathcal{M}(\bm{z}_{t-1}))+\bm{h}+\bm{C}\bm{s}_t\;,
\end{equation}
where $\mathcal{M}$ is a mean-centering operation (see Appx. \ref{sect:mean_centred_plrnn} and \citet{brenner2022tractable}), and with element-wise nonlinearity $\phi(\bullet)=\text{ReLU}(\bullet)=\max(0,\bullet)$. The model describes the temporal evolution of an $M$-dimensional latent state vector, $\bm{z}_t\in\mathbb{R}^M$, with linear self-connections in diagonal matrix $\bm{A}\in\mathbb{R}^{M\times M}$, full weight matrix $\bm{W}\in\mathbb{R}^{M\times M}$, bias term $\bm{h}\in\mathbb{R}^M$, and possibly external inputs $\bm{s}_t\in\mathbb{R}^K$ weighted by $\bm{C}\in\mathbb{R}^{M\times K}$ (the benchmarks we explore in here, however, are all autonomous DS with $\bm{s}_t=0 \ \forall t$).\footnote{Also, mathematically, a non-autonomous system can always be equivalently rewritten as an autonomous system \cite{zhang2009controlling}.} 
The latent PLRNN is linked to the actually observed time series $\bm{X}=\{\bm{x}_1,...,\bm{x}_T\}$ through a decoder (observation) model, in the simplest case given by a linear layer:
\begin{equation}
    \bm{x}_t=G_{\bm{\lambda}}(\bm{z}_t)=\bm{B}\bm{z}_t\;,
\end{equation}
where $\bm{B}\in\mathbb{R}^{N\times M}$. Numerous variations on this basic model have been introduced and benchmarked on DSR problems in the literature \cite{koppe_fmri_2019, brenner2022tractable,hess2023generalized}, but here we will stick to this most basic form to enhance interpretability of our results in graph-theoretical language (but see Appx. \ref{ch:Methodological_details} for further details).\footnote{In fact, as shown in \citet{brenner2022tractable} and \citet{hess2023generalized}, more complex PLRNN variants can be reformulated in terms of Eqn. \ref{eq:PLRNN}.} Of major importance in scientific settings, a crucial advantage of model Eqn. \ref{eq:PLRNN} is that it allows for an equivalent continuous-time formulation (i.e., as a system of ODEs; \citet{pmlr-v119-monfared20a}), and all its fixed points and cycles can be exactly determined by efficient, often linear-time, algorithms \cite{eisenmann2023bifurcations}.

In DSR, we would like to capture long-term statistical and geometrical properties of the underlying DS, beyond mere short-term forecasts \cite{platt2021robust, durstewitz2023reconstructing, platt2023constraining}. It turns out that the actual training algorithm is much more important for this than the RNN architecture \cite{mikhaeil2022difficulty,hess2023generalized}. In particular, efficient training routines often implement control-theoretic ideas like sparse \cite{mikhaeil2022difficulty} or generalized \cite{hess2023generalized} teacher forcing (STF, GTF) that manage the exploding-\&-vanishing gradient problem even for chaotic systems with diverging trajectories. To keep things simple, here we employ STF \cite{mikhaeil2022difficulty,brenner2022tractable} with an identity mapping for the observation model, i.e. 
\begin{equation} \label{eq:identity_model}
    \bm{x}_t=\bm{\mathcal{I}}\bm{z}_t, \;\;\;\;\;\mathcal{I}_{kl}=
    \begin{cases}
        1, & \text{if $k=l$ and $k\leq N$} \\
        0, & \text{else}
    \end{cases}\;.
\end{equation}
STF then replaces the latent states $z_{t',k},k\leq N$, with observations $x_{t',k}$ sparsely at strategically chosen time points $t'\in\mathcal{T}=\{n\tau+1\}$ with $n\in\mathbb{N}$ \cite{mikhaeil2022difficulty}, thereby re-calibrating trajectories during training such that relevant time scales are captured yet too wild divergence and exploding gradients are prevented (for details see Appx. \ref{ch:Methodological_details}; note that no such forcing is used in the actual test or generation phase).

\subsection{Weight Pruning}
We implement weight pruning \cite{NIPS1989_6c9882bb} by applying a mask $\bm{m}$ to the weight matrix $\bm{W}$ using element-wise multiplication:
\begin{equation} \label{eq:graph_PLRNN}
    \bm{z}_{t}=\bm{A} \bm{z}_{t-1}+(\bm{m}\odot \bm{W})\phi(\mathcal{M}(\bm{z}_{t-1}))+\bm{h}\;,
\end{equation}
where $\bm{m}\in\{0,1\}^{M\times M}$ represents the network topology.
Pseudo-code for the iterative pruning procedure, retaining initial parameters $\bm{\theta}_0$ but updating the mask in each iteration, is given in Algorithm \ref{alg:Pruning}. 
\begin{figure}[ht]
\centering
\begin{minipage}{.9\linewidth}
\begin{algorithm}[H]
\caption{Pruning algorithm}
\SetKwInOut{Input}{Input}\SetKwInOut{Output}{Output}
\Input{Initial model $f(\bm{x};\bm{\theta}_0=\{\bm{A}_0,\bm{W}_0,\bm{h}_0\})$ with initial parameters $\bm{\theta}_0$}
\Output{Mask $\bm{m}$ of pruned network}
\BlankLine

Initialize $m_{ij}^1=1\;\forall i,j\in[1,...,M]$

\For{$k \gets 1$ \KwTo $n$}{
    \begin{enumerate}
        \item Train model $f\left(\bm{x};\bm{\theta}_0=\{\bm{A}_0,\bm{m}^k\odot\bm{W}_0,\bm{h}_0\}\right)$ for $j$ epochs, yielding parameters $\bm{\theta}$
        \item  Remove $p\%$ of parameters $w_{ij}\in\bm{\theta}$ based on their contribution $I_{w_{ij}}$ to model performance, resulting in mask $\bm{m}^{k+1}$
        \item Reset parameters to $\bm{\theta_0}$
    \end{enumerate}
}

\label{alg:Pruning}
\end{algorithm}
\end{minipage}
\end{figure}

\begin{figure}[ht!]
    \begin{center}
    \includegraphics[width=0.47\textwidth]{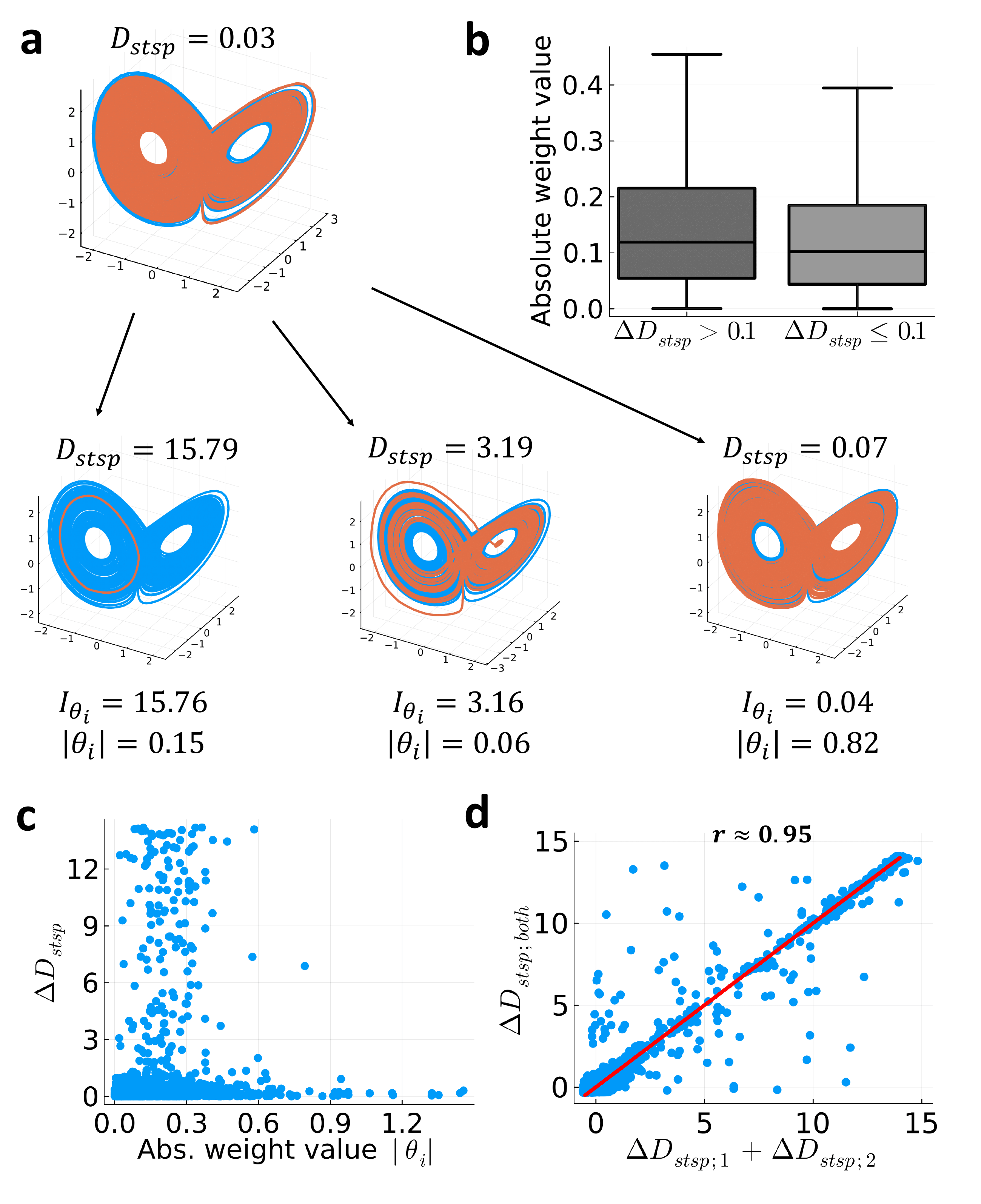}
    \vspace{-.4cm}
    \caption{\textbf{a}) Illustration of geometry-based pruning. Top shows the (ground truth) iconic Lorenz-63 \cite{DeterministicNonperiodicFlow} chaotic attractor (blue) and an optimal PLRNN reconstruction (red), while below three reconstructions are shown with a single weight parameter removed with high (leftmost), medium (center) or low (rightmost) influence on attractor geometry. Measure for geometrical (dis)agreement ($D_{\text{stsp}}$) on top of each graph, and geometric importance score and magnitude of pruned parameter indicated below. \textbf{b}) Weight parameters with large ($\Delta D_{\text{stsp}}>0.1$) vs. low ($\Delta D_{\text{stsp}}\leq0.1$) impact on geometrical reconstruction quality do not substantially differ in absolute magnitude. \textbf{c}) Change in geometrical disagreement ($\Delta D_{\text{stsp}}$) vs. weight magnitude for PLRNNs trained on the Lorenz-63. Note there is no discernible trend for larger weights to associate with stronger effects on attractor geometry. \textbf{d}) 
    The effects of weight removal on $\Delta D_{\text{stsp}}$ are largely additive, with simultaneous removal of two weights having about the same effect as the sum of the individual weight effects.
    }
    \label{fig:geometry-based_pruning}
    \end{center}
\end{figure}

Traditionally in pruning procedures, importance $I_{\theta_i}$ of a parameter is simply measured by its absolute magnitude, i.e. $I_{\theta_i}=|\theta_i|$. As we find below that weight magnitude is only weakly correlated with DSR performance, we introduce \textit{geometric pruning} as a means to examine 1) whether a significant sparsification of the network is possible in this context, and 2) which parameters do have a significant impact. In geometric pruning we iteratively, using the same protocol as in Algorithm \ref{alg:Pruning}, remove those connections that have the least impact on attractor geometry in state space (Fig. \ref{fig:geometry-based_pruning}). Since in DSR we are interested in obtaining a generative model that has the same long-term temporal behavior and geometrical structure in state space as the true underlying DS, this is a direct indicator of DSR quality. Formally we define it through the same measure that has been used to assess geometrical agreement, a Kullback-Leibler (KL) divergence in the system's \textit{state space} \cite{koppe_fmri_2019,hess2023generalized}, namely 
\begin{equation}
\begin{split}
    I_{\theta_i} =& \left|\text{KL}(p_{true}(\bm{x})\lVert p_{gen}^{-i}(\bm{x}|\bm{z})) \right. \\ 
    & \left. -\text{KL}(p_{true}(\bm{x})\lVert p_{gen}(\bm{x}|\bm{z}))\right|\;,
\end{split}
\end{equation}
where $p_{true}(\bm{x})$ is the limit set distribution across state space of the ground truth trajectories, $p_{gen}(\bm{x}|\bm{z})$ the corresponding full model-generated trajectory distribution, and $p_{gen}^{-i}(\bm{x}|\bm{z})$ the model generated distribution with parameter $\theta_i$ removed. In low-dimensional spaces this KL divergence can be approximated by simply discretizing (binning) the space, while in higher-dimensional spaces a Gaussian mixture model approximation is usually employed (see Appx. \ref{sect:evaluation_measures} and \citet{brenner2022tractable} for details). As illustrated in Fig. \ref{fig:geometry-based_pruning}, in geometric pruning $I_{\theta_i}$ naturally picks out those weight parameters which contribute the least to attractor geometry. Computing this measure is generally costly and pruning through this procedure may thus not always be feasible in practical settings. However, here it mainly served to study resulting network topologies, based on which initialization templates can be constructed (see sect. \ref{sect:network_topology_training}).

\subsection{Analysis of Network Topology} \label{sect:analysis_network_topology}
A graph $G=(V,E)$ consists of a set of nodes $V(G)=\{v_i\}$, $i\in1,...,n$, and a set of edges $E(G)=\{e_{ij}\}$ (or links) between nodes \cite{graph_theory_diestel}. Network topology focuses on abstract structural properties of such graphs, as specified through the adjacency matrix $\bm{A}^{\text{adj}}$, which codes the existing links (and their direction) between any two nodes. In our case it is given by the pruning mask $\bm{A}^{\text{adj}}=\bm{m}$. The type of graph and its tendency to form highly connected hubs is determined by its degree distribution $P(k)$, where the degree $k$ refers to the number of connections a node receives \cite{graph_theory_diestel}. The two most important statistical quantities describing the properties of a graph, which we will use here, are, first, the mean average path length $L$ defined as the minimum path length between two nodes averaged over all pairs of nodes \cite{watts_collective_1998}: 
\begin{equation} \label{eq:average_path_length}
    L(G) = \frac{1}{n(n-1)}\sum_{i\neq j}d(v_i,v_j)\;,
\end{equation}
where $d(v_i,v_j)$ is the geodesic distance between node $v_i$ and node $v_j$ in a graph with $n$ nodes. Second, the \textit{clustering} of nodes is calculated as
\cite{PhysRevE.76.026107}
\begin{equation} \label{eq:clustering_coefficient} 
    C(G)=\frac{1}{n}\sum_{i=1}^n\frac{(\bm{A}^{\text{adj}}+(\bm{A}^{\text{adj}})^T)^3_{ii}}{2T_i}\;,
\end{equation}
where $\bm{A}^{\text{adj}}$ is the adjacency matrix of the graph. The clustering $C(G)$ gives the ratio between the number of all triangle connections (i.e., where two neighbors of a chosen node $v_i$ are also directly connected) and the total theoretically possible number of triangles $T_i$. Details on these measures can be found in Appx. \ref{PLRNN topology}. 

\begin{figure*}[!htb]
    \begin{center}
    \includegraphics[width=0.9\textwidth]{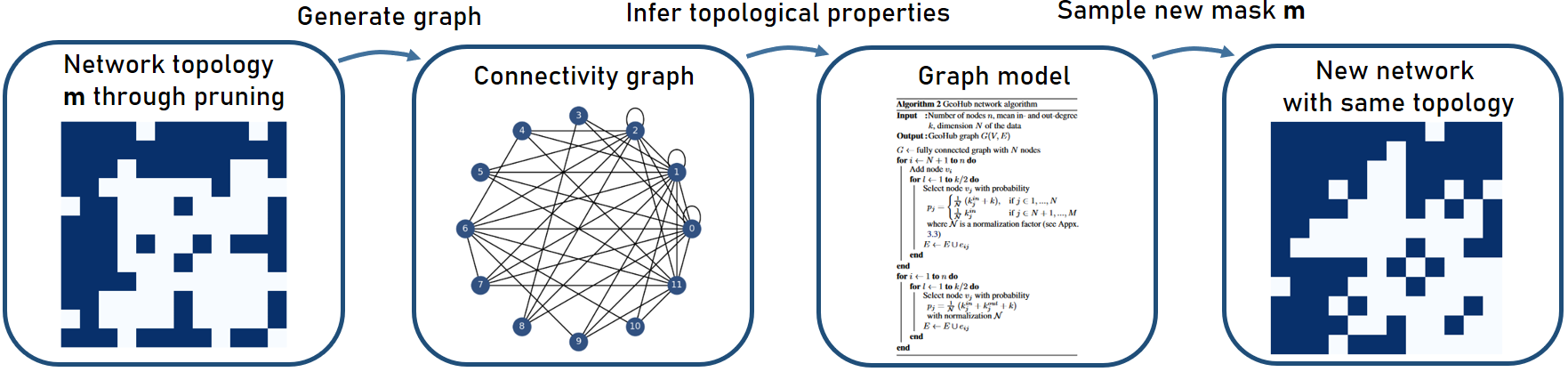}
    \caption{Approach for translating graph-topological properties of trained networks into a general scheme to be used as topological prior.\\
    }
    \label{fig:plrnn_graph_topology}
    \end{center}
\end{figure*}

Based on such a topological characterization of network graphs obtained through geometric pruning of trained PLRNNs, in sect. \ref{sect:network_topology_training} we derive an algorithm that creates an adjacency matrix $\bm{A}^{\text{adj}}$ with the desired properties which can be used as a mask $\bm{m}$ in Eqn. \ref{eq:graph_PLRNN}. Fig. \ref{fig:plrnn_graph_topology} illustrates the general approach. We compare its reconstruction performance with common network structures from graph theory, like the Erdős–Rényi model \cite{erdos59a} generating random graphs, the Watts-Strogatz model \cite{watts_collective_1998} known for its small-world properties, or the Barabási-Albert model \cite{Barabasi99emergenceScaling,Albert_2002} producing central hub topologies.

\begin{figure*}[ht!]
    \centering
    \begin{subfigure}
         \centering
         \includegraphics[width=0.9\textwidth]{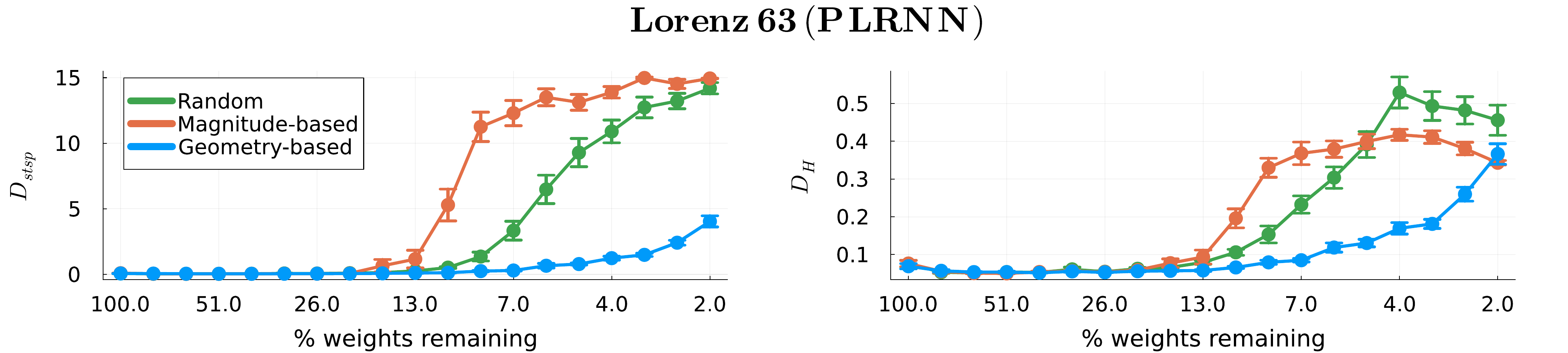}
    \end{subfigure}
    \begin{subfigure}
         \centering
         \includegraphics[width=0.9\textwidth]{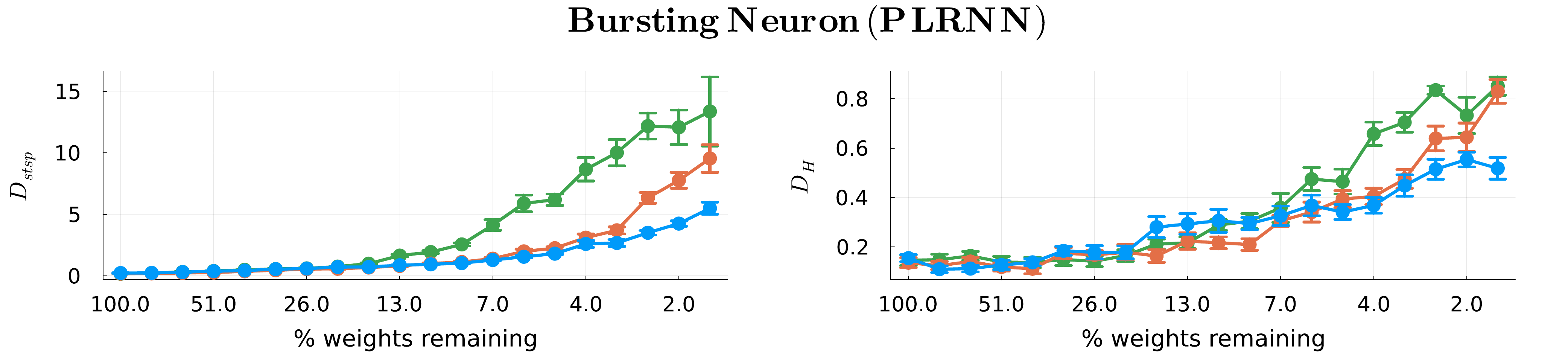}
    \end{subfigure}
    \begin{subfigure}
         \centering
         \includegraphics[width=0.9\textwidth]{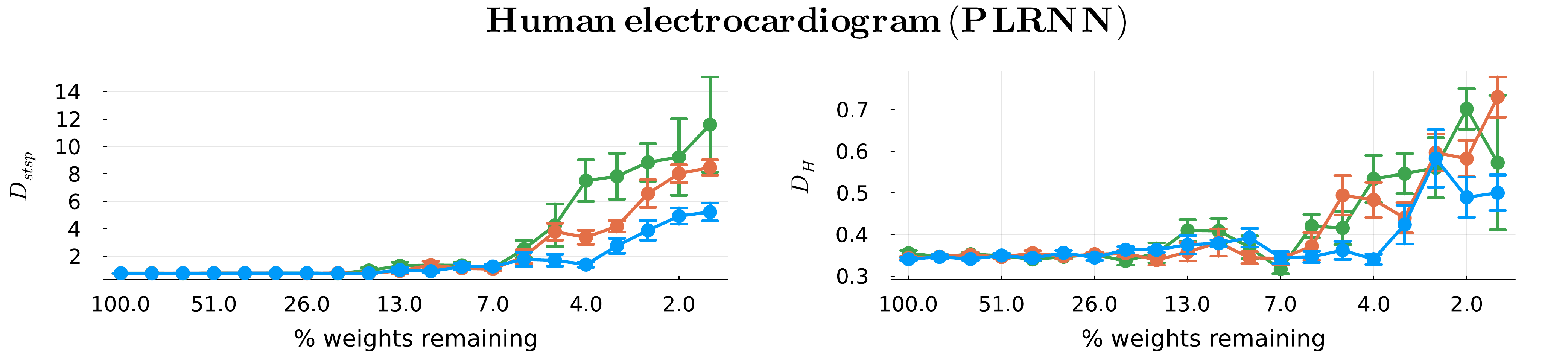}
    \end{subfigure}
    \begin{subfigure}
         \centering
         \includegraphics[width=0.9\textwidth]{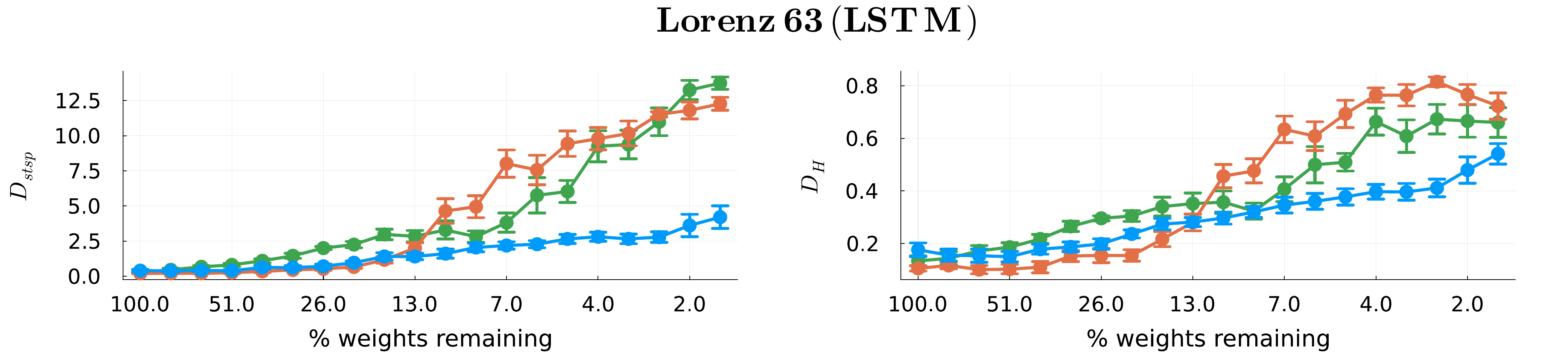}
    \end{subfigure}
    \caption{Quantification of DS reconstruction quality in terms of attractor geometry disagreement ($D_{\text{stsp}}$, left column) and 
    disagreement in long-term temporal structure ($D_{\text{H}}$, right column) as a function of network pruning (x-axis, exponential scale) and different pruning criteria. Error bars = SEM.
    }
    \label{fig:pruning_measures_performance}
\end{figure*}

\section{Results}
\label{results}
\subsection{Performance Evaluation}
We used well-established performance criteria to evaluate the DS reconstruction quality of trained networks \cite{koppe_fmri_2019, brenner2022tractable, hess2023generalized}. Because of exponential trajectory divergence in chaotic systems, mean-squared prediction errors are only of limited use as they may become large quickly even for well-trained systems \cite{wood_statistical_2010, mikhaeil2022difficulty}.\footnote{Counterintuitively, as exemplified in \citet{koppe_fmri_2019}, they can even become \textit{larger} for perfect than for poorer reconstructions, because in the longer run only the mean may be well predictable for chaotic systems.} Instead, we focus on the \textit{geometrical} agreement between true and reconstructed attractors as quantified through a Kullback-Leibler divergence ($D_{\text{stsp}}$) as suggested in (\citet{koppe_fmri_2019}; see also Appx. \ref{sect:evaluation_measures}), and on the long-term \textit{temporal} agreement between true and reconstructed (generated) time series assessed by the average dimension-wise Hellinger distance ($D_{\text{H}}$) between true and reconstructed power spectra (equivalently, autocorrelation coefficients may be used; see Appx. \ref{sect:evaluation_measures} for more details). Note that while $D_{\text{stsp}}$ and $D_{\text{H}}$ will be correlated in good reconstructions, they assess fundamentally different, complementary aspects of the dynamics. 

\subsection{Geometry-Based, but not Magnitude-Based, Pruning Allows for Substantial Reduction in Network Size} \label{sect:Pruning_results}
Figs. \ref{fig:geometry-based_pruning}b \& c show that there is hardly any (or at most very small) difference in the absolute magnitude of PLRNN connection weights contributing substantially vs. essentially non-contributing to geometric reconstruction quality (see Fig. \ref{fig:weight_mag_Dstsp_further}b for a further example), regardless of whether several weights are removed individually or simultaneously (Fig. \ref{fig:geometry-based_pruning}d, Fig. \ref{fig:weight_mag_Dstsp_further}a). This raises the question of whether the parameter size of RNNs trained for DSR can be reduced beyond what would be expected by just random removal of connections. Using geometric pruning, however, we found that reductions by up to $95\%$ for some systems are indeed possible (Fig. \ref{fig:pruning_measures_performance}). More specifically, we evaluated DSR performance on several DS benchmarks for three different iterative pruning protocols (Algorithm \ref{alg:Pruning}; Fig. \ref{fig:pruning_measures_performance}). First, the \textit{Lorenz-63} model of atmospheric convection, proposed by Edward Lorenz \cite{DeterministicNonperiodicFlow}, produces a chaotic attractor with iconic butterfly-wing structure (Fig. \ref{fig:geometry-based_pruning}) and is probably the most commonly employed benchmark in this whole literature. Second, we use a simplified biophysical model of a \textit{bursting neuron} \cite{DURSTEWITZ20091189} which produces fast spiking (action potential) activity on top of a slow oscillation (see Fig. \ref{fig:BurstingNeuron}), thus featuring two widely different time scales. It has been previously used to evaluate DSR models \cite{brenner2022tractable}, and can also generate chaotic activity within some parameter regimes \cite{DURSTEWITZ20091189}, but was employed here as an example of a system with a complex (multi-period) limit cycle as in previous work. Third, as a real-world example, we used \textit{human electrocardiogram (ECG)} data bearing signatures of chaos, with a positive maximum Lyapunov exponent (see \citet{hess2023generalized}). We also tested DSR on the \textit{Rössler} attractor \cite{ROSSLER1976397}, a simplification of the Lorenz-63 system, on the Lorenz-63 system with high levels of observation noise ($25\%$), and on the \textit{Lorenz-96} model \cite{lorenz96}, a higher-dimensional spatial extension of the earlier Lorenz-63 model which also produces (highly) chaotic behavior for the parameter setup chosen here (we use a $5d$ system in our experiments, Fig. \ref{fig:Lorenz96}; see Appx. \ref{Benachmark data} for details on all models, and Appx. \ref{ch:Methodological_details} for detailed hyper-parameter settings used in RNN training). 

Fig. \ref{fig:pruning_measures_performance} illustrates DSR performance on the Lorenz-63, the bursting neuron, and the ECG benchmarks as a function of network size, i.e. percentage of pruned parameters, for magnitude- compared to geometry-based pruning (results for all other benchmarks are in Fig. \ref{fig:pruning_measures_performance_further}). As a baseline, we also included a random pruning protocol, where parameters for removal were just chosen at random. In agreement with the observations in Fig. \ref{fig:geometry-based_pruning}, we found that, on average, for all benchmarks, and for both the geometrical and temporal (dis)agreement measure, the effects of magnitude-based pruning essentially were not that much different than if weights were removed just randomly (statistically, by repeated measures ANOVAs, the differences were indeed insignificant in 5/10 comparisons across both $D_{\text{stsp}}$ and $D_{\text{H}}$, and significant in the remaining 5/10 cases, but with random better than magnitude in one of these). This confirms that the absolute size of a weight parameter is less indicative of its contribution to RNN performance. Yet, using geometric pruning, network size in general could be reduced substantially beyond that of random pruning (significantly so in all 10/10 cases, $p<0.03$), with sometimes just about $5\%$ of the weight parameters sufficient to optimally reconstruct the ground truth DS. This was apparent not only in the geometrical measure $D_{\text{stsp}}$ (left column in Fig. \ref{fig:pruning_measures_performance}), but also in the temporal (dis)agreement measure $D_{\text{H}}$ (right column in Fig. \ref{fig:pruning_measures_performance}) which was not a criterion used in the pruning process (as well as in short-term prediction errors, see Fig. \ref{fig:pruning_performance_PE}). 

Furthermore, these effects were present across networks of different initial sizes (Fig. \ref{fig:pruning_smaller_initial_modelsize}): Specifically, while the LTH suggests starting with strongly over-parameterized systems to enhance the chances for a winning ticket, which then naturally can be substantially pruned down \cite{frankle2019lottery,frankle2020linear,malach2020proving}, the differences between geometry- and magnitude-based pruning persisted in much smaller networks (Fig. \ref{fig:pruning_smaller_initial_modelsize}). Finally, similar results were obtained for other types of RNN architectures (LSTMs: Fig. \ref{fig:pruning_measures_performance}, bottom; vanilla RNNs: Fig. \ref{fig:pruning_RNN}), implying that these observations are not specific to PLRNNs but more general. For LSTMs we furthermore observed that geometrical pruning identified the relevance of the model's different weight matrices to the performance, leading to interpretable results in terms of the inter-cell connectivity (Fig. \ref{fig:LSTM_topology}). We conclude that a substantial reduction in parameter set size is indeed possible, but not so much based on the more traditional criterion of weight magnitude \cite{blalock2020state}.

\subsection{Network Topology, not Weight Configuration is Essential to Performance}
\begin{figure}[ht!]
    \centering
    \includegraphics[width=0.49\textwidth]{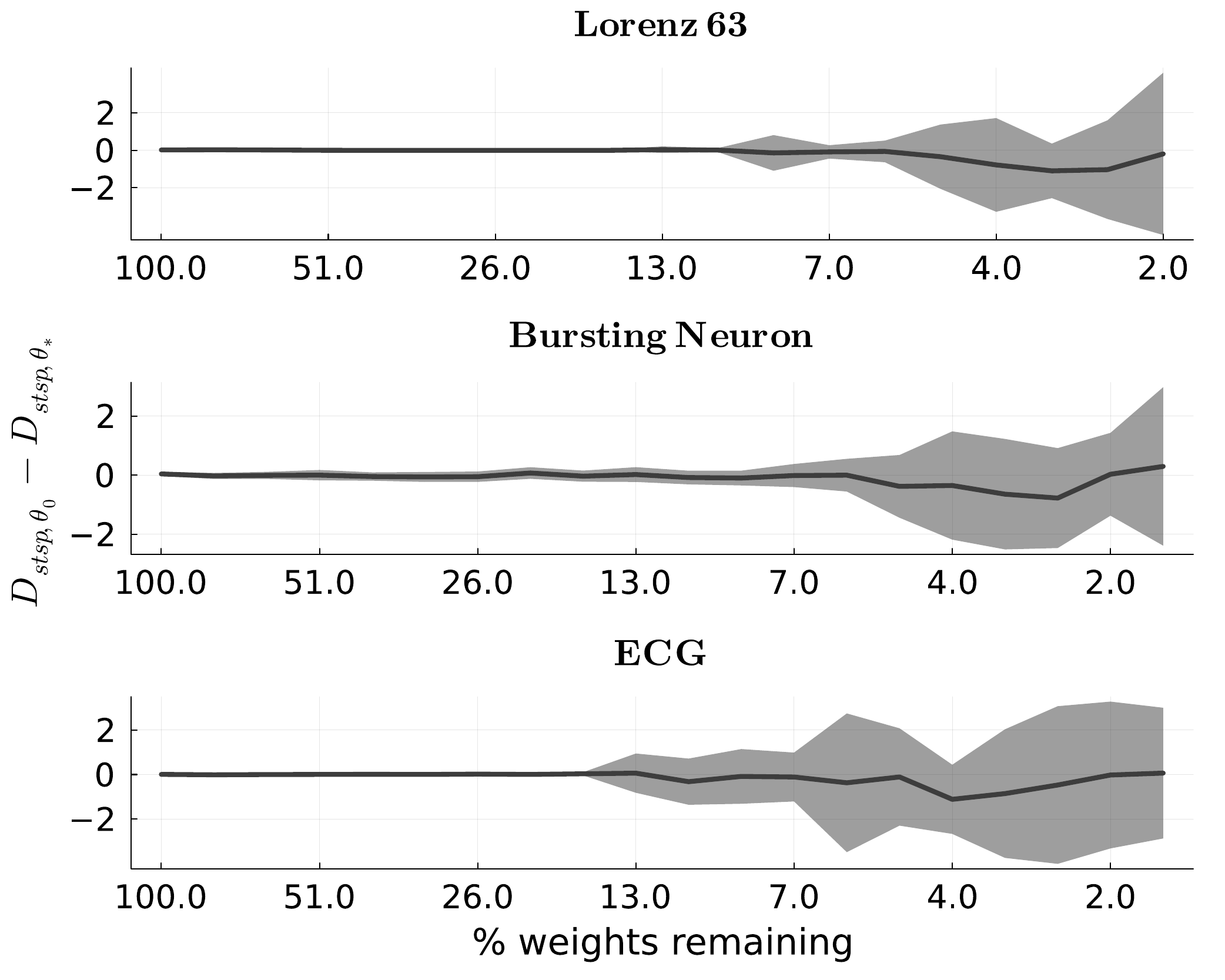}
    \caption{Difference in $D_{\text{stsp}}$ when using the initial weights $\bm{\theta}_0$ and reinitialized weights     $\bm{\theta}_*$ shows there is no strong or consistent influence of the specific weight initialization. Error bands = standard deviation.}
    \label{fig:initial_weights}
\end{figure}

The LTH poses that it is the topology of an embedded subnetwork $\bm{m}$ in conjunction with a specific random initialization of model parameters $\bm{\theta}_0$ of this subnetwork which is crucial for its success. The fact that absolute weight magnitude plays less of a role in performance already sheds doubt on this idea in the context of DSR. To more explicitly test this and disentangle the contributions of mask $\bm{m}$ and weights $\bm{\theta}_0$ to the DSR of geometrically pruned networks, we resampled network parameters $\bm{\theta}_* \sim \mathcal{N}(0, \sigma^2 \bm{I})$ with a fixed mask $\bm{m}$ from the very same distribution, from which the initial estimate $\bm{\theta}_0$ had been drawn, and compared this to the standard LTH case where $\bm{\theta}_0$ is fixed after the initial draw. We found that the influence of the network topology, given by the mask $\bm{m}$, far outweighed the importance of the specific initial weight vector $\bm{\theta}_0$: Redrawing $\bm{\theta}_*$ from scratch vs. fixing it to the initial $\bm{\theta}_0$ did not make much difference for DSR performance (Fig. \ref{fig:initial_weights}; see also Fig. \ref{fig:reinitialization_Dhell} for the same results on $D_{H}$), highlighting the crucial role network topology plays in the context of DSR. This is good news: in the `classical' LTH, masks and weight distributions are tied in a specific way and therefore hard to disentangle. This in turn implies that the specific configuration that led to the winning ticket is difficult to reverse-engineer, and hence computationally costly iterative pruning schemes are required. However, given that in our case performance gains are primarily driven by topological structure and not parameter distribution, this structure can be distilled from trained RNNs and reverse-engineered with tools well-known from graph theory, as discussed next.
 
\subsection{Distilling Network Topology for Enhanced DSR Training} \label{sect:network_topology_training} 
\begin{figure}[h]
    \centering
    \includegraphics[width=0.5\textwidth]{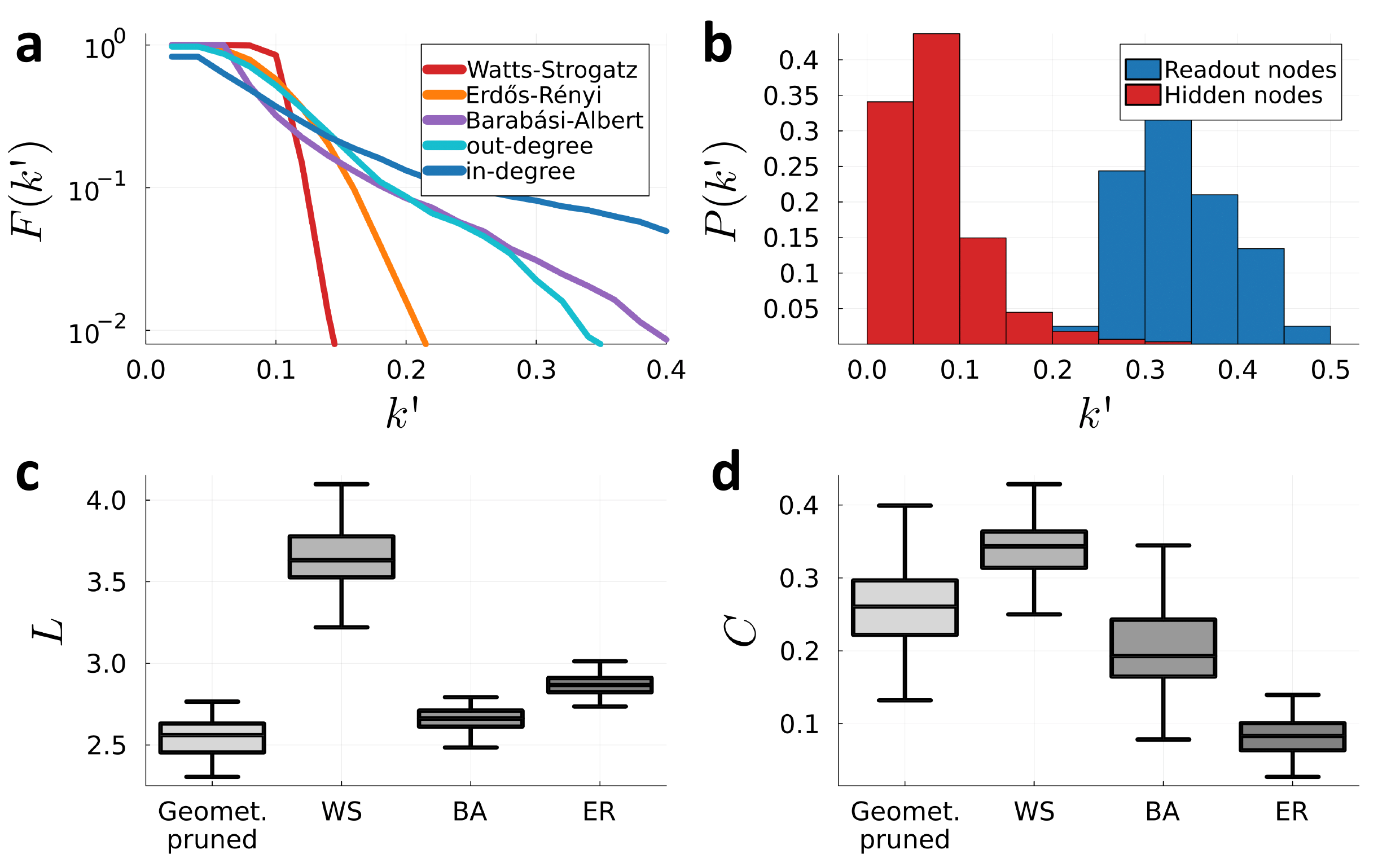}
    \caption{Graph properties of geometrically pruned, Barabási-Albert (BA), Watts-Strogatz (WS), and Erdős–Rényi (ER) networks, with $92.4\%$ of parameters removed and averaged across all datasets with $M=50$. 
    \textbf{a}) Cumulative degree distribution $F(k')$ as a function of normalized degree $k'=\frac{k}{n-1}$, separated according to in- and out-degree (for geometrically pruned network). 
    \textbf{b}) Comparison of degree distributions $P(k')$ for readout vs. hidden nodes of geometrically pruned networks. 
    \textbf{c}) Average path lengths $L$ for all four network topologies. Note that Erdős–Rényi graphs are not a naive baseline here, but are also known to have small path length \cite{watts_collective_1998}. 
    \textbf{d}) Clustering coefficients $C$ for the same. See also Fig. \ref{fig:network_graph_properties_M100}.}
    \label{fig:network_graph_properties}
\end{figure}

Next, we analyzed the topological properties of geometry-pruned networks in order to identify the crucial features that led to superior performance.\footnote{Recall that well trained but unpruned RNNs always have full, all-to-all connectivity.}
We find that these topologies contain both hub-type as well as small-world characteristics (Fig. \ref{fig:network_graph_properties}; details in Appx. \ref{PLRNN topology}, see Fig. \ref{fig:graph_examples} for specific examples of the different network topologies): 
As typical for small-world networks like the Watts-Strogatz model, geometrically pruned RNNs were characterized by a small average path length $L$ (Fig. \ref{fig:network_graph_properties}c; see Fig. \ref{fig:network_graph_properties_M100} for similar effects in larger networks) as well as a high clustering coefficient $C$ (Fig. \ref{fig:network_graph_properties}d, Fig. \ref{fig:network_graph_properties_M100}). At the same time, as in scale-free networks like the Barabási-Albert model (Fig. \ref{fig:graph_examples}), geometrically pruned RNNs bear a hub-like structure with a few highly connected network nodes (Fig. \ref{fig:network_graph_properties}a, Fig. \ref{fig:network_graph_properties_M100}). 
We combined these features into an algorithm (Algorithm \ref{alg:PLRNN_subnetwork_graph}) that automatically produces RNN connectivity structures with these desired properties, which we call `GeoHub' (for geometrically-pruned-hub network). 
\begin{figure}[ht!]
\centering
\begin{minipage}{.9\linewidth}
\begin{algorithm}[H]
\SetKwInOut{Input}{Input}\SetKwInOut{Output}{Output}
\Input{Number of nodes $n$, mean in- and out-degree $k$, dimension $N$ of the data}
\Output{GeoHub graph $G(V,E)$}
\BlankLine
$G \gets$ fully connected graph with $N$ nodes

Add $n-N$ nodes

\For{$i \gets 1$ \KwTo $n$}{

  \For{$l \gets 1$ \KwTo $k/2$}{
  Select node $v_j$ with probability \\
  $p_j=\begin{cases}
        \frac{1}{\mathcal{N}}\;(k^{in}_j+ k/2), & \text{if $j\in 1,...,N$} \\
        \frac{1}{\mathcal{N}}\;k^{in}_j & \text{if $j\in N+1,...,n$}
    \end{cases}$\;
    where $\mathcal{N}$ is a normalization factor (see Appx. \ref{PLRNN topology})
  
    $E \gets E \cup \{e_{ij}\}$ 
  }
}
\For{$i \gets 1$ \KwTo $n$}{
  \For{$l \gets 1$ \KwTo $k/2$}{
  Select node $v_j$ with probability \\
  $p_j=\frac{1}{\mathcal{N}}\;(k^{in}_j+k^{out}_j+ k/4)$\;\\
  with normalization $\mathcal{N}$
  
    $E \gets E \cup \{e_{ji}\}$ 
  }
}
\caption{GeoHub network algorithm}
\label{alg:PLRNN_subnetwork_graph}
\end{algorithm}
\end{minipage}
\vspace{-.5cm}
\end{figure}

\begin{figure}[ht!]
    \centering
    \begin{subfigure}
         \centering
         \includegraphics[width=0.45\textwidth]{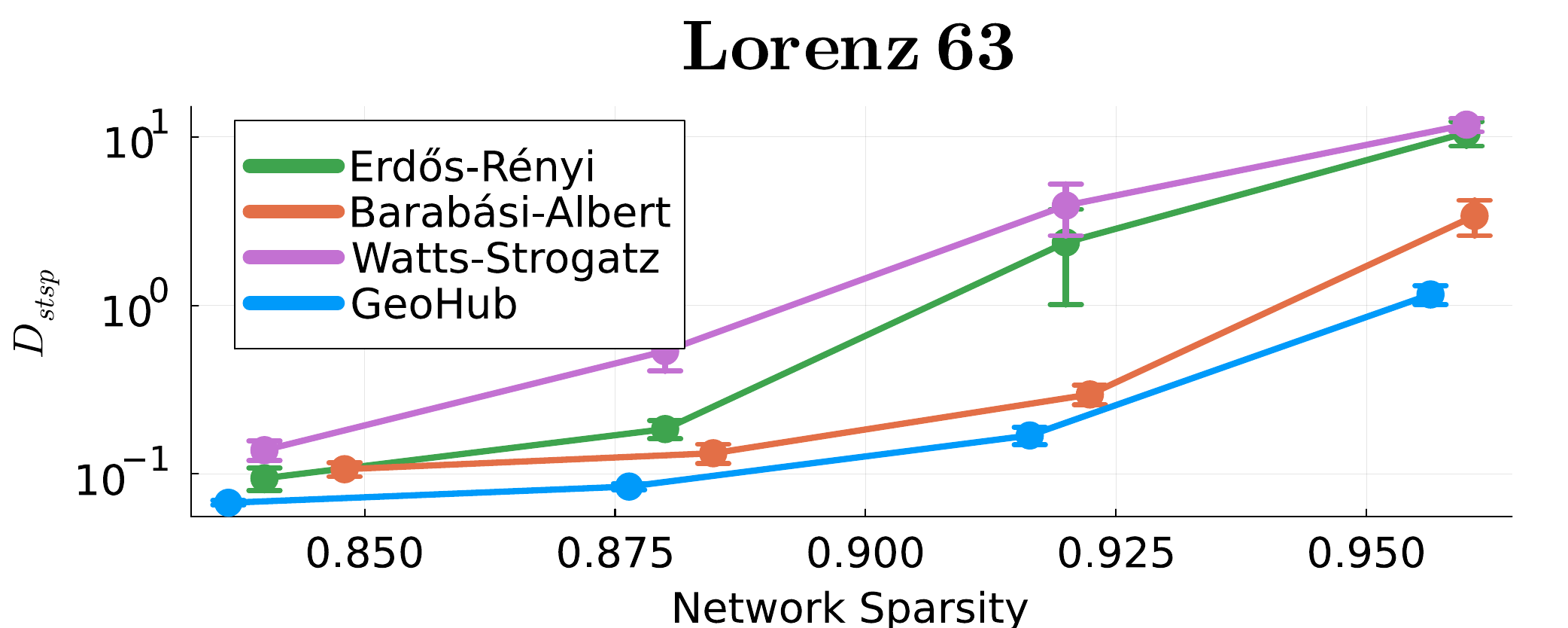}
    \end{subfigure}
    \begin{subfigure}
         \centering
         \includegraphics[width=0.45\textwidth]{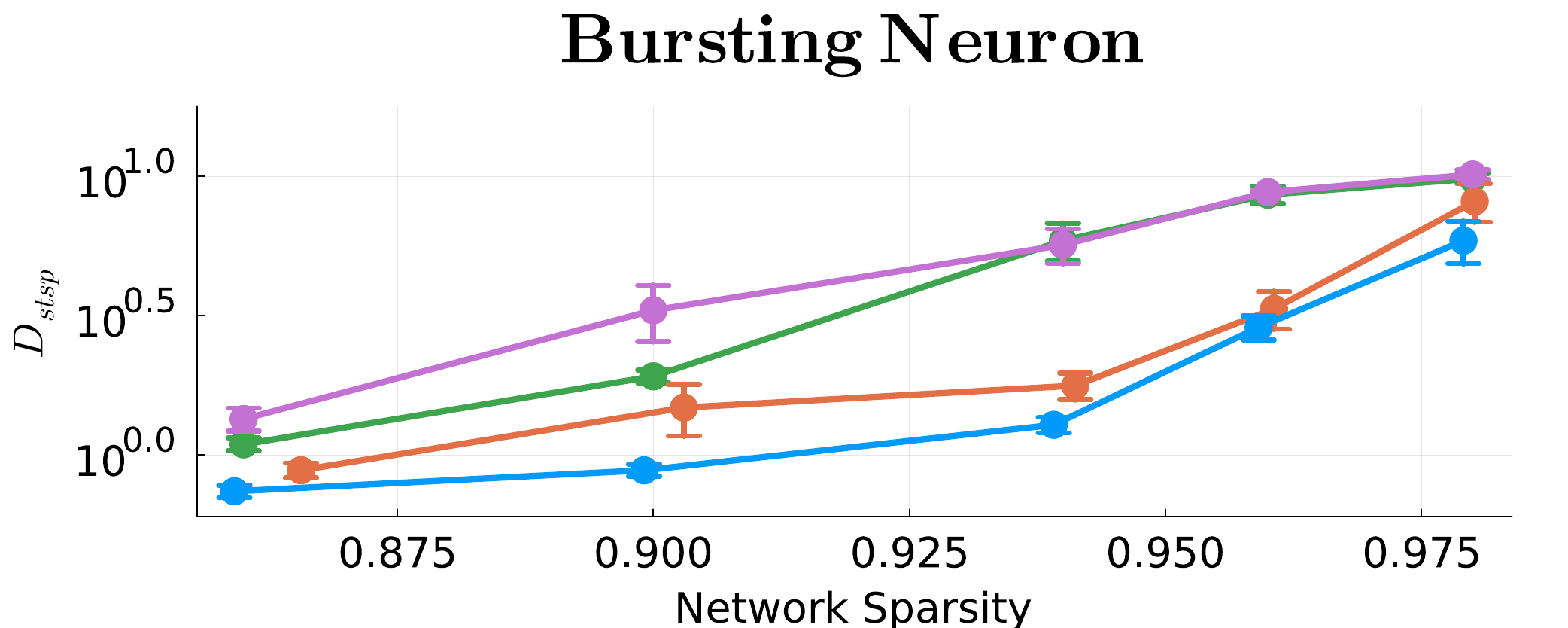}
    \end{subfigure}
    \begin{subfigure}
         \centering
         \includegraphics[width=0.45\textwidth]{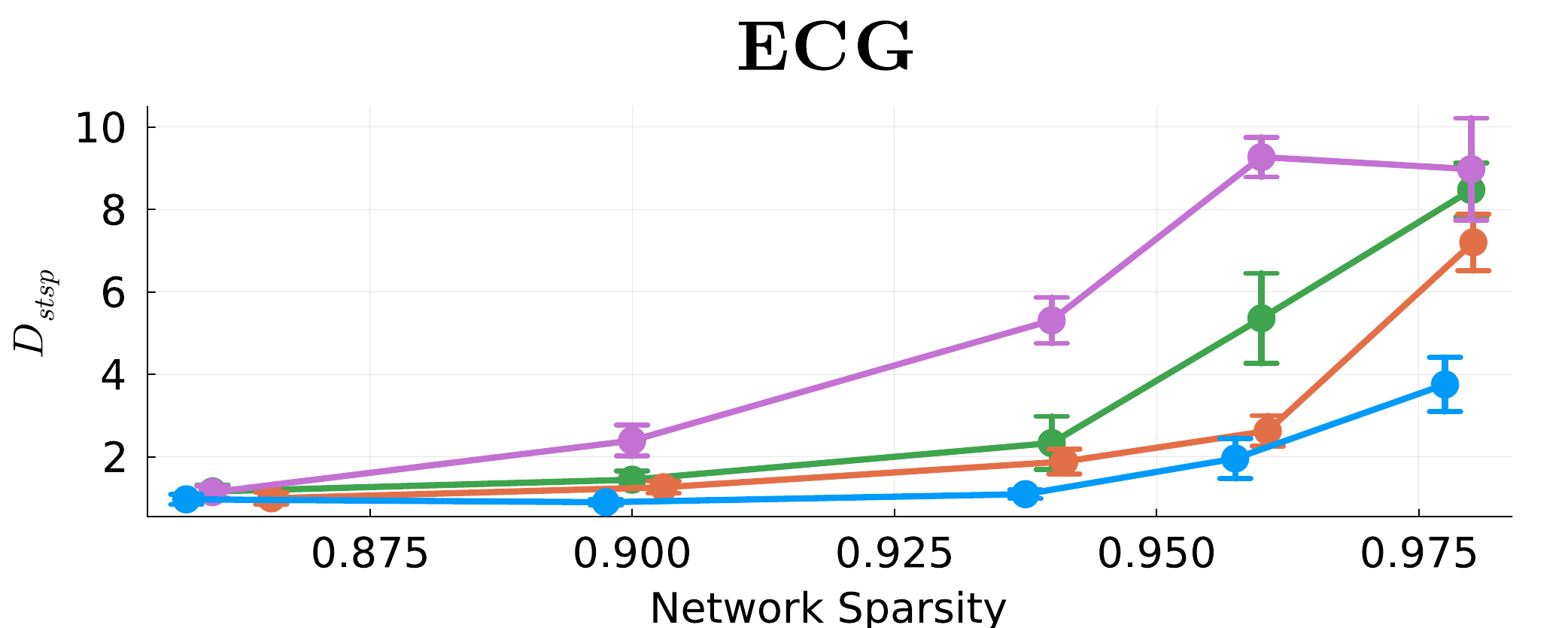}
    \end{subfigure}
    \vspace{-.2cm}
    \caption{Reconstruction results in terms of state space divergence $D_{\text{stsp}}$ as a function of network sparsity $s=1-\frac{| \bm{m}|}{|\bm{W}|}$. Erdős–Rényi, Barabási-Albert, and Watts-Strogatz graph algorithms are described in detail in \ref{ch:graph_models}. Error bars = SEM.}
    \label{fig:graph_netork_performance}
\end{figure}

\begin{figure}[ht!]
    \centering
    \includegraphics[width=0.5\textwidth]{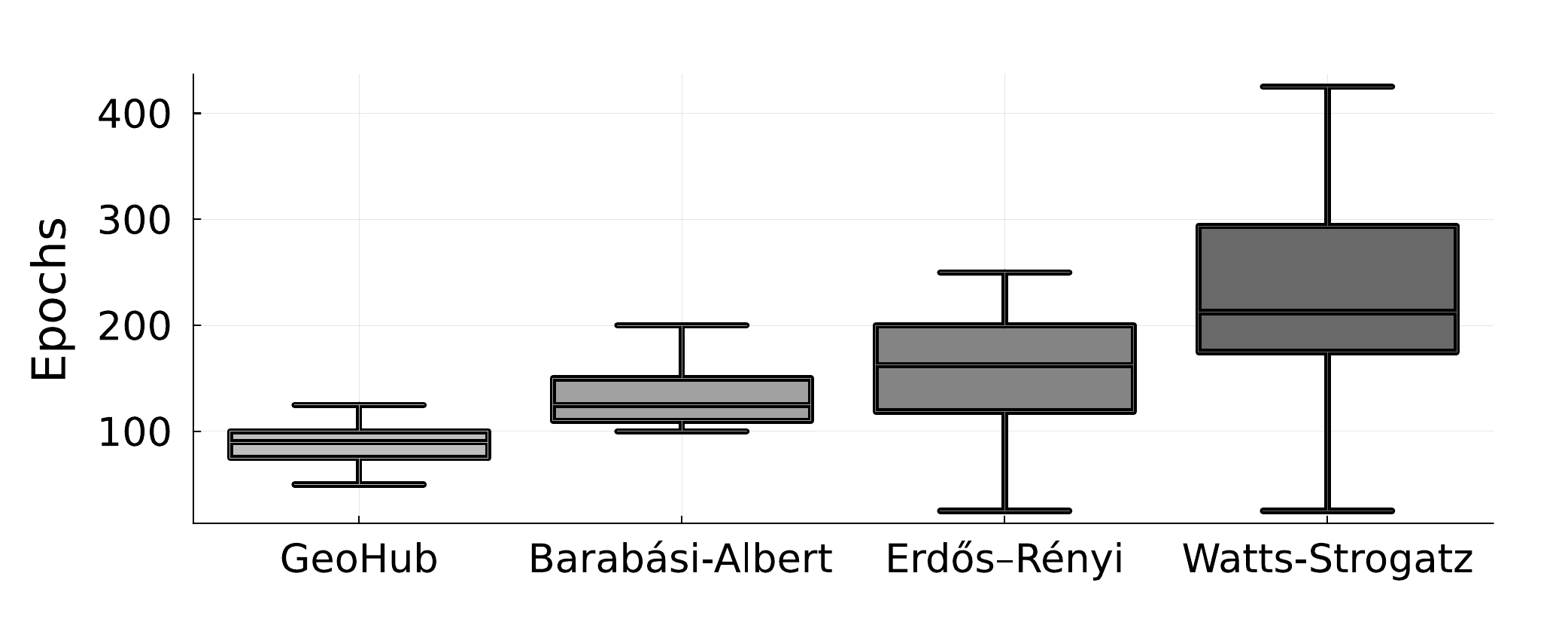}
    \vspace{-.7cm}
    \caption{Epoch at which reasonable DSR performance ($D_{\text{stsp}}<1.0$) is obtained for the different network topologies when trained on the Lorenz-63. Each network contains $\approx 300$ parameters.}
    \label{fig:train_speed}
    
\end{figure}

In Fig. \ref{fig:graph_netork_performance} we compare the DSR performance of RNNs trained with GeoHub topology to those based on the classical Watts-Strogatz and Barabási-Albert models. As evident, GeoHub-based networks perform best on all three benchmark setups employed in this comparison, whether chaotic (Lorenz-63), complex but non-chaotic (bursting neuron), or real-world ECG data, closely followed by RNNs with a Barabási-Albert graph (see Fig. \ref{fig:graph_performance_Lorenz96} for comparisons on other benchmarks). As a further baseline, we also included RNNs based on an Erdős–Rényi random graph in this comparison. Surprisingly, although small-world features appeared to be necessary to move DSR performance beyond that obtained by the scale-free Barabási-Albert structure alone, a pure small-world structure (Watts-Strogatz model) actually appeared to \textit{diminish} performance compared to the Erdős–Rényi graph models. Finally, we observed that RNNs initialized with optimal topology do not only outperform other graph structures, but also train significantly faster, i.e. reach satisfying DSR performance in fewer epochs than other topologies, as illustrated in Fig. \ref{fig:train_speed}.

In summary, our results show that the best performing graph model is the one which replicates the topology empirically obtained through geometry-based pruning, and that initializing based on this topology alone yields sparse networks with performance rivaling that of fully connected RNNs.

\section{Conclusions}
\label{discussion}
In this work we reported on a surprising observation: Setting up the \textit{right network topology alone} is sufficient to produce a highly sparsified generative RNN which recapitulates the geometrical and temporal properties of observed DS about as well as a fully parameterized network. Beyond proper network topology, neither the specific parameter initialization, nor their absolute magnitude, were that crucial to DSR performance. Such networks, as obtained by geometric pruning, with a mixture of hub-type and small-world characteristics, not only profoundly reduce parameter and memory load, but also train much faster than comparably sized RNNs with random topology. Since we observed this for a diverse set of benchmarks, both chaotic and non-chaotic, low- and higher-dimensional, evolving on multiple time scales, as well as for different types of RNNs, these results appear to be more general, although follow-up studies to confirm this for a wider range of systems and NN architectures are desirable. 

Our results suggest a new type of LTH \cite{frankle2019lottery} in the context of DSR, where the winning tickets in largely over-parameterized networks depend less on specific weights but more on abstract topological features. It remains to be examined why exactly this is the case, how widely this holds, and whether the type of topology will be similar across different systems. For instance, the LTH so far has been mainly explored in the context of feed-forward architectures like CNNs. Is recurrence in the network a crucial feature and do our results therefore also generalize to other sequence and time series models? Or is it specific to the DSR problem, where we also want to capture geometric and long-term temporal properties of the data-generating system? Since many real-world systems (on which many of the benchmarks are based) have specific topological properties due to physical constraints, like scale-free or small-worldness, it is also conceivable that for DSR optimal RNN structure to some extent mirrors this empirical observation. It would be interesting to explore whether these results hold beyond the natural science domain, in areas like NLP for instance. 

Finally, note that geometric pruning here mainly served as a tool to examine topological factors important in DSR. However, computationally efficient implementations of it, which make this technique directly applicable, are also conceivable, e.g. based on efficient (dimension-wise and parallelizable) proxies for $D_{stsp}$. There is also room for improvement, for instance by incorporating invariant temporal structure into the pruning process.

\section*{Software and Data}
All code created is available at \url{https://github.com/DurstewitzLab/RNNtopoDSR}.

\section*{Acknowledgements}
This work was funded by the German Research Foundation (DFG) through individual grant Du 354/15-1 to DD, and within Germany’s Excellence Strategy EXC 2181/1 – 390900948 (STRUCTURES).

\section*{Impact Statement}

This paper presents work whose goal is to advance the field of theoretical Machine Learning. There are many potential societal consequences of our work, none which we feel must be specifically highlighted here.

\bibliography{references}

\begin{thebibliography}{95}
\providecommand{\natexlab}[1]{#1}
\providecommand{\url}[1]{\texttt{#1}}
\expandafter\ifx\csname urlstyle\endcsname\relax
  \providecommand{\doi}[1]{doi: #1}\else
  \providecommand{\doi}{doi: \begingroup \urlstyle{rm}\Url}\fi

\bibitem[Albert \& Barab{\'{a}}si(2002)Albert and Barab{\'{a}}si]{Albert_2002}
Albert, R. and Barab{\'{a}}si, A.-L.
\newblock Statistical mechanics of complex networks.
\newblock \emph{Reviews of Modern Physics}, 74\penalty0 (1):\penalty0 47--97, 2002.
\newblock \doi{10.1103/revmodphys.74.47}.
\newblock URL \url{https://doi.org/10.1103%2Frevmodphys.74.47}.

\bibitem[Albert \& Barab{\'a}si(2002)Albert and Barab{\'a}si]{albert2002statistical}
Albert, R. and Barab{\'a}si, A.-L.
\newblock Statistical mechanics of complex networks.
\newblock \emph{Reviews of modern physics}, 74\penalty0 (1):\penalty0 47, 2002.

\bibitem[Alvarez et~al.(2020)Alvarez, Roşca, and Fălcuţescu]{alvarez_dynode_2020}
Alvarez, V. M.~M., Roşca, R., and Fălcuţescu, C.~G.
\newblock {DyNODE}: {Neural} {Ordinary} {Differential} {Equations} for {Dynamics} {Modeling} in {Continuous} {Control}, September 2020.
\newblock URL \url{http://arxiv.org/abs/2009.04278}.
\newblock arXiv:2009.04278 [cs, eess, stat].

\bibitem[Amaral et~al.(2000)Amaral, Scala, Barthelemy, and Stanley]{amaral2000classes}
Amaral, L. A.~N., Scala, A., Barthelemy, M., and Stanley, H.~E.
\newblock Classes of small-world networks.
\newblock \emph{Proceedings of the national academy of sciences}, 97\penalty0 (21):\penalty0 11149--11152, 2000.

\bibitem[Ba et~al.(2016)Ba, Kiros, and Hinton]{ba2016layer}
Ba, J.~L., Kiros, J.~R., and Hinton, G.~E.
\newblock Layer normalization.
\newblock \emph{arXiv preprint arXiv:1607.06450}, 2016.

\bibitem[Barabási \& Albert(1999)Barabási and Albert]{Barabasi99emergenceScaling}
Barabási, A.-L. and Albert, R.
\newblock Emergence of scaling in random networks.
\newblock \emph{Science}, 286\penalty0 (5439):\penalty0 509--512, 1999.
\newblock \doi{10.1126/science.286.5439.509}.
\newblock URL \url{http://www.sciencemag.org/cgi/content/abstract/286/5439/509}.

\bibitem[Bassett \& Bullmore(2006)Bassett and Bullmore]{bassett2006small}
Bassett, D.~S. and Bullmore, E.
\newblock Small-world brain networks.
\newblock \emph{The neuroscientist}, 12\penalty0 (6):\penalty0 512--523, 2006.

\bibitem[Beggs \& Plenz(2003)Beggs and Plenz]{beggs_neuronal_2003}
Beggs, J.~M. and Plenz, D.
\newblock Neuronal avalanches in neocortical circuits.
\newblock \emph{The Journal of Neuroscience: The Official Journal of the Society for Neuroscience}, 23\penalty0 (35):\penalty0 11167--11177, December 2003.
\newblock ISSN 1529-2401.
\newblock \doi{10.1523/JNEUROSCI.23-35-11167.2003}.

\bibitem[Belkin et~al.(2019)Belkin, Hsu, Ma, and Mandal]{Belkin_2019}
Belkin, M., Hsu, D., Ma, S., and Mandal, S.
\newblock Reconciling modern machine-learning practice and the classical bias–variance trade-off.
\newblock \emph{Proceedings of the National Academy of Sciences}, 116\penalty0 (32):\penalty0 15849–15854, 2019.
\newblock ISSN 1091-6490.
\newblock \doi{10.1073/pnas.1903070116}.
\newblock URL \url{http://dx.doi.org/10.1073/pnas.1903070116}.

\bibitem[Blalock et~al.(2020)Blalock, Gonzalez~Ortiz, Frankle, and Guttag]{blalock2020state}
Blalock, D., Gonzalez~Ortiz, J.~J., Frankle, J., and Guttag, J.
\newblock What is the state of neural network pruning?
\newblock \emph{Proceedings of machine learning and systems}, 2:\penalty0 129--146, 2020.

\bibitem[Brenner et~al.(2022)Brenner, Hess, Mikhaeil, Bereska, Monfared, Kuo, and Durstewitz]{brenner2022tractable}
Brenner, M., Hess, F., Mikhaeil, J.~M., Bereska, L., Monfared, Z., Kuo, P.-C., and Durstewitz, D.
\newblock Tractable dendritic rnns for reconstructing nonlinear dynamical systems.
\newblock In \emph{Proceedings of the 39th International Conference on Machine Learning}, pp.\  2292--2320. PMLR, 2022.

\bibitem[Brunton et~al.(2016)Brunton, Proctor, and Kutz]{brunton2016discovering}
Brunton, S.~L., Proctor, J.~L., and Kutz, J.~N.
\newblock Discovering governing equations from data by sparse identification of nonlinear dynamical systems.
\newblock \emph{Proceedings of the national academy of sciences}, 113\penalty0 (15):\penalty0 3932--3937, 2016.

\bibitem[Bullmore \& Sporns(2009)Bullmore and Sporns]{bullmore2009complex}
Bullmore, E. and Sporns, O.
\newblock {Complex brain networks: graph theoretical analysis of structural and functional systems}.
\newblock \emph{Nature Reviews Neuroscience}, 10\penalty0 (3):\penalty0 186--198, 2009.

\bibitem[Bullmore \& Sporns(2012)Bullmore and Sporns]{bullmore2012economy}
Bullmore, E. and Sporns, O.
\newblock The economy of brain network organization.
\newblock \emph{Nature reviews neuroscience}, 13\penalty0 (5):\penalty0 336--349, 2012.

\bibitem[Burkholz et~al.(2022)Burkholz, Laha, Mukherjee, and Gotovos]{burkholz_existence_2022}
Burkholz, R., Laha, N., Mukherjee, R., and Gotovos, A.
\newblock On the {Existence} of {Universal} {Lottery} {Tickets}, March 2022.
\newblock URL \url{http://arxiv.org/abs/2111.11146}.
\newblock arXiv:2111.11146 [cs, stat].

\bibitem[Carroll \& Pecora(2019)Carroll and Pecora]{carroll_network_2019}
Carroll, T.~L. and Pecora, L.~M.
\newblock Network {Structure} {Effects} in {Reservoir} {Computers}.
\newblock \emph{Chaos: An Interdisciplinary Journal of Nonlinear Science}, 29\penalty0 (8):\penalty0 083130, August 2019.
\newblock ISSN 1054-1500, 1089-7682.
\newblock \doi{10.1063/1.5097686}.
\newblock URL \url{http://arxiv.org/abs/1903.12487}.
\newblock arXiv:1903.12487 [nlin].

\bibitem[Champion et~al.(2019)Champion, Lusch, Kutz, and Brunton]{champion_data-driven_2019}
Champion, K., Lusch, B., Kutz, J.~N., and Brunton, S.~L.
\newblock Data-driven discovery of coordinates and governing equations.
\newblock \emph{Proceedings of the National Academy of Sciences}, 116\penalty0 (45):\penalty0 22445--22451, November 2019.
\newblock ISSN 0027-8424, 1091-6490.
\newblock \doi{10.1073/pnas.1906995116}.
\newblock URL \url{http://arxiv.org/abs/1904.02107}.
\newblock arXiv:1904.02107 [stat].

\bibitem[Chatzikonstantinou et~al.(2021)Chatzikonstantinou, Konstantinidis, Dimitropoulos, and Daras]{CHATZIKONSTANTINOU2021475}
Chatzikonstantinou, C., Konstantinidis, D., Dimitropoulos, K., and Daras, P.
\newblock Recurrent neural network pruning using dynamical systems and iterative fine-tuning.
\newblock \emph{Neural Networks}, 143:\penalty0 475--488, 2021.
\newblock ISSN 0893-6080.
\newblock \doi{https://doi.org/10.1016/j.neunet.2021.07.001}.
\newblock URL \url{https://www.sciencedirect.com/science/article/pii/S0893608021002641}.

\bibitem[Chen et~al.(2018)Chen, Rubanova, Bettencourt, and Duvenaud]{chen2018neural}
Chen, R.~T., Rubanova, Y., Bettencourt, J., and Duvenaud, D.~K.
\newblock Neural ordinary differential equations.
\newblock \emph{Advances in neural information processing systems}, 31, 2018.

\bibitem[Dale et~al.(2021)Dale, O’Keefe, Sebald, Stepney, and Trefzer]{dale_reservoir_2021}
Dale, M., O’Keefe, S., Sebald, A., Stepney, S., and Trefzer, M.~A.
\newblock Reservoir computing quality: connectivity and topology.
\newblock \emph{Natural Computing}, 20\penalty0 (2):\penalty0 205--216, June 2021.
\newblock ISSN 1572-9796.
\newblock \doi{10.1007/s11047-020-09823-1}.
\newblock URL \url{https://doi.org/10.1007/s11047-020-09823-1}.

\bibitem[Diestel(2005)]{graph_theory_diestel}
Diestel, R.
\newblock \emph{Graph Theory (Graduate Texts in Mathematics)}.
\newblock Springer, 2005.
\newblock ISBN 3540261826.

\bibitem[Durstewitz(2009)]{DURSTEWITZ20091189}
Durstewitz, D.
\newblock Implications of synaptic biophysics for recurrent network dynamics and active memory.
\newblock \emph{Neural Networks}, 22\penalty0 (8):\penalty0 1189--1200, 2009.
\newblock ISSN 0893-6080.
\newblock \doi{https://doi.org/10.1016/j.neunet.2009.07.016}.
\newblock URL \url{https://www.sciencedirect.com/science/article/pii/S0893608009001622}.
\newblock Cortical Microcircuits.

\bibitem[Durstewitz(2017)]{10.1371/journal.pcbi.1005542}
Durstewitz, D.
\newblock A state space approach for piecewise-linear recurrent neural networks for identifying computational dynamics from neural measurements.
\newblock \emph{PLOS Computational Biology}, 13\penalty0 (6):\penalty0 1--33, 06 2017.
\newblock \doi{10.1371/journal.pcbi.1005542}.
\newblock URL \url{https://doi.org/10.1371/journal.pcbi.1005542}.

\bibitem[Durstewitz et~al.(2023)Durstewitz, Koppe, and Thurm]{durstewitz2023reconstructing}
Durstewitz, D., Koppe, G., and Thurm, M.~I.
\newblock Reconstructing computational system dynamics from neural data with recurrent neural networks.
\newblock \emph{Nature Reviews Neuroscience}, pp.\  1--18, 2023.

\bibitem[Dutoit et~al.(2009)Dutoit, Schrauwen, Van~Campenhout, Stroobandt, Van~Brussel, and Nuttin]{dutoit_pruning_2009}
Dutoit, X., Schrauwen, B., Van~Campenhout, J., Stroobandt, D., Van~Brussel, H., and Nuttin, M.
\newblock Pruning and regularization in reservoir computing.
\newblock \emph{Neurocomputing}, 72\penalty0 (7):\penalty0 1534--1546, March 2009.
\newblock ISSN 0925-2312.
\newblock \doi{10.1016/j.neucom.2008.12.020}.
\newblock URL \url{https://www.sciencedirect.com/science/article/pii/S0925231209000186}.

\bibitem[Eguiluz et~al.(2005)Eguiluz, Chialvo, Cecchi, Baliki, and Apkarian]{eguiluz2005scale}
Eguiluz, V.~M., Chialvo, D.~R., Cecchi, G.~A., Baliki, M., and Apkarian, A.~V.
\newblock Scale-free brain functional networks.
\newblock \emph{Physical review letters}, 94\penalty0 (1):\penalty0 018102, 2005.

\bibitem[Eisenmann et~al.(2023)Eisenmann, Monfared, G{\"o}ring, and Durstewitz]{eisenmann2023bifurcations}
Eisenmann, L., Monfared, Z., G{\"o}ring, N.~A., and Durstewitz, D.
\newblock Bifurcations and loss jumps in {RNN} training.
\newblock In \emph{Thirty-seventh Conference on Neural Information Processing Systems}, 2023.

\bibitem[Emmert-Streib(2006)]{emmert-streib_influence_2006}
Emmert-Streib, F.
\newblock Influence of the neural network topology on the learning dynamics.
\newblock \emph{Neurocomputing}, 69\penalty0 (10):\penalty0 1179--1182, June 2006.
\newblock ISSN 0925-2312.
\newblock \doi{10.1016/j.neucom.2005.12.070}.
\newblock URL \url{https://www.sciencedirect.com/science/article/pii/S0925231205003966}.

\bibitem[Erd\"{o}s \& R\'{e}nyi(1959)Erd\"{o}s and R\'{e}nyi]{erdos59a}
Erd\"{o}s, P. and R\'{e}nyi, A.
\newblock On random graphs i.
\newblock \emph{Publicationes Mathematicae Debrecen}, 6:\penalty0 290, 1959.

\bibitem[Fagiolo(2007)]{PhysRevE.76.026107}
Fagiolo, G.
\newblock Clustering in complex directed networks.
\newblock \emph{Phys. Rev. E}, 76:\penalty0 026107, 2007.
\newblock \doi{10.1103/PhysRevE.76.026107}.
\newblock URL \url{https://link.aps.org/doi/10.1103/PhysRevE.76.026107}.

\bibitem[Floyd(1962)]{Floyd1962Algorithm9S}
Floyd, R.~W.
\newblock Algorithm 97: Shortest path.
\newblock \emph{Communications of the ACM}, 5:\penalty0 345, 1962.
\newblock URL \url{https://api.semanticscholar.org/CorpusID:2003382}.

\bibitem[Frankle \& Carbin(2019)Frankle and Carbin]{frankle2019lottery}
Frankle, J. and Carbin, M.
\newblock The lottery ticket hypothesis: Finding sparse, trainable neural networks.
\newblock In \emph{International Conference on Learning Representations}, 2019.
\newblock URL \url{https://openreview.net/forum?id=rJl-b3RcF7}.

\bibitem[Frankle et~al.(2020)Frankle, Dziugaite, Roy, and Carbin]{frankle2020linear}
Frankle, J., Dziugaite, G.~K., Roy, D., and Carbin, M.
\newblock Linear mode connectivity and the lottery ticket hypothesis.
\newblock In \emph{International Conference on Machine Learning}, pp.\  3259--3269. PMLR, 2020.

\bibitem[Girish et~al.(2021)Girish, Maiya, Gupta, Chen, Davis, and Shrivastava]{girish_lottery_2021}
Girish, S., Maiya, S.~R., Gupta, K., Chen, H., Davis, L., and Shrivastava, A.
\newblock The {Lottery} {Ticket} {Hypothesis} for {Object} {Recognition}.
\newblock In \emph{2021 {IEEE}/{CVF} {Conference} on {Computer} {Vision} and {Pattern} {Recognition} ({CVPR})}, pp.\  762--771, Nashville, TN, USA, June 2021. IEEE.
\newblock ISBN 978-1-66544-509-2.
\newblock \doi{10.1109/CVPR46437.2021.00082}.
\newblock URL \url{https://ieeexplore.ieee.org/document/9578168/}.

\bibitem[Glorot \& Bengio(2010)Glorot and Bengio]{glorot2010understanding}
Glorot, X. and Bengio, Y.
\newblock Understanding the difficulty of training deep feedforward neural networks.
\newblock In \emph{Proceedings of the thirteenth international conference on artificial intelligence and statistics}, pp.\  249--256. JMLR Workshop and Conference Proceedings, 2010.

\bibitem[Han et~al.(2015{\natexlab{a}})Han, Mao, and Dally]{han2015deep}
Han, S., Mao, H., and Dally, W.~J.
\newblock Deep compression: Compressing deep neural networks with pruning, trained quantization and huffman coding.
\newblock \emph{arXiv preprint arXiv:1510.00149}, 2015{\natexlab{a}}.

\bibitem[Han et~al.(2015{\natexlab{b}})Han, Pool, Tran, and Dally]{NIPS2015_ae0eb3ee}
Han, S., Pool, J., Tran, J., and Dally, W.
\newblock Learning both weights and connections for efficient neural network.
\newblock In Cortes, C., Lawrence, N., Lee, D., Sugiyama, M., and Garnett, R. (eds.), \emph{Advances in Neural Information Processing Systems}, volume~28. Curran Associates, Inc., 2015{\natexlab{b}}.
\newblock URL \url{https://proceedings.neurips.cc/paper_files/paper/2015/file/ae0eb3eed39d2bcef4622b2499a05fe6-Paper.pdf}.

\bibitem[Han et~al.(2015{\natexlab{c}})Han, Pool, Tran, and Dally]{han2015learning}
Han, S., Pool, J., Tran, J., and Dally, W.
\newblock Learning both weights and connections for efficient neural network.
\newblock \emph{Advances in neural information processing systems}, 28, 2015{\natexlab{c}}.

\bibitem[Han et~al.(2022)Han, Zhao, and Small]{han_tighter_2022}
Han, X., Zhao, Y., and Small, M.
\newblock A tighter generalization bound for reservoir computing.
\newblock \emph{Chaos: An Interdisciplinary Journal of Nonlinear Science}, 32\penalty0 (4):\penalty0 043115, April 2022.
\newblock ISSN 1054-1500.
\newblock \doi{10.1063/5.0082258}.
\newblock URL \url{https://doi.org/10.1063/5.0082258}.

\bibitem[Hassibi \& Stork(1992)Hassibi and Stork]{NIPS1992_303ed4c6}
Hassibi, B. and Stork, D.
\newblock Second order derivatives for network pruning: Optimal brain surgeon.
\newblock In Hanson, S., Cowan, J., and Giles, C. (eds.), \emph{Advances in Neural Information Processing Systems}, volume~5. Morgan-Kaufmann, 1992.
\newblock URL \url{https://proceedings.neurips.cc/paper_files/paper/1992/file/303ed4c69846ab36c2904d3ba8573050-Paper.pdf}.

\bibitem[He \& Xiao(2023)He and Xiao]{He_2023}
He, Y. and Xiao, L.
\newblock Structured pruning for deep convolutional neural networks: A survey.
\newblock \emph{IEEE Transactions on Pattern Analysis and Machine Intelligence}, pp.\  1–20, 2023.
\newblock ISSN 1939-3539.
\newblock \doi{10.1109/tpami.2023.3334614}.
\newblock URL \url{http://dx.doi.org/10.1109/TPAMI.2023.3334614}.

\bibitem[Hess et~al.(2023)Hess, Monfared, Brenner, and Durstewitz]{hess2023generalized}
Hess, F., Monfared, Z., Brenner, M., and Durstewitz, D.
\newblock Generalized teacher forcing for learning chaotic dynamics.
\newblock In Krause, A., Brunskill, E., Cho, K., Engelhardt, B., Sabato, S., and Scarlett, J. (eds.), \emph{Proceedings of the 40th International Conference on Machine Learning}, volume 202 of \emph{Proceedings of Machine Learning Research}, pp.\  13017--13049. PMLR, 23--29 Jul 2023.
\newblock URL \url{https://proceedings.mlr.press/v202/hess23a.html}.

\bibitem[Hochreiter \& Schmidhuber(1997)Hochreiter and Schmidhuber]{lstm}
Hochreiter, S. and Schmidhuber, J.
\newblock {Long Short-Term Memory}.
\newblock \emph{Neural Computation}, 9\penalty0 (8):\penalty0 1735--1780, 1997.
\newblock ISSN 0899-7667.
\newblock \doi{10.1162/neco.1997.9.8.1735}.
\newblock URL \url{https://doi.org/10.1162/neco.1997.9.8.1735}.

\bibitem[Jaeger \& Haas(2004)Jaeger and Haas]{jaeger_harnessing_2004}
Jaeger, H. and Haas, H.
\newblock Harnessing {Nonlinearity}: {Predicting} {Chaotic} {Systems} and {Saving} {Energy} in {Wireless} {Communication}.
\newblock \emph{Science}, 304\penalty0 (5667):\penalty0 78--80, April 2004.
\newblock \doi{10.1126/science.1091277}.
\newblock URL \url{https://www.science.org/doi/10.1126/science.1091277}.
\newblock Publisher: American Association for the Advancement of Science.

\bibitem[Jiang \& Claramunt(2004)Jiang and Claramunt]{jiang2004topological}
Jiang, B. and Claramunt, C.
\newblock Topological analysis of urban street networks.
\newblock \emph{Environment and Planning B: Planning and design}, 31\penalty0 (1):\penalty0 151--162, 2004.

\bibitem[Junior et~al.(2020)Junior, Stelzer, and Zhao]{junior_clustered_2020}
Junior, L.~O., Stelzer, F., and Zhao, L.
\newblock Clustered {Echo} {State} {Networks} for {Signal} {Observation} and {Frequency} {Filtering}.
\newblock In \emph{Anais do {Symposium} on {Knowledge} {Discovery}, {Mining} and {Learning} ({KDMiLe})}, pp.\  25--32. SBC, October 2020.
\newblock \doi{10.5753/kdmile.2020.11955}.
\newblock URL \url{https://sol.sbc.org.br/index.php/kdmile/article/view/11955}.
\newblock ISSN: 2763-8944.

\bibitem[Karlsson \& Svanström(2019)Karlsson and Svanström]{karlsson_modelling_2019}
Karlsson, D. and Svanström, O.
\newblock Modelling {Dynamical} {Systems} {Using} {Neural} {Ordinary} {Differential} {Equations}, 2019.

\bibitem[Kingma \& Ba(2015)Kingma and Ba]{kingma2014adam}
Kingma, D.~P. and Ba, J.
\newblock Adam: {A} {Method} for {Stochastic} {Optimization}.
\newblock In \emph{Proceedings of the 3rd {International} {Conference} on {Learning} {Representations}}, 2015.
\newblock URL \url{http://arxiv.org/abs/1412.6980}.

\bibitem[Kleinberg(2000)]{kleinberg2000small}
Kleinberg, J.
\newblock The small-world phenomenon: An algorithmic perspective.
\newblock In \emph{Proceedings of the thirty-second annual ACM symposium on Theory of computing}, pp.\  163--170, 2000.

\bibitem[Ko et~al.(2023)Ko, Koh, Park, and Jhe]{ko_homotopy-based_2023}
Ko, J.-H., Koh, H., Park, N., and Jhe, W.
\newblock Homotopy-based training of neuralodes for accurate dynamics discovery.
\newblock \emph{Advances in Neural Information Processing Systems}, 36:\penalty0 64725--64752, 2023.

\bibitem[Koppe et~al.(2019)Koppe, Toutounji, Kirsch, Lis, and Durstewitz]{koppe_fmri_2019}
Koppe, G., Toutounji, H., Kirsch, P., Lis, S., and Durstewitz, D.
\newblock Identifying nonlinear dynamical systems via generative recurrent neural networks with applications to fmri.
\newblock \emph{PLOS Computational Biology}, 15\penalty0 (8):\penalty0 1--35, 2019.
\newblock \doi{10.1371/journal.pcbi.1007263}.
\newblock URL \url{https://doi.org/10.1371/journal.pcbi.1007263}.

\bibitem[Kramer et~al.(2022)Kramer, Bommer, Tombolini, Koppe, and Durstewitz]{kramer2021reconstructing}
Kramer, D., Bommer, P.~L., Tombolini, C., Koppe, G., and Durstewitz, D.
\newblock Reconstructing nonlinear dynamical systems from multi-modal time series.
\newblock In Chaudhuri, K., Jegelka, S., Song, L., Szepesvari, C., Niu, G., and Sabato, S. (eds.), \emph{Proceedings of the 39th International Conference on Machine Learning}, volume 162 of \emph{Proceedings of Machine Learning Research}, pp.\  11613--11633. PMLR, 17--23 Jul 2022.
\newblock URL \url{https://proceedings.mlr.press/v162/kramer22a.html}.

\bibitem[LeCun et~al.(1989)LeCun, Denker, and Solla]{NIPS1989_6c9882bb}
LeCun, Y., Denker, J., and Solla, S.
\newblock Optimal brain damage.
\newblock In Touretzky, D. (ed.), \emph{Advances in Neural Information Processing Systems}, volume~2. Morgan-Kaufmann, 1989.
\newblock URL \url{https://proceedings.neurips.cc/paper_files/paper/1989/file/6c9882bbac1c7093bd25041881277658-Paper.pdf}.

\bibitem[Li et~al.(2020)Li, Li, and Wang]{li_echo_2020}
Li, F., Li, Y., and Wang, X.
\newblock Echo {State} {Network} with {Hub} {Property}.
\newblock In Deng, Z. (ed.), \emph{Proceedings of 2019 {Chinese} {Intelligent} {Automation} {Conference}}, Lecture {Notes} in {Electrical} {Engineering}, pp.\  537--544, Singapore, 2020. Springer.
\newblock ISBN 978-981-329-050-1.
\newblock \doi{10.1007/978-981-32-9050-1_61}.

\bibitem[Liu et~al.(2020)Liu, Jiang, He, Chen, Liu, Gao, and Han]{Liu2020On}
Liu, L., Jiang, H., He, P., Chen, W., Liu, X., Gao, J., and Han, J.
\newblock On the variance of the adaptive learning rate and beyond.
\newblock In \emph{International Conference on Learning Representations}, 2020.
\newblock URL \url{https://openreview.net/forum?id=rkgz2aEKDr}.

\bibitem[Liu et~al.(2021)Liu, Mocanu, Pei, and Pechenizkiy]{liu2021selfish}
Liu, S., Mocanu, D.~C., Pei, Y., and Pechenizkiy, M.
\newblock Selfish sparse rnn training.
\newblock In \emph{International Conference on Machine Learning}, pp.\  6893--6904. PMLR, 2021.

\bibitem[Liu et~al.(2018)Liu, Sun, Zhou, Huang, and Darrell]{liu2018rethinking}
Liu, Z., Sun, M., Zhou, T., Huang, G., and Darrell, T.
\newblock Rethinking the value of network pruning.
\newblock \emph{arXiv preprint arXiv:1810.05270}, 2018.

\bibitem[Lorenz(1963)]{DeterministicNonperiodicFlow}
Lorenz, E.~N.
\newblock Deterministic nonperiodic flow.
\newblock \emph{Journal of Atmospheric Sciences}, 20\penalty0 (2):\penalty0 130 -- 141, 1963.
\newblock \doi{https://doi.org/10.1175/1520-0469(1963)020<0130:DNF>2.0.CO;2}.
\newblock URL \url{https://journals.ametsoc.org/view/journals/atsc/20/2/1520-0469_1963_020_0130_dnf_2_0_co_2.xml}.

\bibitem[Lorenz(1996)]{lorenz96}
Lorenz, E.~N.
\newblock Predictability: A problem partly solved.
\newblock In \emph{Proc. Seminar on predictability}, volume~1. Reading, 1996.

\bibitem[Malach et~al.(2020{\natexlab{a}})Malach, Yehudai, Shalev-Schwartz, and Shamir]{malach2020proving}
Malach, E., Yehudai, G., Shalev-Schwartz, S., and Shamir, O.
\newblock Proving the lottery ticket hypothesis: Pruning is all you need.
\newblock In \emph{International Conference on Machine Learning}, pp.\  6682--6691. PMLR, 2020{\natexlab{a}}.

\bibitem[Malach et~al.(2020{\natexlab{b}})Malach, Yehudai, Shalev-Schwartz, and Shamir]{malach_proving_2020}
Malach, E., Yehudai, G., Shalev-Schwartz, S., and Shamir, O.
\newblock Proving the {Lottery} {Ticket} {Hypothesis}: {Pruning} is {All} {You} {Need}.
\newblock In \emph{Proceedings of the 37th {International} {Conference} on {Machine} {Learning}}, pp.\  6682--6691. PMLR, November 2020{\natexlab{b}}.
\newblock URL \url{https://proceedings.mlr.press/v119/malach20a.html}.
\newblock ISSN: 2640-3498.

\bibitem[Messenger \& Bortz(2021)Messenger and Bortz]{messenger_weak_2021}
Messenger, D.~A. and Bortz, D.~M.
\newblock Weak {SINDy}: {Galerkin}-{Based} {Data}-{Driven} {Model} {Selection}.
\newblock \emph{Multiscale Modeling \& Simulation}, 19\penalty0 (3):\penalty0 1474--1497, January 2021.
\newblock ISSN 1540-3459.
\newblock \doi{10.1137/20M1343166}.
\newblock URL \url{https://epubs.siam.org/doi/10.1137/20M1343166}.
\newblock Publisher: Society for Industrial and Applied Mathematics.

\bibitem[Mikhaeil et~al.(2022)Mikhaeil, Monfared, and Durstewitz]{mikhaeil2022difficulty}
Mikhaeil, J., Monfared, Z., and Durstewitz, D.
\newblock On the difficulty of learning chaotic dynamics with rnns.
\newblock \emph{Advances in Neural Information Processing Systems}, 35:\penalty0 11297--11312, 2022.

\bibitem[Monfared \& Durstewitz(2020)Monfared and Durstewitz]{pmlr-v119-monfared20a}
Monfared, Z. and Durstewitz, D.
\newblock Transformation of {R}e{LU}-based recurrent neural networks from discrete-time to continuous-time.
\newblock In III, H.~D. and Singh, A. (eds.), \emph{Proceedings of the 37th International Conference on Machine Learning}, volume 119 of \emph{Proceedings of Machine Learning Research}, pp.\  6999--7009. PMLR, 2020.
\newblock URL \url{https://proceedings.mlr.press/v119/monfared20a.html}.

\bibitem[Muldoon et~al.(2016)Muldoon, Bridgeford, and Bassett]{muldoon2016small}
Muldoon, S.~F., Bridgeford, E.~W., and Bassett, D.~S.
\newblock Small-world propensity and weighted brain networks.
\newblock \emph{Scientific reports}, 6\penalty0 (1):\penalty0 22057, 2016.

\bibitem[NEAL(2017)]{neal_2017}
NEAL, Z.~P.
\newblock How small is it? comparing indices of small worldliness.
\newblock \emph{Network Science}, 5\penalty0 (1):\penalty0 30–44, 2017.
\newblock \doi{10.1017/nws.2017.5}.

\bibitem[Orseau et~al.(2020)Orseau, Hutter, and Rivasplata]{orseau_logarithmic_2020}
Orseau, L., Hutter, M., and Rivasplata, O.
\newblock Logarithmic {Pruning} is {All} {You} {Need}.
\newblock In \emph{Advances in {Neural} {Information} {Processing} {Systems}}, volume~33, pp.\  2925--2934. Curran Associates, Inc., 2020.

\bibitem[Pajevic \& Plenz(2012)Pajevic and Plenz]{pajevic2012organization}
Pajevic, S. and Plenz, D.
\newblock The organization of strong links in complex networks.
\newblock \emph{Nature Physics}, 8\penalty0 (5):\penalty0 429--436, 2012.

\bibitem[Pathak et~al.(2018)Pathak, Hunt, Girvan, Lu, and Ott]{pathak2018model}
Pathak, J., Hunt, B., Girvan, M., Lu, Z., and Ott, E.
\newblock Model-free prediction of large spatiotemporally chaotic systems from data: A reservoir computing approach.
\newblock \emph{Physical review letters}, 120\penalty0 (2):\penalty0 024102, 2018.

\bibitem[Platt et~al.(2021)Platt, Wong, Clark, Penny, and Abarbanel]{platt2021robust}
Platt, J.~A., Wong, A., Clark, R., Penny, S.~G., and Abarbanel, H.~D.
\newblock Robust forecasting using predictive generalized synchronization in reservoir computing.
\newblock \emph{Chaos: An Interdisciplinary Journal of Nonlinear Science}, 31\penalty0 (12), 2021.

\bibitem[Platt et~al.(2023)Platt, Penny, Smith, Chen, and Abarbanel]{platt2023constraining}
Platt, J.~A., Penny, S.~G., Smith, T.~A., Chen, T.-C., and Abarbanel, H.~D.
\newblock Constraining chaos: Enforcing dynamical invariants in the training of reservoir computers.
\newblock \emph{Chaos: An Interdisciplinary Journal of Nonlinear Science}, 33\penalty0 (10), 2023.

\bibitem[Ravasz \& Barabási(2003)Ravasz and Barabási]{ravasz_hierarchical_2003}
Ravasz, E. and Barabási, A.-L.
\newblock Hierarchical organization in complex networks.
\newblock \emph{Physical Review E}, 67\penalty0 (2):\penalty0 026112, February 2003.
\newblock \doi{10.1103/PhysRevE.67.026112}.
\newblock URL \url{https://link.aps.org/doi/10.1103/PhysRevE.67.026112}.
\newblock Publisher: American Physical Society.

\bibitem[Reiss et~al.(2019)Reiss, Indlekofer, Schmidt, and Van~Laerhoven]{reiss2019deep}
Reiss, A., Indlekofer, I., Schmidt, P., and Van~Laerhoven, K.
\newblock Deep ppg: Large-scale heart rate estimation with convolutional neural networks.
\newblock \emph{Sensors}, 19\penalty0 (14):\penalty0 3079, 2019.

\bibitem[Rubinov et~al.(2009)Rubinov, Knock, Stam, Micheloyannis, Harris, Williams, and Breakspear]{rubinov2009small}
Rubinov, M., Knock, S.~A., Stam, C.~J., Micheloyannis, S., Harris, A.~W., Williams, L.~M., and Breakspear, M.
\newblock Small-world properties of nonlinear brain activity in schizophrenia.
\newblock \emph{Human brain mapping}, 30\penalty0 (2):\penalty0 403--416, 2009.

\bibitem[Rumelhart et~al.(1986)Rumelhart, Hinton, and Williams]{rumelhart1986learning}
Rumelhart, D.~E., Hinton, G.~E., and Williams, R.~J.
\newblock Learning representations by back-propagating errors.
\newblock \emph{Nature}, 323\penalty0 (6088):\penalty0 533--536, 1986.

\bibitem[Rössler(1976)]{ROSSLER1976397}
Rössler, O.
\newblock An equation for continuous chaos.
\newblock \emph{Physics Letters A}, 57\penalty0 (5):\penalty0 397--398, 1976.
\newblock ISSN 0375-9601.
\newblock \doi{https://doi.org/10.1016/0375-9601(76)90101-8}.
\newblock URL \url{https://www.sciencedirect.com/science/article/pii/0375960176901018}.

\bibitem[Shandilya \& Timme(2011)Shandilya and Timme]{shandilya_inferring_2011}
Shandilya, S.~G. and Timme, M.
\newblock Inferring network topology from complex dynamics.
\newblock \emph{New Journal of Physics}, 13\penalty0 (1):\penalty0 013004, January 2011.
\newblock ISSN 1367-2630.
\newblock \doi{10.1088/1367-2630/13/1/013004}.
\newblock URL \url{https://dx.doi.org/10.1088/1367-2630/13/1/013004}.

\bibitem[Sreenivasan et~al.(2022)Sreenivasan, Sohn, Yang, Grinde, Nagle, Wang, Xing, Lee, and Papailiopoulos]{sreenivasan_rare_2022}
Sreenivasan, K., Sohn, J.-y., Yang, L., Grinde, M., Nagle, A., Wang, H., Xing, E., Lee, K., and Papailiopoulos, D.
\newblock Rare gems: Finding lottery tickets at initialization.
\newblock \emph{Advances in neural information processing systems}, 35:\penalty0 14529--14540, 2022.

\bibitem[Stoer \& Bulirsch(2002)Stoer and Bulirsch]{bulirsch2002}
Stoer, J. and Bulirsch, R.
\newblock \emph{Introduction to numerical analysis}.
\newblock Texts in applied mathematics. Springer, 2002.
\newblock ISBN 9780387954523.

\bibitem[Talathi \& Vartak(2016)Talathi and Vartak]{talathi2016improving}
Talathi, S.~S. and Vartak, A.
\newblock Improving performance of recurrent neural network with relu nonlinearity.
\newblock In \emph{Proceedings of the 4th {International} {Conference} on {Learning} {Representations}}, 2016.
\newblock URL \url{http://arxiv.org/abs/1511.03771}.

\bibitem[Tziperman et~al.(1997)Tziperman, Scher, Zebiak, and Cane]{tziperman-97}
Tziperman, E., Scher, H., Zebiak, S.~E., and Cane, M.~A.
\newblock Controlling spatiotemporal chaos in a realistic el ni\~no prediction model.
\newblock \emph{Phys. Rev. Lett.}, 79:\penalty0 1034--1037, Aug 1997.
\newblock \doi{10.1103/PhysRevLett.79.1034}.
\newblock URL \url{https://link.aps.org/doi/10.1103/PhysRevLett.79.1034}.

\bibitem[van~den Heuvel et~al.(2008)van~den Heuvel, Stam, Boersma, and Pol]{van2008small}
van~den Heuvel, M.~P., Stam, C.~J., Boersma, M., and Pol, H.~H.
\newblock Small-world and scale-free organization of voxel-based resting-state functional connectivity in the human brain.
\newblock \emph{Neuroimage}, 43\penalty0 (3):\penalty0 528--539, 2008.

\bibitem[Vlachas et~al.(2018)Vlachas, Byeon, Wan, Sapsis, and Koumoutsakos]{vlachas2018data}
Vlachas, P.~R., Byeon, W., Wan, Z.~Y., Sapsis, T.~P., and Koumoutsakos, P.
\newblock Data-driven forecasting of high-dimensional chaotic systems with long short-term memory networks.
\newblock \emph{Proceedings of the Royal Society A: Mathematical, Physical and Engineering Sciences}, 474\penalty0 (2213):\penalty0 20170844, 2018.

\bibitem[Voss et~al.(2004)Voss, Timmer, and Kurths]{voss_nonlinear_2004}
Voss, H.~U., Timmer, J., and Kurths, J.
\newblock Nonlinear dynamical system identification from uncertain and indirect measurements.
\newblock \emph{International Journal of Bifurcation and Chaos}, 14\penalty0 (06):\penalty0 1905--1933, June 2004.
\newblock ISSN 0218-1274.
\newblock \doi{10.1142/S0218127404010345}.
\newblock URL \url{https://www.worldscientific.com/doi/abs/10.1142/S0218127404010345}.
\newblock Publisher: World Scientific Publishing Co.

\bibitem[Wang et~al.(2016)Wang, Wu, Feng, Lu, and Lü]{wang_topology_2016}
Wang, Y., Wu, X., Feng, H., Lu, J., and Lü, J.
\newblock Topology inference of uncertain complex dynamical networks and its applications in hidden nodes detection.
\newblock \emph{Science China Technological Sciences}, 59\penalty0 (8):\penalty0 1232--1243, August 2016.
\newblock ISSN 1869-1900.
\newblock \doi{10.1007/s11431-016-6050-1}.
\newblock URL \url{https://doi.org/10.1007/s11431-016-6050-1}.

\bibitem[Warshall(1962)]{Warshall1962ATO}
Warshall, S.
\newblock A theorem on boolean matrices.
\newblock \emph{J. ACM}, 9:\penalty0 11--12, 1962.
\newblock URL \url{https://api.semanticscholar.org/CorpusID:33763989}.

\bibitem[Watts \& Strogatz(1998)Watts and Strogatz]{watts_collective_1998}
Watts, D.~J. and Strogatz, S.~H.
\newblock Collective dynamics of ‘small-world’ networks.
\newblock \emph{Nature}, 393\penalty0 (6684):\penalty0 440--442, 1998.
\newblock \doi{10.1038/30918}.

\bibitem[Williams \& Zipser(1989)Williams and Zipser]{williams1989learning}
Williams, R.~J. and Zipser, D.
\newblock A learning algorithm for continually running fully recurrent neural networks.
\newblock \emph{Neural computation}, 1\penalty0 (2):\penalty0 270--280, 1989.

\bibitem[Wood(2010)]{wood_statistical_2010}
Wood, S.~N.
\newblock Statistical inference for noisy nonlinear ecological dynamic systems.
\newblock \emph{Nature}, 466\penalty0 (7310):\penalty0 1102--1104, August 2010.
\newblock ISSN 1476-4687.
\newblock \doi{10.1038/nature09319}.
\newblock URL \url{https://www.nature.com/articles/nature09319}.
\newblock Number: 7310 Publisher: Nature Publishing Group.

\bibitem[Yin \& Meng(2012)Yin and Meng]{yin_self-organizing_2012}
Yin, J. and Meng, Y.
\newblock Self-organizing reservior computing with dynamically regulated cortical neural networks.
\newblock In \emph{The 2012 {International} {Joint} {Conference} on {Neural} {Networks} ({IJCNN})}, pp.\  1--7, June 2012.
\newblock \doi{10.1109/IJCNN.2012.6252772}.
\newblock URL \url{https://ieeexplore.ieee.org/document/6252772}.
\newblock ISSN: 2161-4407.

\bibitem[You et~al.(2020)You, Leskovec, He, and Xie]{you2020graph}
You, J., Leskovec, J., He, K., and Xie, S.
\newblock Graph structure of neural networks.
\newblock In \emph{International Conference on Machine Learning}, pp.\  10881--10891. PMLR, 2020.

\bibitem[Yu et~al.(2019)Yu, Edunov, Tian, and Morcos]{yu2019playing}
Yu, H., Edunov, S., Tian, Y., and Morcos, A.~S.
\newblock Playing the lottery with rewards and multiple languages: lottery tickets in rl and nlp.
\newblock \emph{arXiv preprint arXiv:1906.02768}, 2019.

\bibitem[Zhang et~al.(2009{\natexlab{a}})Zhang, Liu, and Wang]{zhang2009controlling}
Zhang, H., Liu, D., and Wang, Z.
\newblock \emph{Controlling chaos: suppression, synchronization and chaotification}.
\newblock Springer Science \& Business Media, 2009{\natexlab{a}}.

\bibitem[Zhang et~al.(2021)Zhang, Wang, Liu, Chen, and Xiong]{zhang_why_2021}
Zhang, S., Wang, M., Liu, S., Chen, P.-Y., and Xiong, J.
\newblock Why {Lottery} {Ticket} {Wins}? {A} {Theoretical} {Perspective} of {Sample} {Complexity} on {Sparse} {Neural} {Networks}.
\newblock In \emph{Advances in {Neural} {Information} {Processing} {Systems}}, volume~34, pp.\  2707--2720. Curran Associates, Inc., 2021.

\bibitem[Zhang et~al.(2009{\natexlab{b}})Zhang, Lin, Gao, Zhou, Guan, and Li]{zhang_trapping_2009}
Zhang, Z., Lin, Y., Gao, S., Zhou, S., Guan, J., and Li, M.
\newblock Trapping in scale-free networks with hierarchical organization of modularity.
\newblock \emph{Physical Review. E, Statistical, Nonlinear, and Soft Matter Physics}, 80\penalty0 (5 Pt 1):\penalty0 051120, November 2009{\natexlab{b}}.
\newblock ISSN 1550-2376.
\newblock \doi{10.1103/PhysRevE.80.051120}.

\end{thebibliography}
\bibliographystyle{icml2024}

\newpage
\appendix
\onecolumn

\setcounter{figure}{0} 
\renewcommand{\thefigure}{A\arabic{figure}}
\renewcommand{\theHfigure}{A\arabic{figure}}

\setcounter{algorithm}{0} 
\renewcommand{\thealgorithm}{A\arabic{algorithm}}
\renewcommand{\theHalgorithm}{A\arabic{algorithm}}

\section{Appendix}

\subsection{Methodological Details} \label{ch:Methodological_details}

\subsubsection{Mean-centered PLRNN} \label{sect:mean_centred_plrnn} 
Layer normalization is often beneficial for training RNNs \cite{ba2016layer} and has been modified for PLRNNs in order to retain their piecewise linear structure \cite{brenner2022tractable}. \citet{brenner2022tractable} observed that mean-centering before applying the nonlinearity at each time step is already sufficient to obtain the usual performance boosts, implemented as 
\begin{equation}
    \mathcal{M}(\bm{z}_{t-1})=\bm{z}_{t-1}-\mu_{t-1}=\bm{z}_{t-1}-\bm{1}\frac{1}{M}\sum_{i=1}^M\bm{z}_{t-1,i}\;,
\end{equation}
where $\bm{1}\in\mathbb{R}^M$ is a vector of ones. Since mean-centering is linear, it can be rewritten as a matrix multiplication with the latent state vector, 
\begin{equation}
    \mathcal{M}(\bm{z}_{t-1})=\bm{M}\bm{z}_{t-1}=
    \frac{1}{M}\begin{pmatrix}
    M-1 & -1 & \hdots & -1\\
    -1 & M-1 & \hdots & -1 \\
    \vdots & \vdots & \ddots & \vdots \\
    -1 & -1 & \hdots & M-1
    \end{pmatrix} \bm{z}_{t-1}\;.
\end{equation}

\subsubsection{BPTT + identity-TF}
Training RNNs via BPTT runs into the exploding \& vanishing gradient problem, which is aggravated when training on chaotic systems \cite{mikhaeil2022difficulty}. \citet{mikhaeil2022difficulty,brenner2022tractable} suggested STF as a remedy, which in the case of a direct (identity) mapping from a subset of latent states to the observed time series values takes a particularly simple form. Let $\bm{X}=\{\bm{x}_t\}$ be the observed time series, and $\bm{Z}=\{\bm{z}_t\}$ the RNN latent states. We then create a control series $\Tilde{\bm{Z}}$ by inverting the observation model, which in identity-TF simply comes down to 
\begin{equation}
    \Tilde{z}_{t,k}=
    \begin{cases}
        x_{t,k}, & \text{for $k\leq N$} \\
        z_{t,k}, & \text{for $k> N$}
    \end{cases}\;,
\end{equation}
i.e. the $\Tilde{\bm{z}}_t$ are just the latent states with the first $N$ components replaced by the actual observations. 
In STF, the latent states $\bm{z}_t$ are replaced by the control states $\Tilde{\bm{z}}_t$ sparsely at times $\mathcal{T}=\{n\tau+1\}$, $n\in\mathbb{N}$, separated by an interval $\tau$:
\begin{equation}
    \bm{z}_{t}=
    \begin{cases}
        PLRNN(\Tilde{\bm{z}}_{t-1}), & \text{if $t\in\mathcal{T}$} \\
        PLRNN(\bm{z}_{t-1}), & \text{else}
    \end{cases}\;,
\end{equation}
where this forcing is always applied \textit{after} calculating the loss. To allow the system to capture relevant time scales while avoiding divergence, ideally the forcing interval $\tau$ is chosen according to the predictability time based on the system's maximum Lyapunov exponent \cite{mikhaeil2022difficulty}, but here we simply determined the optimal $\tau$ by grid search as in \cite{brenner2022tractable}. Importantly, STF is only applied during model training and not at test time. 

\subsubsection{Training Protocol}
Given a time series $\{\bm{x}_{1:T}\}$ from a DS, we train the model using BPTT + identity-STF. For each training epoch we sample several subsequences of length $\Tilde{T}$, $\Tilde{x}^{(p)}_{1:\Tilde{T}}=x_{t_p:t_p+\Tilde{T}}$ where $t_p\in[1,T-\Tilde{T}]$ is chosen randomly. These subsequences $\left\{ \Tilde{\bm{x}}^{(p)}_{1:\Tilde{T}} \right\}^S_{p=1}$ are then arranged into a batch of size $S$. On each sequence, the PLRNN is initialized with the first forcing signal $\Tilde{\bm{z}}_1^{(p)}$, and from there forward-iterated in time, yielding predictions $\left\{\hat{\Tilde{\bm{x}}}_{2:\Tilde{T}}^{(p)}\right\} = \left\{\bm{z}_{2:\Tilde{T},1:N}^{(p)}\right\}$ using Eqn. \ref{eq:PLRNN}. The loss is then computed as the MSE between predicted and ground truth time series
\begin{equation}
    \mathcal{L}_{\text{MSE}}\left( \left\{\Tilde{\bm{x}}_{2:\Tilde{T}}^{(p)}\right\}, \left\{\hat{\Tilde{\bm{x}}}_{2:\Tilde{T}}^{(p)}\right\} \right) = \frac{1}{S(\Tilde{T}-1)}\sum_{p=1}^S\sum_{t=2}^{\Tilde{T}}\left\lVert\Tilde{\bm{x}}_t^{(p)}-\hat{\Tilde{\bm{x}}}_t^{(p)}\right\rVert_2^2\;.
\end{equation}
We took rectified adaptive moment estimation (RADAM) \cite{Liu2020On} as the optimizer, using $L=50$ batches of size $S=16$ in each epoch. We chose $M=\{50,100,100,50,100\}$, $\tau=\{16,10,5,8,8\}$, $T=\{200,50,50,300,200\}$, $\eta_{\text{start}}=\{10^{-2},10^{-3},10^{-3},5\cdot10^{-3},5\cdot10^{-3}\}$, and $epochs=\{2000,3000,4000,3000,3000\}$ for the \{Lorenz-63, ECG, Bursting Neuron, Rössler, Lorenz-96\}, respectively, and $\eta_{\text{end}}=10^{-5}$ for all settings. Parameters in $\bm{W}$ were initialized using a Gaussian initialization with $\sigma=0.01$, $\bm{h}$ simply as a vector of zeros, and $\bm{A}$ as the diagonal of a normalized positive-definite random matrix \cite{brenner2022tractable,talathi2016improving}. Across all training epochs of a given run, we consistently (for all comparisons and protocols) selected the model with the lowest $D_{\text{stsp}}$. Failed trainings (yielding NaN entries) were discarded. Failures in training were indeed much more common for the random and magnitude-based pruning protocols, further reinforcing our points about the importance of network graph topology.

The vanilla RNN used is given by 
\begin{equation}\label{eq:vanillaRNN}
    \bm{z}_t=\phi(\bm{W}\bm{z}_{t-1}+\bm{C}\bm{x}_t+\bm{b})\;.
\end{equation}
The LSTM architecture is defined by \cite{lstm}
\begin{equation}\label{eq:LSTM}
    \begin{gathered}
        \bm{f}_t=\sigma(\bm{W}_f\bm{z}_{t-1}+\bm{C}_f\bm{x}_{t-1}+\bm{b}_f)\;,\\
        \bm{i}_t=\sigma(\bm{W}_i\bm{z}_{t-1}+\bm{C}_i\bm{x}_{t-1}+\bm{b}_i)\;,\\
        \bm{o}_t=\sigma(\bm{W}_o\bm{z}_{t-1}+\bm{C}_o\bm{x}_{t-1}+\bm{b}_o)\;,\\
        \Tilde{\bm{c}}_t=\tanh(\bm{W}_c\bm{z}_{t-1}+\bm{C}_c\bm{x}_{t-1}+\bm{b}_c)\;,\\
        \bm{c}_t=\bm{f}_t\odot \bm{c}_{t-1}+\bm{i}_t\odot \Tilde{\bm{c}}_t\;,\\
        \bm{z}_t=\bm{o}_t\odot \tanh(\bm{c}_t)\;,
    \end{gathered}
\end{equation}
where $\sigma(\cdot)$ is the standard sigmoid function. For both these architectures, a linear observation model 
\begin{equation}
    \bm{x}_t=\bm{B}\bm{z}_t+\bm{h}
\end{equation}
was used. Models were trained via standard BPTT \cite{rumelhart1986learning} with MSE loss, using the ADAM optimizer \cite{kingma2014adam}. For both architectures batch size was $S=16$, with $L=20$ batches per epoch, $M=50$, $T=200$, $\eta_{\text{start}}=10^{-3}$, $\eta_{\text{end}}=10^{-6}$, and $epochs=500$. All weight parameters were initialized using a Gaussian scheme with $\sigma=0.01$, and biases as vectors of zeros. The $\bm{B}$ matrix of the observation model was initialized using glorot-uniform initialization \cite{glorot2010understanding}. Like for the PLRNN, pruning was applied to all weight matrices $\bm{W}$, $\bm{C}$ (for vanilla RNN), and $\bm{W}_f$, $\bm{W}_i$, $\bm{W}_o$, $\bm{W}_c$, $\bm{C}_f$, $\bm{C}_i$, $\bm{C}_o$, $\bm{C}_c$ (for LSTM). 

\newpage
\subsection{Details on the Benchmark Datasets} \label{Benachmark data}
From all benchmark systems, as detailed below, trajectories of $10^5$ time steps were drawn for training, all dimensions were individually standardized, and Gaussian observation noise was added (Lorenz-63: $5\%$, Lorenz-96: $1\%$, Bursting Neuron: $2\%$, Rössler: $5\%$, ECG: $5\%$). All systems (except for the human ECG data) were numerically integrated using a fourth-order Runge-Kutta scheme \cite{bulirsch2002}. 

\subsubsection{Lorenz-63 System}
\begin{figure}[ht]
    \begin{center}
    \includegraphics[width=0.9\textwidth]{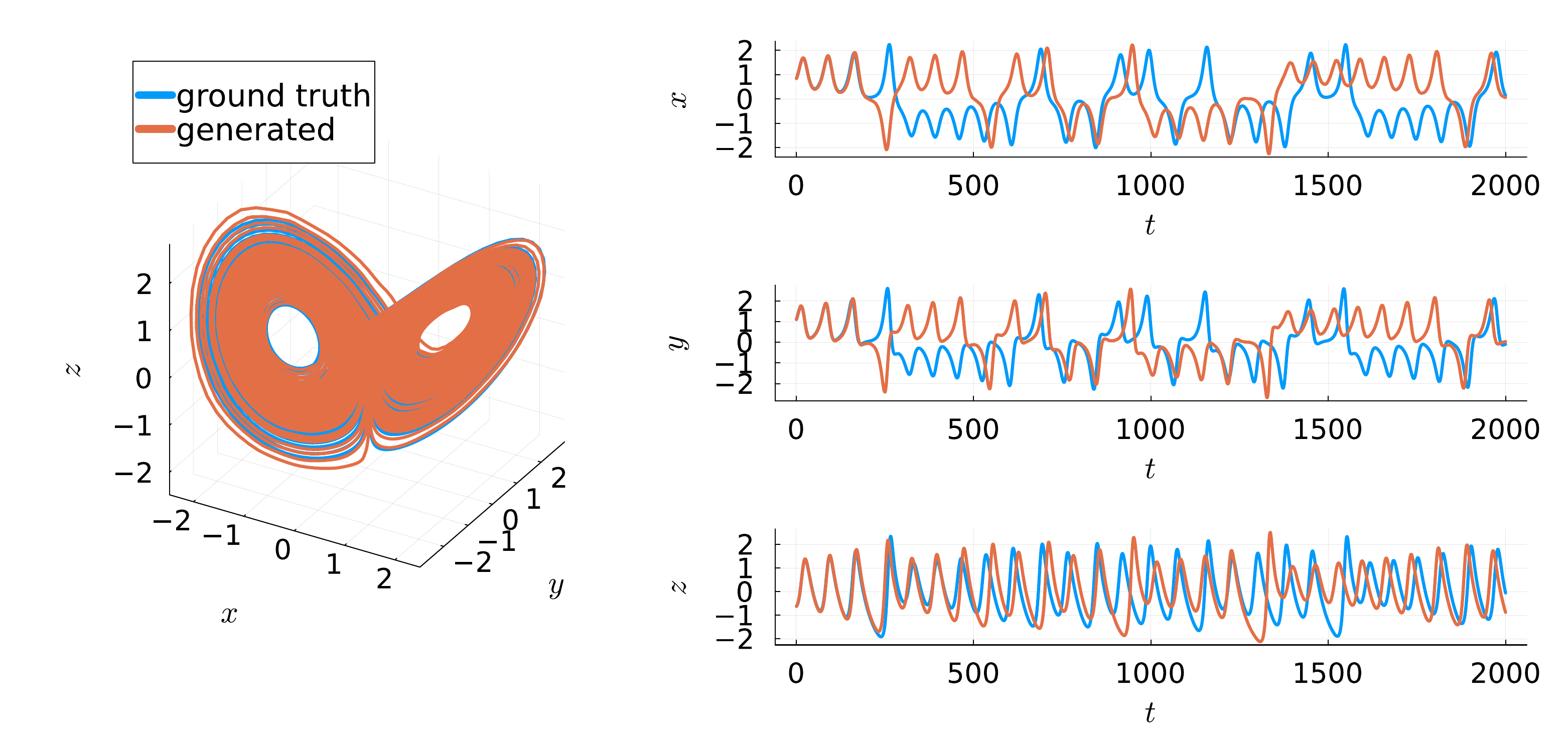}
    \caption{State space (left) and time graphs (right) for Lorenz-63 system with $\sigma=10$, $\beta=\frac{8}{3}$ and $\rho=28$ (blue), and a PLRNN reconstruction (red). Note that despite the essentially perfect reconstruction in state space, the Lorenz system's positive Lyapunov exponent causes the true and reconstructed trajectories to eventually diverge (yet their temporal structure remains the same).
    }
    \label{fig:Lorenz63}
    \end{center}
\end{figure}

The Lorenz-63 system, introduced by Edward Lorenz in 1963 \cite{DeterministicNonperiodicFlow} as a model of atmospheric convection, is given by
\begin{align}
    \frac{\text{d}x}{\text{d}t} & = \sigma(y-x)\\ \nonumber
    \frac{\text{d}y}{\text{d}t} & = x(\rho-z)-y\\ \nonumber
    \frac{\text{d}z}{\text{d}t} & = xy-\beta z,
\end{align}
where $\sigma, \rho, \beta$, are parameters that control the behavior of the system (set to $\sigma=10$, $\beta=\frac{8}{3}$, and $\rho=28$ here, within the chaotic regime). The Lorenz-63 is one of the most popular examples in chaos theory and in the literature on dynamical systems reconstruction. Here we solved this system with integration time step $\Delta t = 0.01$. See Fig. \ref{fig:Lorenz63} for an illustration.
\newpage

\subsubsection{Lorenz-96 System}
\begin{figure}[ht]
    \begin{center}
    \includegraphics[width=0.8\textwidth]{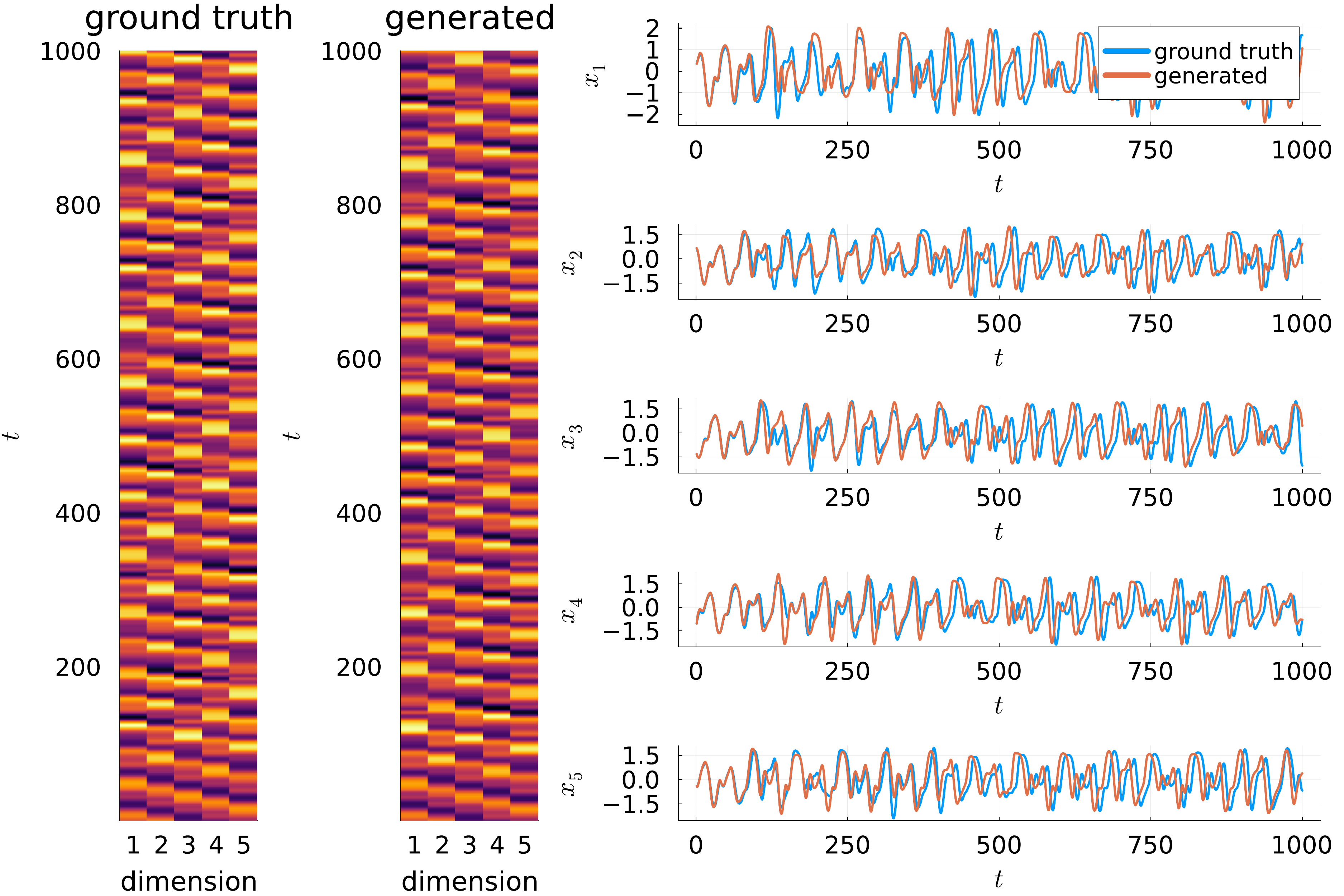}
    \caption{Illustration of the Lorenz-96 system ($N=5$, $F=8$), with spatiotemporal evolution on the left and single time series on the right. 
    }
    \label{fig:Lorenz96}
    \end{center}
\end{figure}

The spatially extended Lorenz-96 system \cite{lorenz96} is defined by 
\begin{equation}
    \frac{\text{d}x_i}{\text{d}t}=(x_{i+1}-x_{i-2})x_{i-1}-x_i+F,
\end{equation}
with system variables $x_i$, $i=1,...,N$, and forcing term $F$ (where $F=8$ puts the system into the chaotic regime). Furthermore, cyclic boundary conditions are assumed with $x_{-1}=x_{N-1}$, $x_0=x_N$, $x_{N+1}=x_1$, and $\Delta t = 0.04$ is used for numerical integration. Fig. \ref{fig:Lorenz96} provides an illustration. 
\newpage

\subsubsection{Bursting Neuron Model}
\begin{figure}[ht]
    \begin{center}
    \includegraphics[width=0.9\textwidth]{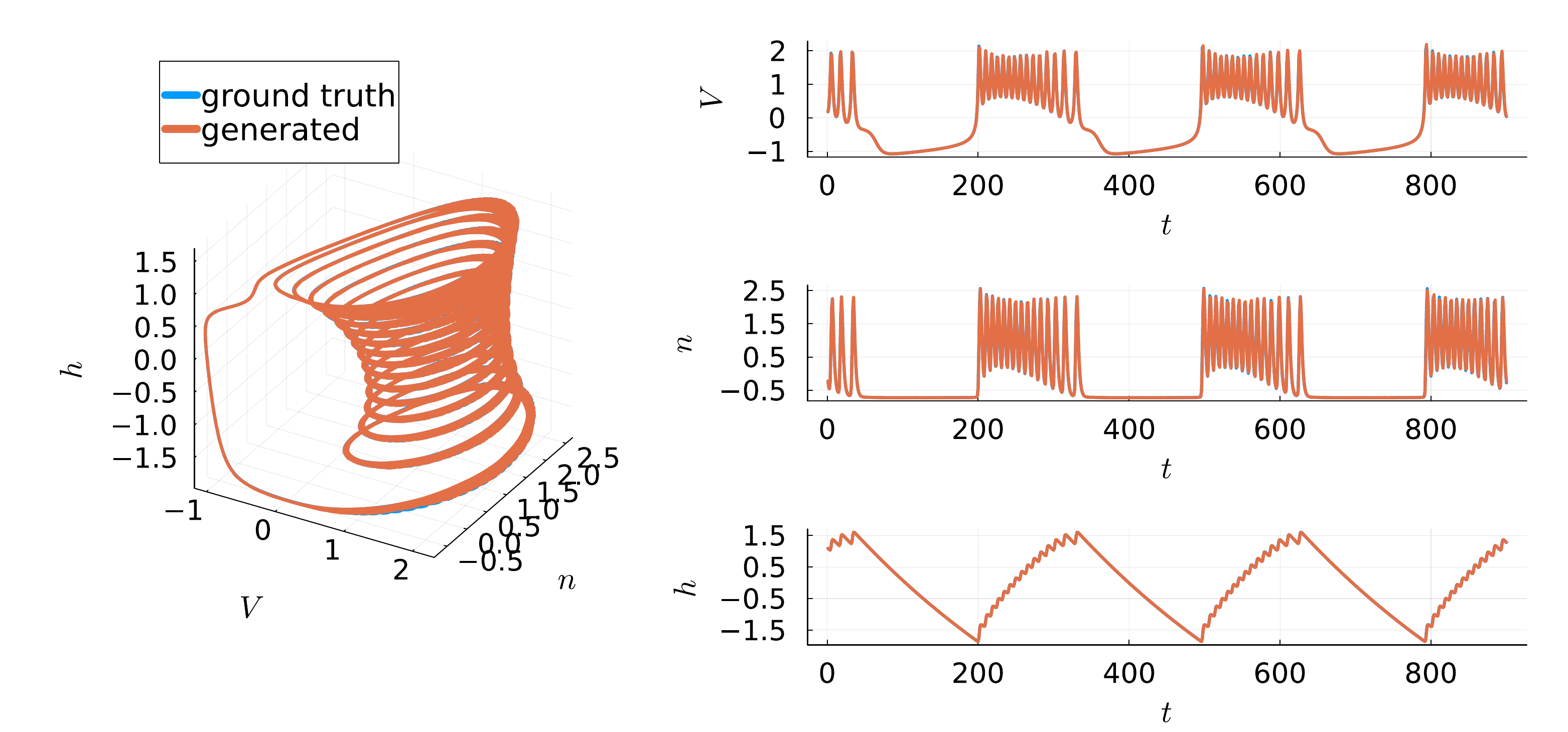}
    \caption[Bursting Neuron dataset]{Illustration of the biophysical neuron activity (blue; \citet{DURSTEWITZ20091189}) and a reconstruction using the PLRNN (red). Left: state space; right: time graphs. As this system is non-chaotic, true and reconstructed trajectories precisely overlap. 
    }
    \label{fig:BurstingNeuron}
    \end{center}
\end{figure}

The simplified $3$d biophysical model of a neuron used here is defined by \cite{DURSTEWITZ20091189} 
\begin{equation}
    \begin{gathered}
        -C_m\frac{\text{d}V}{\text{d}t}=g_L(V-E_L)+g_{N_a}m_\infty(V)(V-E_{N_a})+g_Kn(V-E_K)\\
        \;\;\;\;\;\;\;\;\;\;\;\;\;\;\;\;\;\;\;\;\;\;\;\;\;\;\;\;\;\;+g_Mh(V-E_K)+g_{NMDA}(1+0.33e^{-0.0625V})^{-1}(V-E_{NMDA})\\
        \frac{\text{d}h}{\text{d}t}=\frac{h_\infty(V)-h}{\tau_h}\\
        \frac{\text{d}n}{\text{d}t}=\frac{n_\infty(V)-n}{\tau_n},
    \end{gathered}
\end{equation}
where $V$ is the membrane voltage and $n$, $h$, are so-called gating variables controlling current flux through voltage-gated ion channels with vastly different time constants. Specifically, we used a standard set for the parameters which produces bursting activity (fast spikes riding on top of slow oscillations, see Fig. \ref{fig:BurstingNeuron}): 
\begin{equation}
    \begin{gathered}
        C_m=6\mu F,\;g_L=8mS,\;E_L=-80mV,\;g_{Na}=20mS \\
        E_{Na}=60mV,\;V_{hNa}=-20mV,\;k_{Na}=15,\;g_K=10mS, \\
        E_K=-90mV,\;V_{hK}=-25mV,\;k_K=5,\;\tau_n=1ms,\;g_M=25mS \\
        V_{hM}=-15mV,\;k_M=5,\;\tau_h=200ms,g_{NMDA}=10.2mS
    \end{gathered}
\end{equation}
The limit values of the ionic gating variables are given by
\begin{equation}
    \{m_\infty,n_\infty,h_\infty\}=\sigma\left(\frac{V-\{V_{hNa},V_{hK},V_{hM}\}}{\{k_{Na},k_K,k_M\}}\right)\;,
\end{equation}
where $\sigma(\cdot)$ is the common sigmoid function. 
\newpage

\subsubsection{Rössler System}
\begin{figure}[ht]
    \begin{center}
    \includegraphics[width=0.9\textwidth]{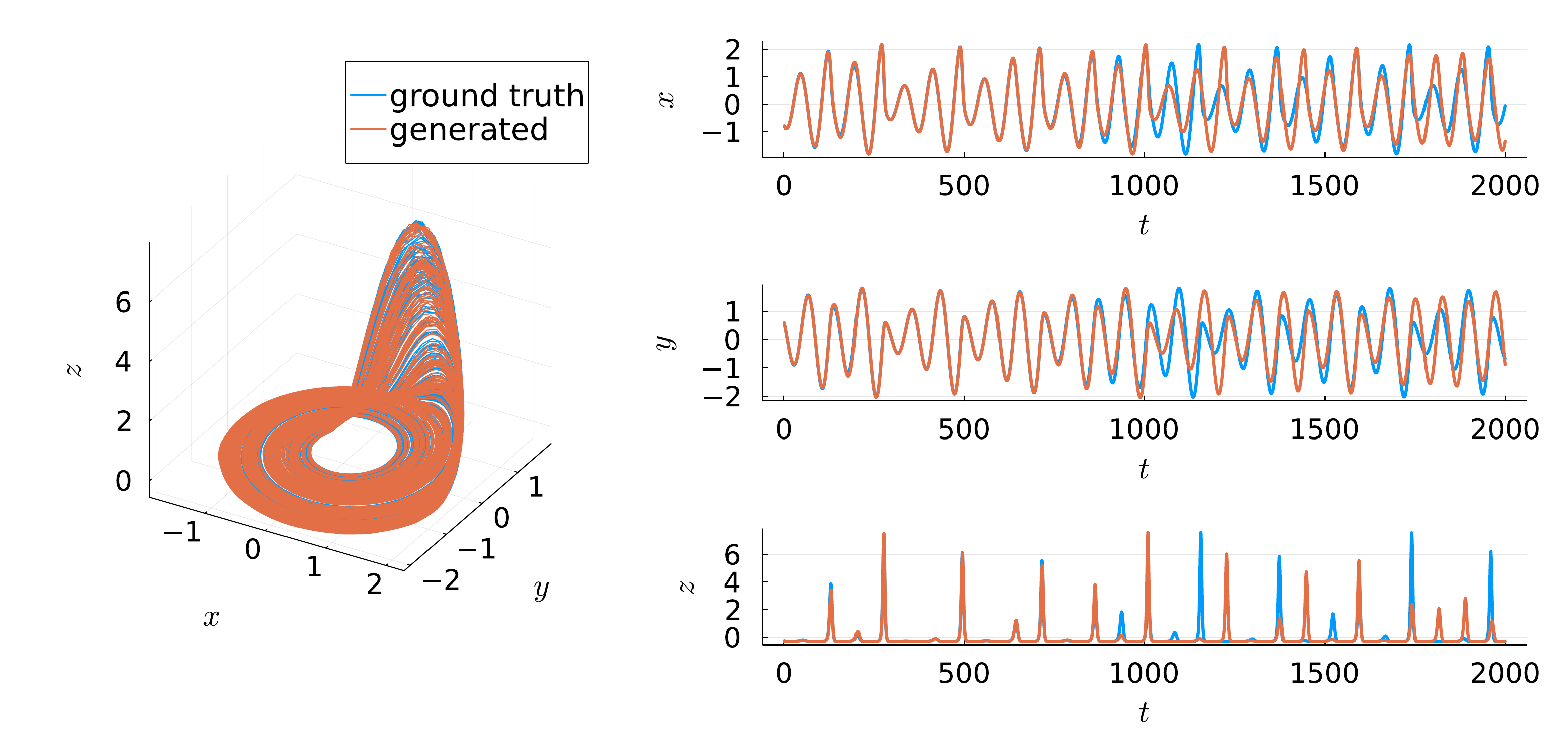}
    \caption{State space (left) and time graphs (right) for Rössler system with $a=0.2$, $b=0.2$, and $c=5.7$ (blue), and a PLRNN reconstruction (red). Note that despite the essentially perfect reconstruction in state space, the Rössler system's positive Lyapunov exponent causes the true and reconstructed trajectories to eventually diverge (yet their temporal structure remains the same).
    }
    \label{fig:Rossler}
    \end{center}
\end{figure}

The Rössler system, introduced by Otto Rössler in 1976 \cite{ROSSLER1976397} is a model that produces chaotic dynamics with nonlinearity in only one state variable, given by
\begin{align}
    \frac{\text{d}x}{\text{d}t} & = -y-z\\ \nonumber
    \frac{\text{d}y}{\text{d}t} & = x+ay\\ \nonumber
    \frac{\text{d}z}{\text{d}t} & = b+z(x-c),
\end{align}
where $a, b, c$, are parameters that control the behavior of the system (set to $a=0.2$, $b=0.2$, and $c=5.7$ here, within the chaotic regime). Here we solved this system with integration time step $\Delta t = 0.08$. See Fig. \ref{fig:Rossler} for an illustration.
\newpage

\subsubsection{Electrocardiogram (ECG) Data}
Electrocardiogram (ECG) time series were taken from the PPG-DaLiAdataset \cite{reiss2019deep}. Using a sampling frequency of $700Hz$, this translates to a recording duration of 600 seconds resulting in a time series spanning $T=419,973$ time points. The data were initially smoothed by applying a Gaussian filter (with $\sigma=6$, $l=8\sigma+1=49$). The time series is then standardized, followed by a delay embedding, using the \texttt{DynamicalSystems.jl} Julia library with embedding dimension $m=5$. In our experiments we use the first $T=100,000$ samples (approximately 143 seconds).
\begin{figure}[ht]
    \begin{center}
    \includegraphics[width=0.8\textwidth]{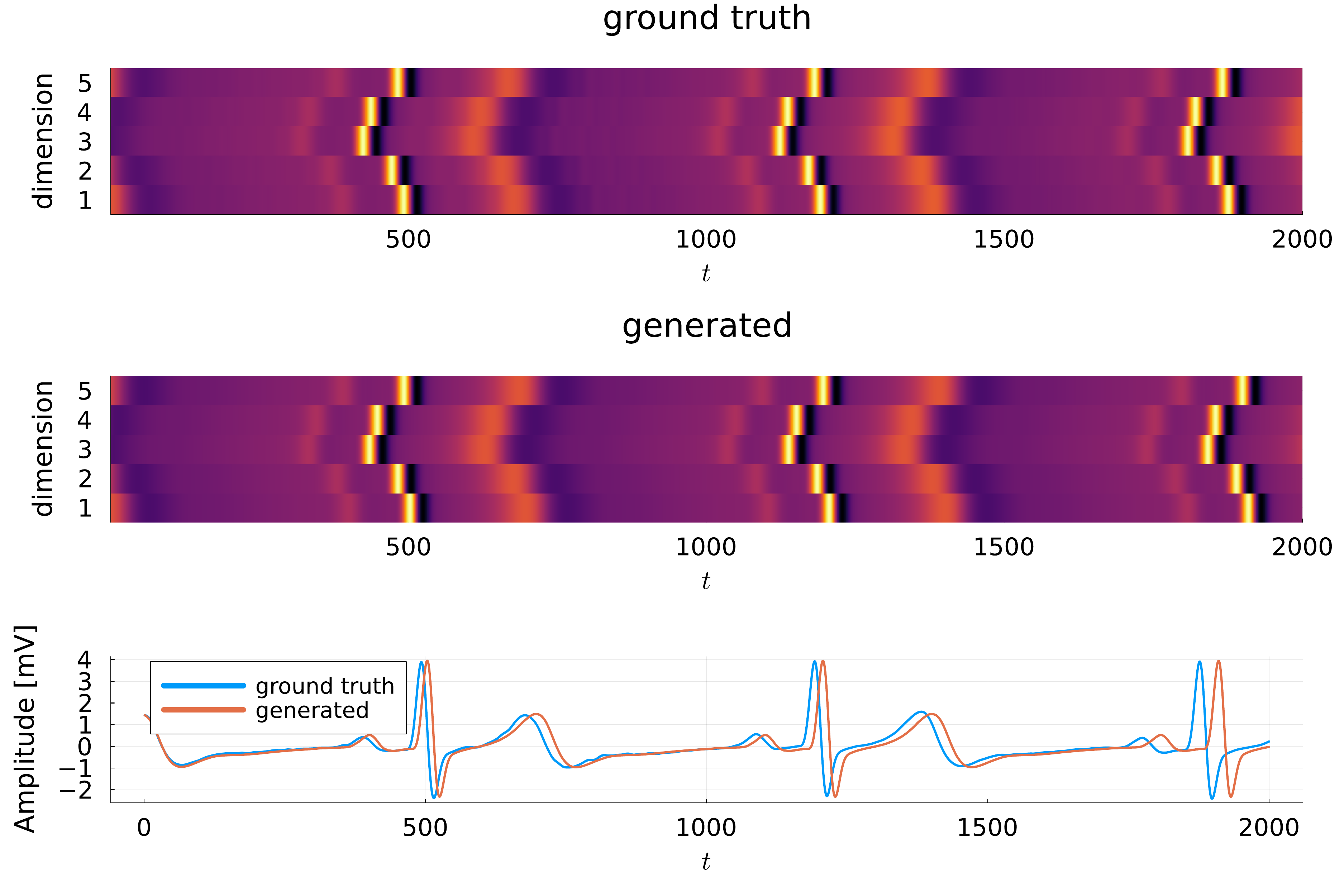}
    \caption{Illustration of a DS reconstruction of human ECG, with spatiotemporal evolution of delay-embedded time series (top panels) and single voltage signal (bottom). Note that the ECG signal is slightly chaotic \cite{hess2023generalized}, ultimately leading to divergence.
    }
    \label{fig:ECG}
    \end{center}
\end{figure}\newpage

\newpage
\subsection{Evaluation Measures} \label{sect:evaluation_measures}
To measure the geometrical (dis)agreement $D_{\text{stsp}}$ between true and model-generated attractors we use a \textit{Kullback-Leibler divergence}, as first suggested in \cite{koppe_fmri_2019}. It assesses the overlap between the true distribution $p_{true}(\bm{x})$ of trajectory points, and the distribution generated by the model $p_{gen}(\bm{x}|\bm{z})$, as 
\begin{equation} \label{eq:KL-divergence}
    \text{KL}(p_{true}(\bm{x})\lVert p_{gen}(\bm{x}|\bm{z}))=\int p_{true}(\bm{x})\log\frac{p_{true}(\bm{x})}{p_{gen}(\bm{x}|\bm{z})} d\bm{x}.
\end{equation}
Practically, this is evaluated by binning space into $k^N$ bins, with $k=30$ bins per dimension for $N=3$ and $k=8$ for $N=5$, estimating the occupation probabilities through the relative frequencies 
\begin{equation}
    p_i=\frac{n_i}{T},
\end{equation}
where $n_i$ is the number of time points falling into bin $i$, and taking 
\begin{equation}
    D_{\text{stsp}} \approx \sum_{i=1}^{k^N} p_{true;i}\log\frac{p_{true;i}}{p_{gen;i}}.
\end{equation}

To assess the agreement in long-term temporal structure between true and reconstructed systems, the \textit{Hellinger distance} $D_{\text{H}}$ between power spectra $f_i(\omega)$ and $g_i(\omega)$ of the true and generated time series, respectively, are computed separately for each dimension $i$ \cite{mikhaeil2022difficulty,hess2023generalized}. It is defined as 
\begin{equation} \label{eq:hellinger_distance}
    H(f_i(\omega),g_i(\omega))=\sqrt{1-\int_{-\infty}^\infty\sqrt{f_i(\omega)g_i(\omega)}\;d\omega}\;.
\end{equation}
Power spectra are computed through the Fast Fourier Transform, slightly smoothed using a Gaussian kernel, and normalized (see \citet{hess2023generalized} for details). The total measure $D_{\text{H}}$ is then defined as the average across all dimension-wise distances $H(f_i(\omega),g_i(\omega))$.

\subsection{Further Results}

\begin{figure}[ht!]
    \centering
    \includegraphics[width=0.7\textwidth]{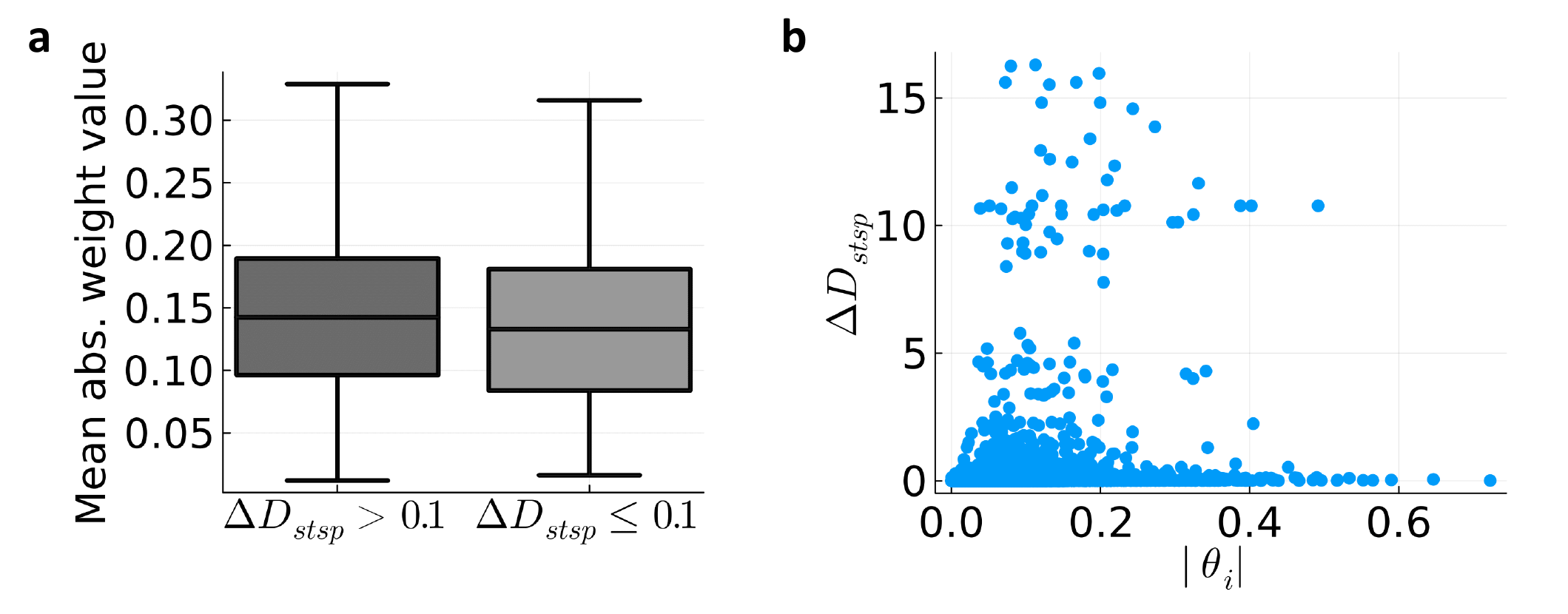}
    \caption{\textbf{a}) Weights with large ($\Delta D_{\text{stsp}}>0.1$) vs. low ($\Delta D_{\text{stsp}}\leq0.1$) impact on geometrical reconstruction quality when removing 3 weights simultaneously in each iteration. \textbf{b}) Change in geometrical agreement ($\Delta D_{\text{stsp}}$) as a function of pruned weight magnitude for PLRNNs trained on human ECG, cf. Fig \ref{fig:geometry-based_pruning}c.}
    \label{fig:weight_mag_Dstsp_further}
\end{figure}\newpage

\begin{figure}[ht!]
    \centering
    \begin{subfigure}
         \centering
         \includegraphics[width=0.9\textwidth]{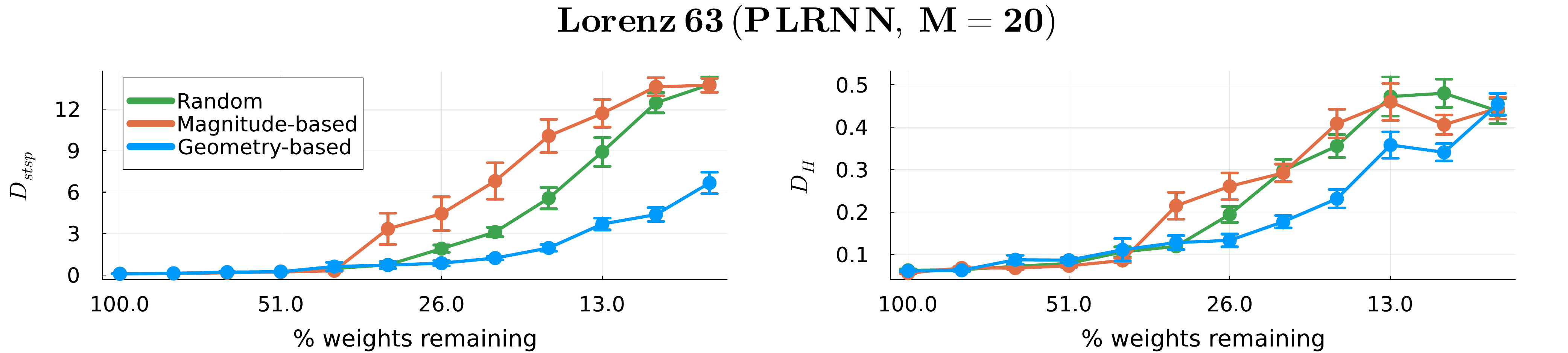}
    \end{subfigure}
    \begin{subfigure}
         \centering
         \includegraphics[width=0.9\textwidth]{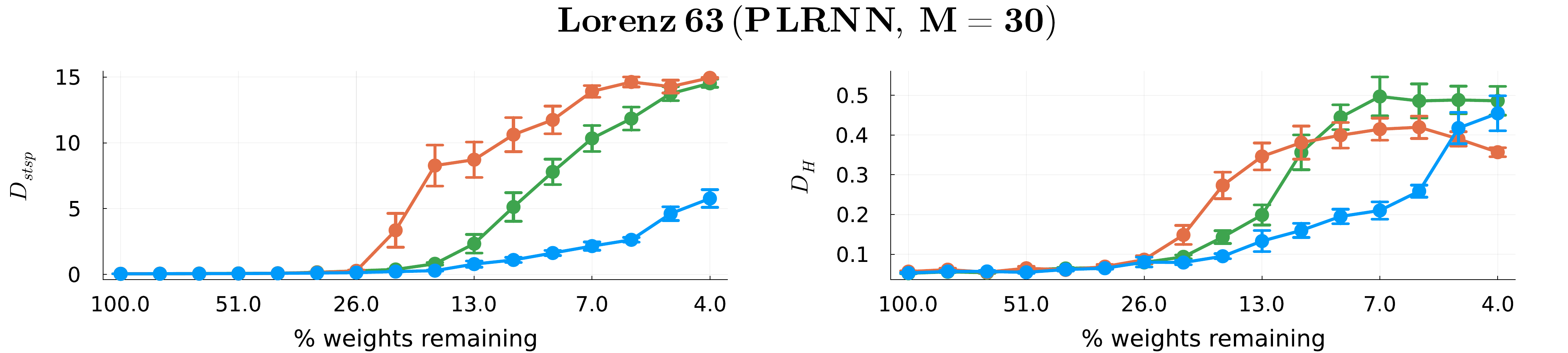}
    \end{subfigure}
    \caption{Same as Fig. \ref{fig:pruning_measures_performance} for PLRNNs of different (smaller) network (latent space) size (Fig. \ref{fig:pruning_measures_performance} was produced for $M=50$). Error bars = SEM.
    }
    \label{fig:pruning_smaller_initial_modelsize}
\end{figure}

\begin{figure}[h!]
    \centering
    \begin{subfigure}
         \centering
         \includegraphics[width=0.9\textwidth]{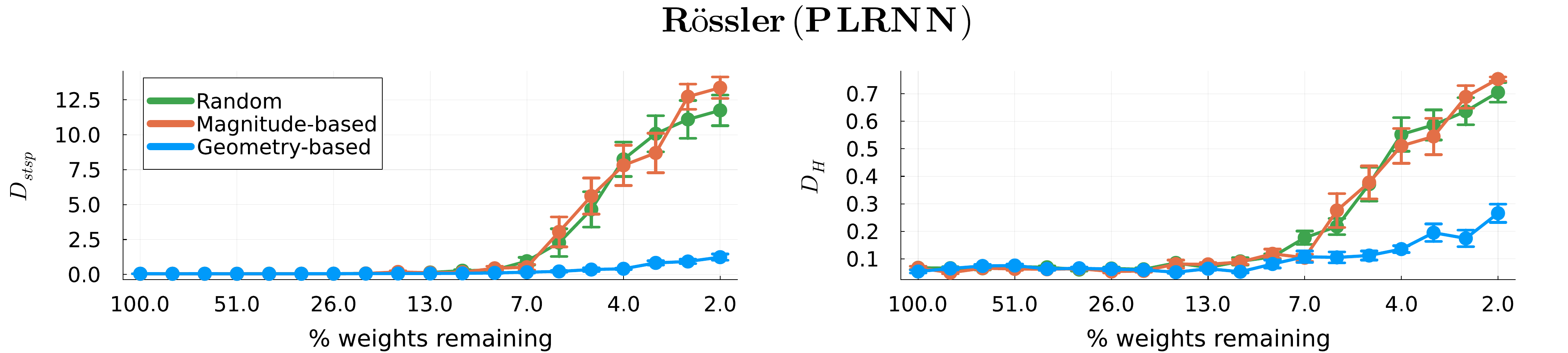}
    \end{subfigure}
    \begin{subfigure}
         \centering
         \includegraphics[width=0.9\textwidth]{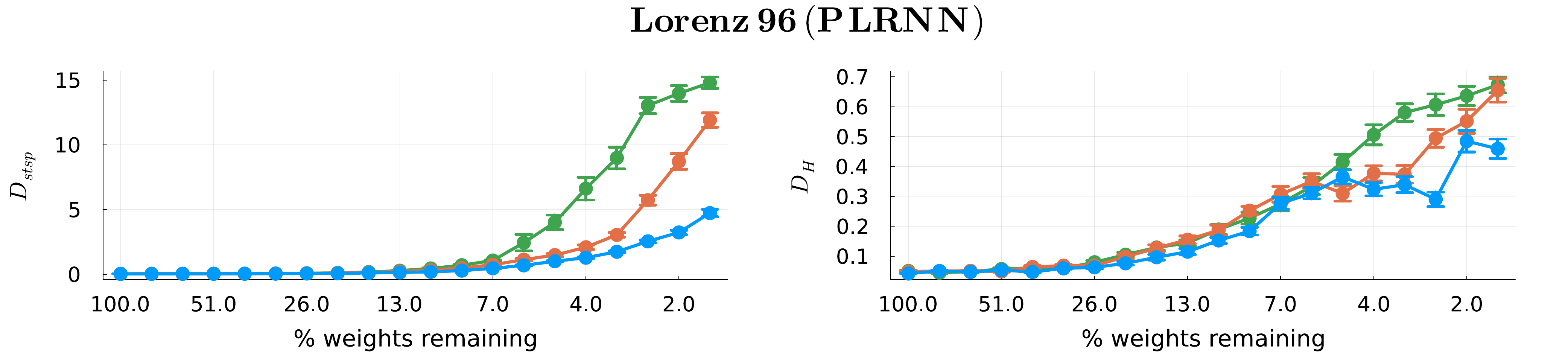}
    \end{subfigure}
    \begin{subfigure}
         \centering
         \includegraphics[width=0.9\textwidth]{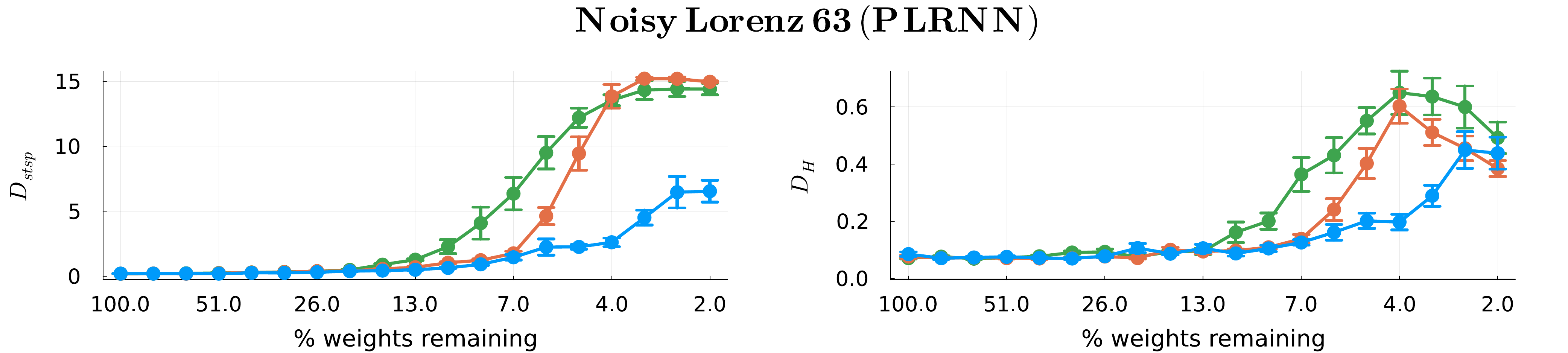}
    \end{subfigure}
    \caption{
    Same as Fig. \ref{fig:pruning_measures_performance} for the chaotic Rössler system (top row), the chaotic Lorenz-96 system (center), and the chaotic Lorenz-63 system with high ($25\%$) noise level (bottom). Error bars = SEM.
    }
    \label{fig:pruning_measures_performance_further}
\end{figure}

\begin{figure}[ht!]
    \centering
    \includegraphics[width=0.9\textwidth]{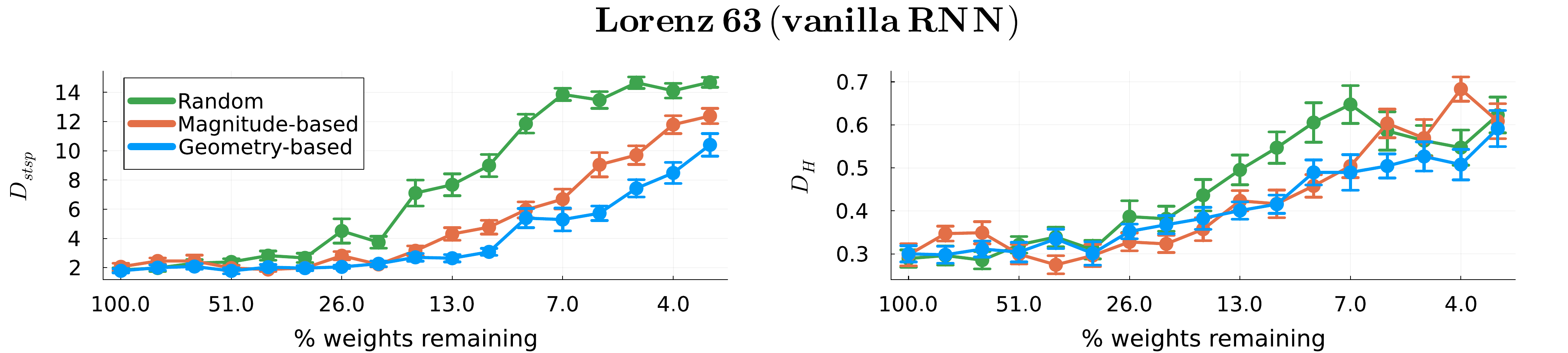}
    \caption{Same as Fig. \ref{fig:pruning_measures_performance} for a vanilla RNN, $\bm{z}_t=\phi(\bm{W}\bm{z}_{t-1}+\bm{C}\bm{x}_t+\bm{b})$ with $\hat{\bm{x}}_t=\bm{B}\bm{z}_t+\bm{h}$. Error bars = SEM.
    }
    \label{fig:pruning_RNN}
\end{figure}

\begin{figure}[ht!]
    \centering
    \begin{subfigure}
         \centering
         \includegraphics[width=0.45\textwidth]{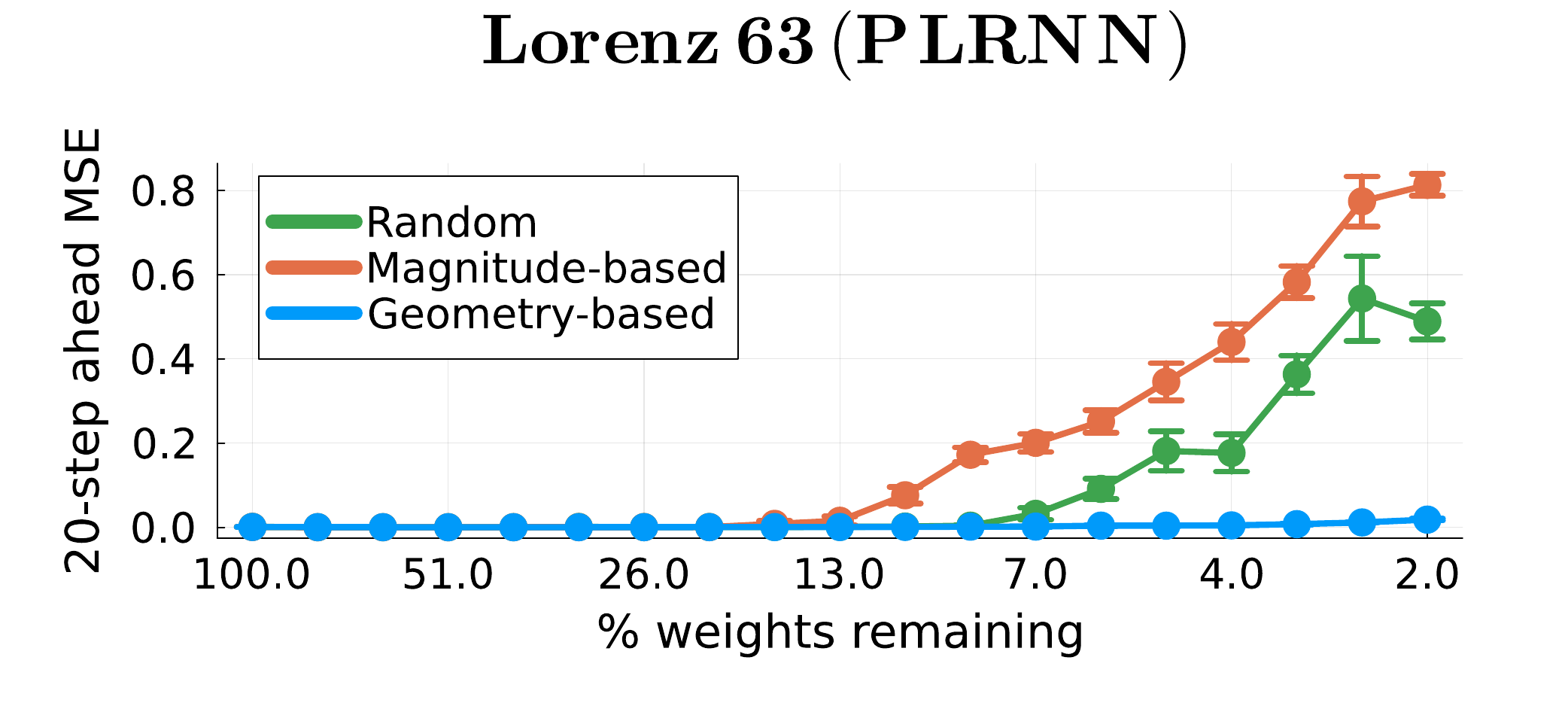}
    \end{subfigure}
    \begin{subfigure}
         \centering
         \includegraphics[width=0.45\textwidth]{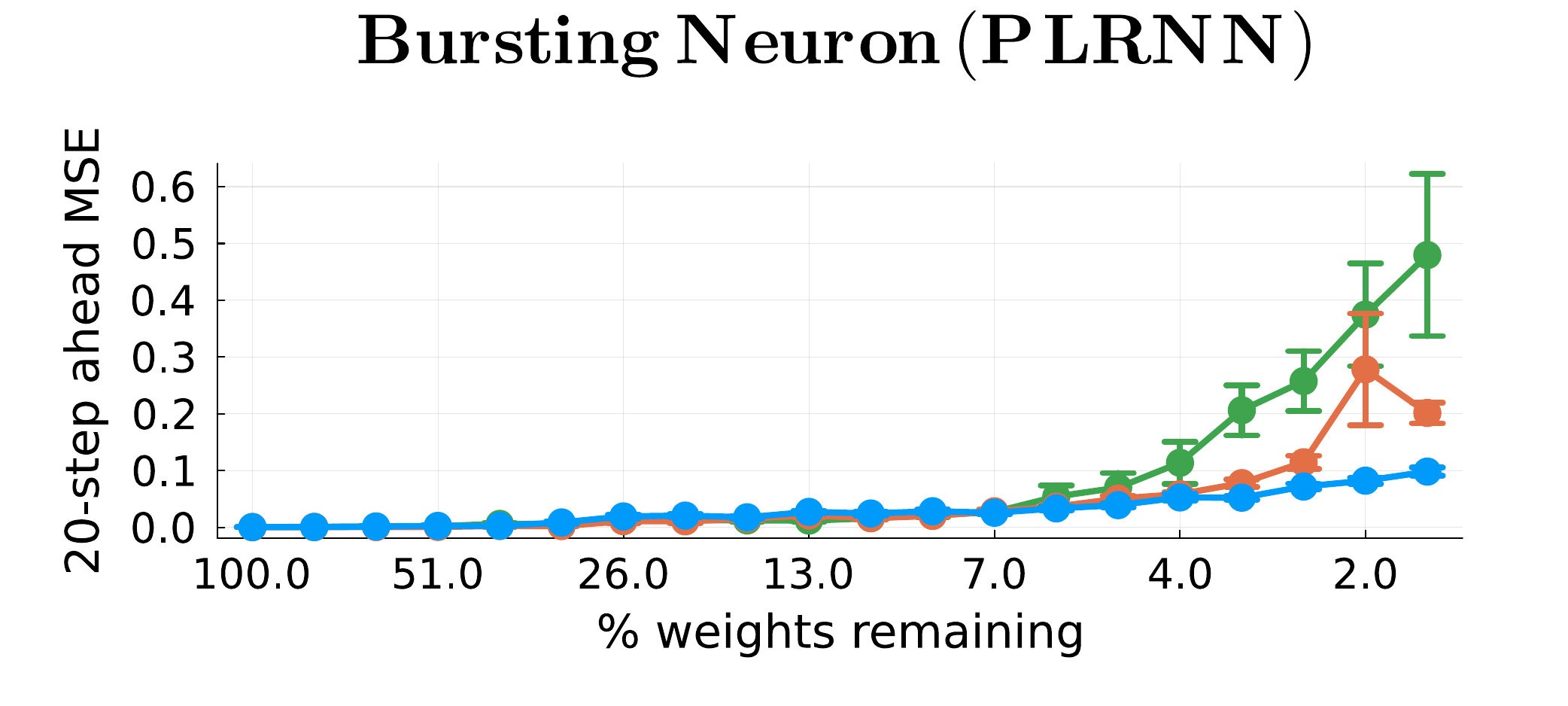}
    \end{subfigure}
    \vskip\floatsep
    \begin{subfigure}
         \centering
         \includegraphics[width=0.45\textwidth]{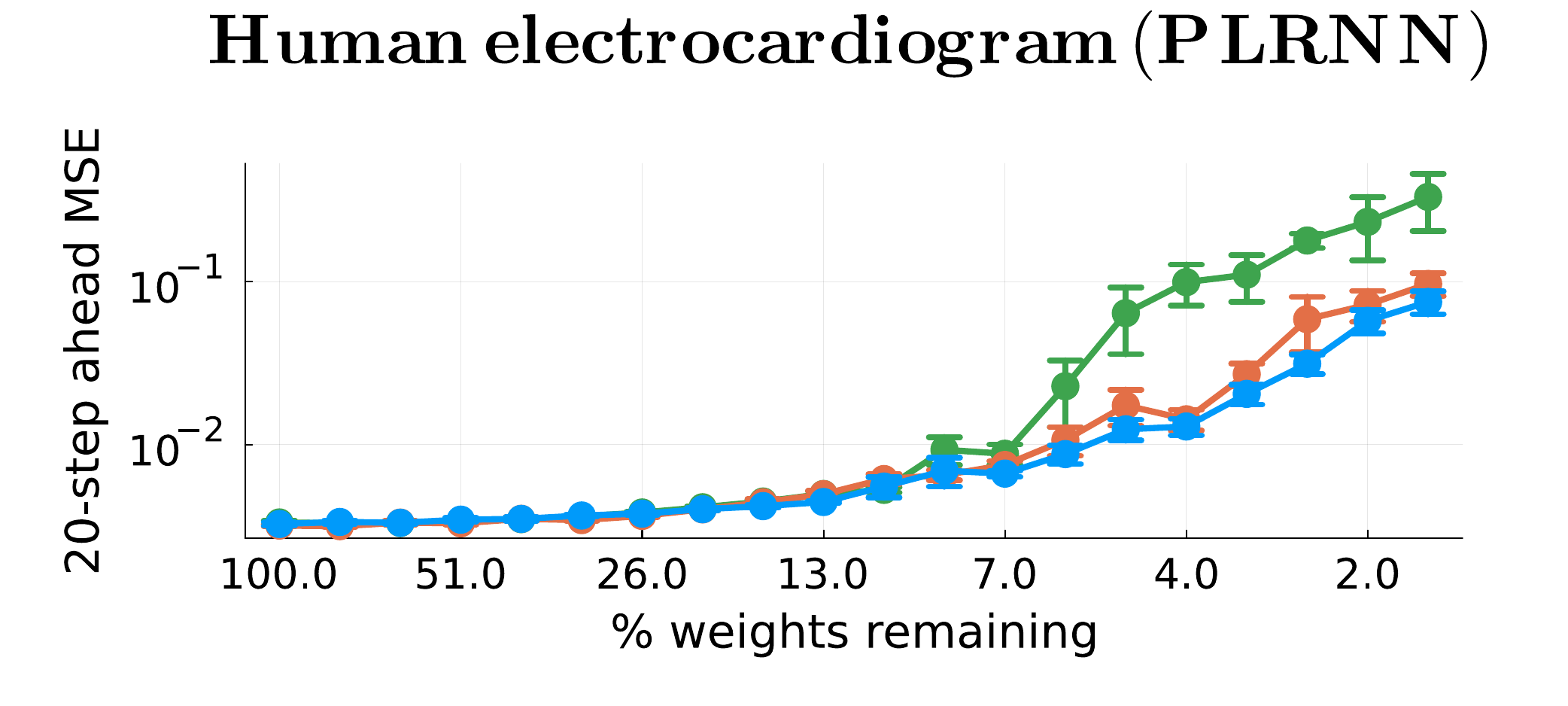}
    \end{subfigure}
    \caption{Quantification of DS reconstruction quality in terms of the mean-squared 20-step-ahead prediction error as a function of network pruning (x-axis, exponential scale) and different pruning criteria for the Lorenz-63, bursting neuron, and ECG. Error bars = SEM.
    }
    \label{fig:pruning_performance_PE}
\end{figure}

\begin{figure}[ht!]
    \centering
    \includegraphics[width=0.5\textwidth]{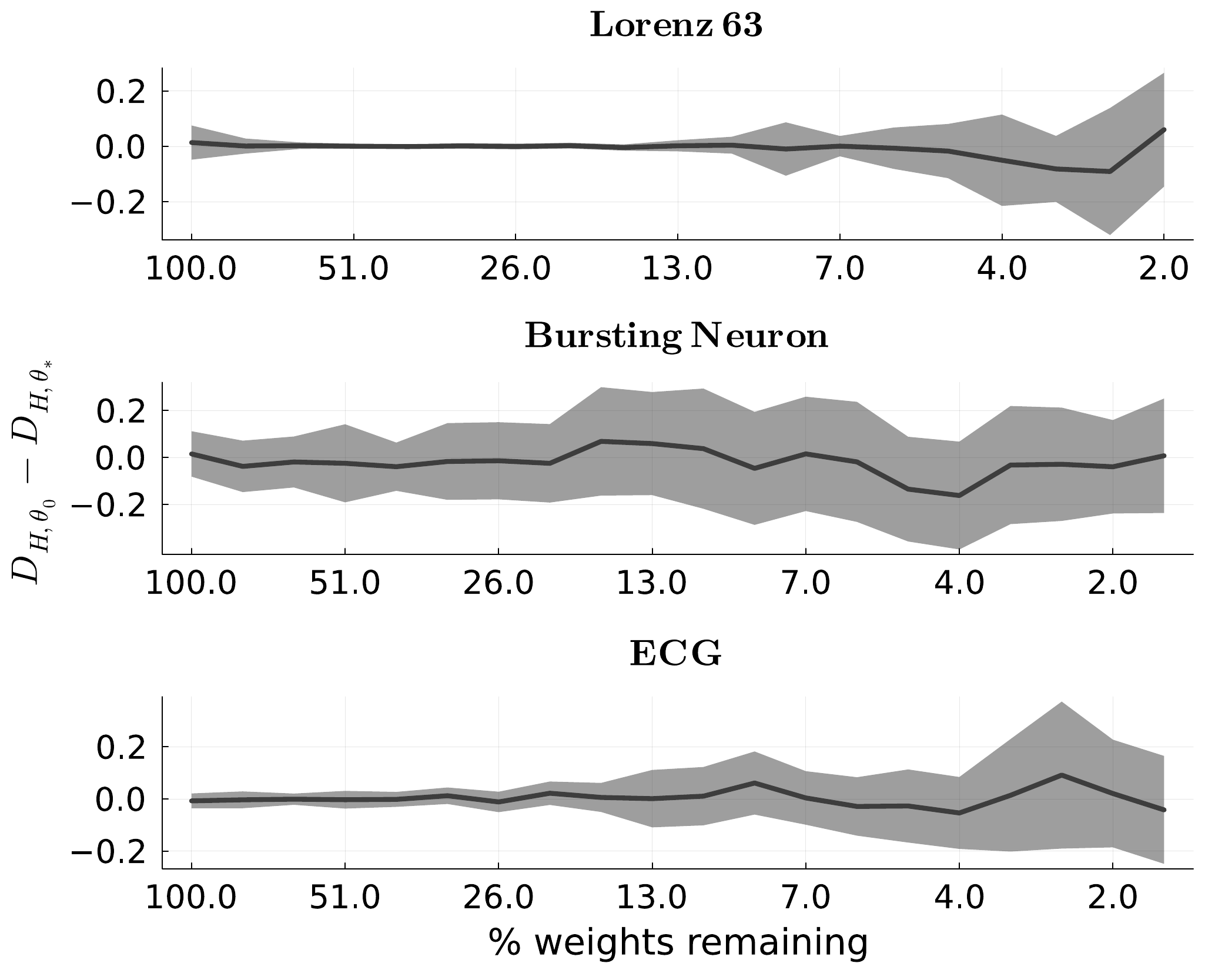}
    \caption{Same as Fig. \ref{fig:initial_weights} for $D_{\text{H}}$: Difference in $D_{\text{H}}$ when using the initial weights $\bm{\theta}_0$ and reinitialized weights $\bm{\theta}_*$ shows there is no notable influence of the specific weight initialization.
    }
    \label{fig:reinitialization_Dhell}
\end{figure}\newpage

\begin{figure}[ht!]
    \centering
    \includegraphics[width=0.65\textwidth]{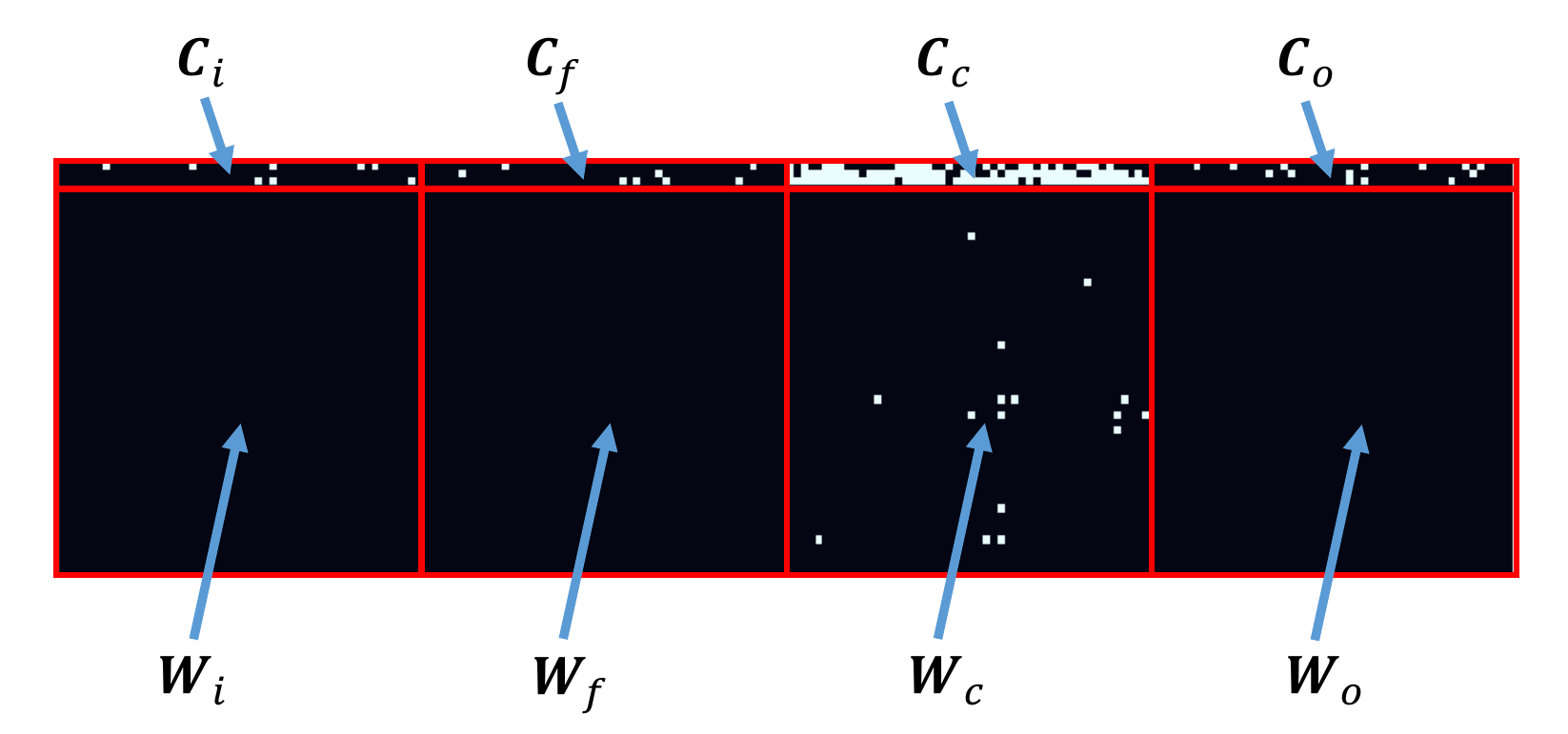}
    \caption{LSTM pruning masks for all types of LSTM weight matrices (input, forget, cell state, and output): Geometry-based pruning is able to identify the relevance of different weight matrices to the performance, therefore excluding $\bm{W}_i,\;\bm{W}_f,\;\bm{W}_o$ from the model, leading to interpretable pruning results in terms of inter-cell connections and their links to observable outputs.}
    \label{fig:LSTM_topology}
\end{figure}

\begin{figure}[ht!]
    \centering
    \includegraphics[width=0.5\textwidth]{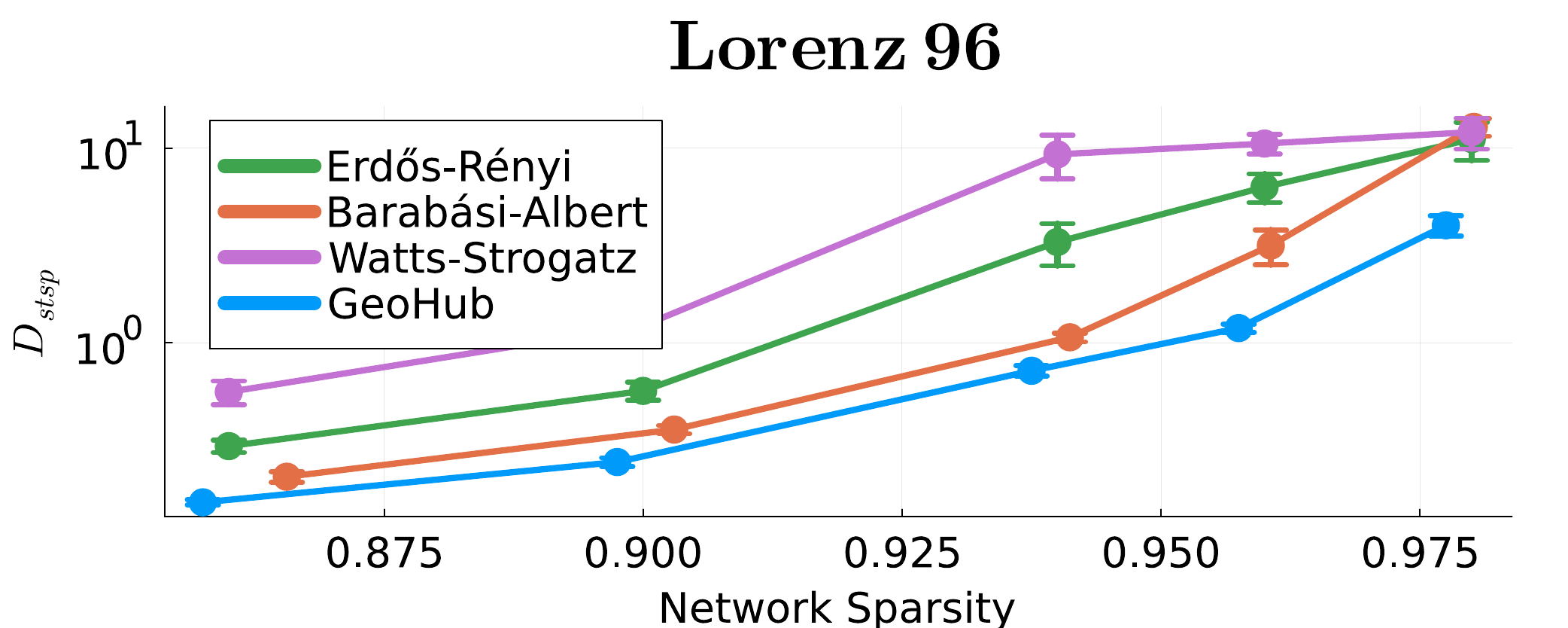}
    \caption{Reconstruction results in terms of state space divergence $D_{\text{stsp}}$ as a function of network sparsity $s=1-\frac{| \bm{m}|}{|\bm{W}|}$ for Erdős–Rényi, Barabási-Albert, Watts-Strogatz and GeoHub graph algorithms. Error bars = SEM.}
    \label{fig:graph_performance_Lorenz96}
\end{figure}

\subsection{Network Topology} \label{PLRNN topology}

\subsubsection{Graph Properties} 
When characterizing graphs, we seek properties that are preserved under different isomorphisms. A fundamental characteristic is the degree distribution $P(k)$, where the degree is the number of connections associated with a node, which may be further distinguished into incoming ($k^{in}$) and outgoing connections ($k^{out}$). Two major types of graphs are \textit{single-scale} and \textit{scale-free} graphs, where the latter is surprisingly common in many real-world systems \cite{Barabasi99emergenceScaling}. While single-scale graphs are often well captured by the binomial or Poisson distribution, $P(k)$ in scale-free graphs follows a power law 
\begin{equation}
    P(k)\propto k^{-\gamma}\;.
\end{equation}
Such graphs are characterized by the presence of many nodes with low and a few nodes with high degrees, often referred to as hubs.

Many algorithms exist for finding the shortest path between two nodes in a graph, as required to compute the mean average path length $L(G)$ in Eqn. \ref{eq:average_path_length}. Here we use the Floyd-Warshall algorithm \cite{Floyd1962Algorithm9S,Warshall1962ATO}. While originally the clustering coefficient $C(G)$, as used in Eqn. \ref{eq:clustering_coefficient}, was defined only for undirected graphs \cite{watts_collective_1998}, here we use the formulation proposed in \cite{PhysRevE.76.026107} which considers the more general case with directed edges. When two neighbors of a node $v_i$ are connected by any type of edge, they form a triangle. The number of directed triangles $t_i$ formed by all neighbors of $v_i$ can be calculated from the adjacency matrix $\bm{A}^{\text{adj}}$ as \cite{PhysRevE.76.026107}
\begin{equation}
    t_i=\frac{1}{2}\sum_j\sum_h(A^{\text{adj}}_{ij}+A^{\text{adj}}_{ji})(A^{\text{adj}}_{ih}+A^{\text{adj}}_{hi})(A^{\text{adj}}_{jh}+A^{\text{adj}}_{hj})=\frac{1}{2}\left(\bm{A}^{\text{adj}}+\left(\bm{A}^{\text{adj}}\right)^T\right)^3_{ii}\;.
\end{equation}
The total number of possible triangles $T_i$ is given by \cite{PhysRevE.76.026107} 
\begin{equation}
    T_i=k_i^{tot}(k_i^{tot}-1)-2k_i^{\leftrightarrow}\;,
\end{equation}
where $k_i^{tot}=k_i^{out}+k_i^{in}$ is the total degree and $k_i^{\leftrightarrow}$ is the number of bilateral edges connected to node $v_i$. The ratio of these two quantities yields the clustering coefficient of a node $v_i$, 
\begin{equation}
    C_i=\frac{(\bm{A}^{\text{adj}}+\left(\bm{A}^{\text{adj}}\right)^T)^3_{ii}}{2[d_i^{tot}(d_i^{tot}-1)-2d_i^{\leftrightarrow}]},
\end{equation}
and the average across all nodes yields Eqn. \ref{eq:clustering_coefficient} in sect. \ref{sect:analysis_network_topology} \cite{watts_collective_1998}. Intuitively, this quantity measures how likely it is that any two neighbors of a given node are also immediate neighbors. Based on these quantities, \citet{watts_collective_1998} introduced the idea of small-world graphs, which are characterized by a high clustering coefficient yet short average path length. To measure the small-worldness of any graph, these characteristics can be combined in a single measure called the small-world index $\text{SWI}$ \cite{neal_2017}. Its definition requires a reference, for which Erdős–Rényi random graphs \cite{erdos59a} and ring lattice graphs are used. Random graphs have a short average path length $L_r$ and a low clustering coefficient $C_r$, while, on the contrary, lattice graphs have a high average path length $L_l$ and high clustering coefficient $C_l$. Based on these one defines
\begin{equation} \label{eq:SWI}
    \text{SWI}=\frac{L-L_l}{L_r-L_l}\cdot\frac{C-C_r}{C_l-C_r}\;.
\end{equation}
This measure is thus normalized and further clipped into the range $0\leq \text{SWI}\leq1$, where a value close to $1$ indicates higher small-worldness.

\subsubsection{Graphs Obtained from Geometry-based Pruning}
\begin{figure}[h!]
    \begin{center}
    \includegraphics[width=0.9\textwidth]{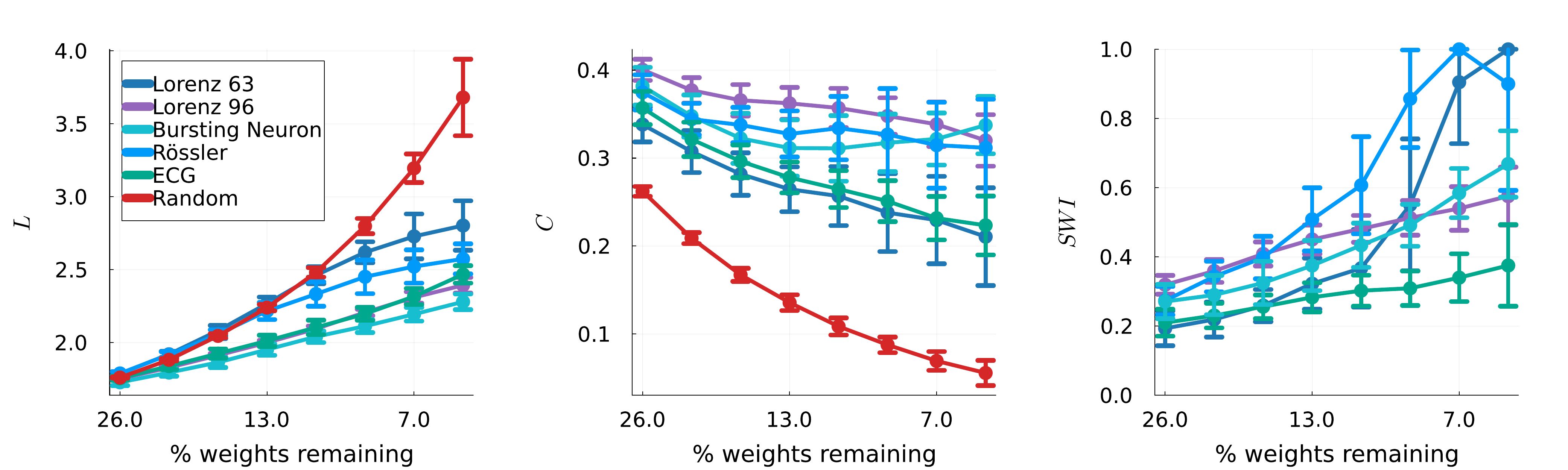}
    \caption{Average path length $L$ (left), clustering coefficient $C$ (center) and SWI (right) for the PLRNN topologies obtained through geometry-based pruning, as a function of the \% of weights remaining, and for the different benchmarks used in here. Random graph for comparison in red. Note that x-axes are on an exponential scale for better visualization.
    }
    \label{fig:subnetwork_smallworld}
    \end{center}
\end{figure}

The graph structure of pruned PLRNNs is given through their pruning masks $\bm{m}$. Investigating the complementary cumulative degree distribution in $\bm{m}$ on a logarithmic scale (Fig. \ref{fig:network_graph_properties}a) reveals a scale-free distribution, indicating the existence of hubs. Those hubs are primarily associated with PLRNN nodes that directly link to the outputs (observations) via the identity mapping used here (see Eqn. \ref{eq:identity_model} and sect. \ref{DSR_model_training}), see Fig. \ref{fig:network_graph_properties}b. The difference between the $k^{in}$- and the $k^{out}$-distribution furthermore indicates directedness of the edges, i.e. directed information flow. The pruned PLRNNs also exhibit properties of small-world graphs with a higher clustering $C$ and a path length $L$ not exceeding the one of random graphs (Fig. \ref{fig:subnetwork_smallworld} left), and as evidenced by the SWI (Eqn. \ref{eq:SWI}; Fig. \ref{fig:subnetwork_smallworld} right). Algorithm \ref{alg:PLRNN_subnetwork_graph} implements a procedure that respects all these properties, namely directedness in edges, hub nodes preferentially associated with in-going connections and, mainly, for all output units, and small-worldness.

In more detail, since the network topologies obtained by pruning highlight an important difference between readout nodes and hidden nodes, we start constructing the graph by a fully connected model with readout dimension $N$. Due to the strong hub characteristics we expand this network based on the preferential attachment mechanism of the Barabási-Albert model. To endow specifically the readout nodes with hub-like features, the probability of connecting to these nodes is increased by a term  $k_{mean}^{in}=\frac{1}{n}\sum_{i=1}^n k_i^{in}$ as observed empirically. To introduce directness within the graph, we further decouple the generation of incoming and outgoing edges, establishing incoming edges first. The probability for a random node connecting to node $v_i$ is thus given by
\begin{equation} \label{eq:PLRNN_subnetwork_graph_ingoing}
    p^{in}_i=\begin{cases}
        \frac{1}{\mathcal{N}}\;\left(k^{in}_i+ \frac{1}{2}k_{mean}^{in}\right), & \text{if $i\in 1,...,N$} \\
        \frac{1}{\mathcal{N}}\;k^{in}_i & \text{if $i\in N+1,...,n$}
    \end{cases}\;
\end{equation}
where the normalization is $\mathcal{N}=\sum_{j=1}^n k^{in}_j+\frac{N}{2} k_{mean}^{in}$. In a next step, for each node outgoing edges are created, taking the already established structure into account. To prevent strong hub formation in outgoing edges, a term proportional to $k_{mean}^{out}=\frac{1}{n}\sum_{i=1}^n k_i^{out}$, related to the empirically observed out-degree, is added, equalizing the connection probabilities $p_i$. Specifically, the probability for a random node to receive a connection from node $v_i$ is given by
\begin{equation} \label{eq:PLRNN_subnetwork_graph_outgoing}
    p^{out}_i=\frac{1}{\mathcal{N}}\;\left(k^{in}_i+k^{out}_i+ \frac{1}{4}k_{mean}^{out}\right)\;,
\end{equation}
where the normalization is $\mathcal{N}=\sum_{j=1}^n (k^{in}_j+k^{out}_j)+\frac{n}{4} k_{mean}^{out}$. (For numerical stability, a constant $0.05$ is added to each $k^{in}_i$ and $k^{out}_i$.) The algorithm is described in detail in Algorithm \ref{alg:PLRNN_subnetwork_graph}. The user needs to specify a hyperparameter $k$ that determines the number of edges connected to a node, based on which the mean in-degree $k_{mean}^{in}$ and mean out-degree $k_{mean}^{out}$ required for calculating the probabilities is exactly given, making the calculations tractable. Using this model, we are able to generate graphs that fulfill the characteristics found empirically as described further above. 

\begin{figure}[ht]
    \begin{center}
    \includegraphics[width=0.4\textwidth]{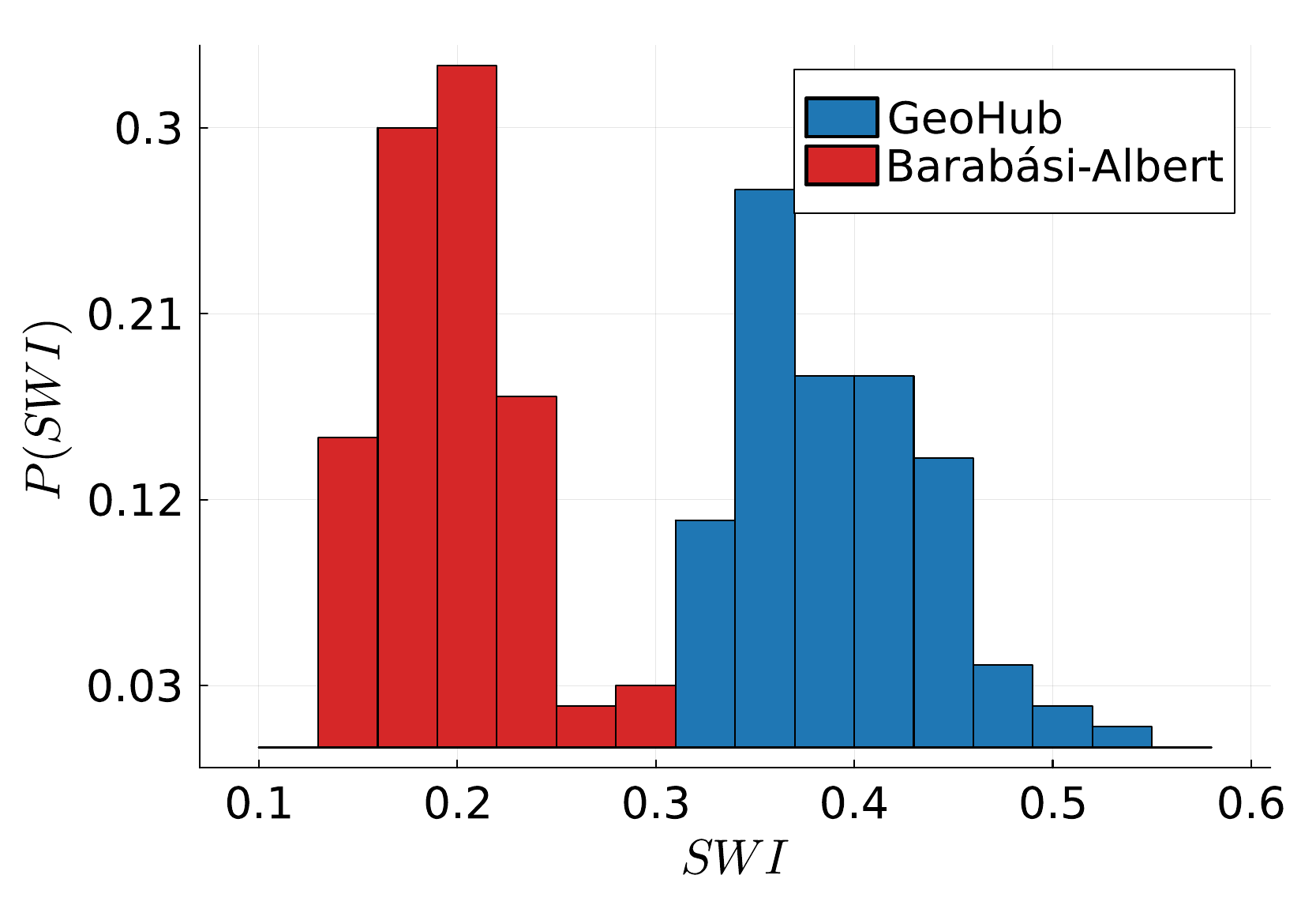}
    \caption[Small-world index(SWI) of generalized PLRNN subnetwork graphs]{Distribution of the SWI, Eqn. \ref{eq:SWI}, for the GeoHub graph (blue) and the Barabási-Albert model (red), for $n=100$ and $s\approx90\%$.}
    \label{fig:PLRNN_subnetwork_SWI}
    \end{center}
\end{figure}

\begin{figure}[h]
    \centering
    \includegraphics[width=0.7\textwidth]{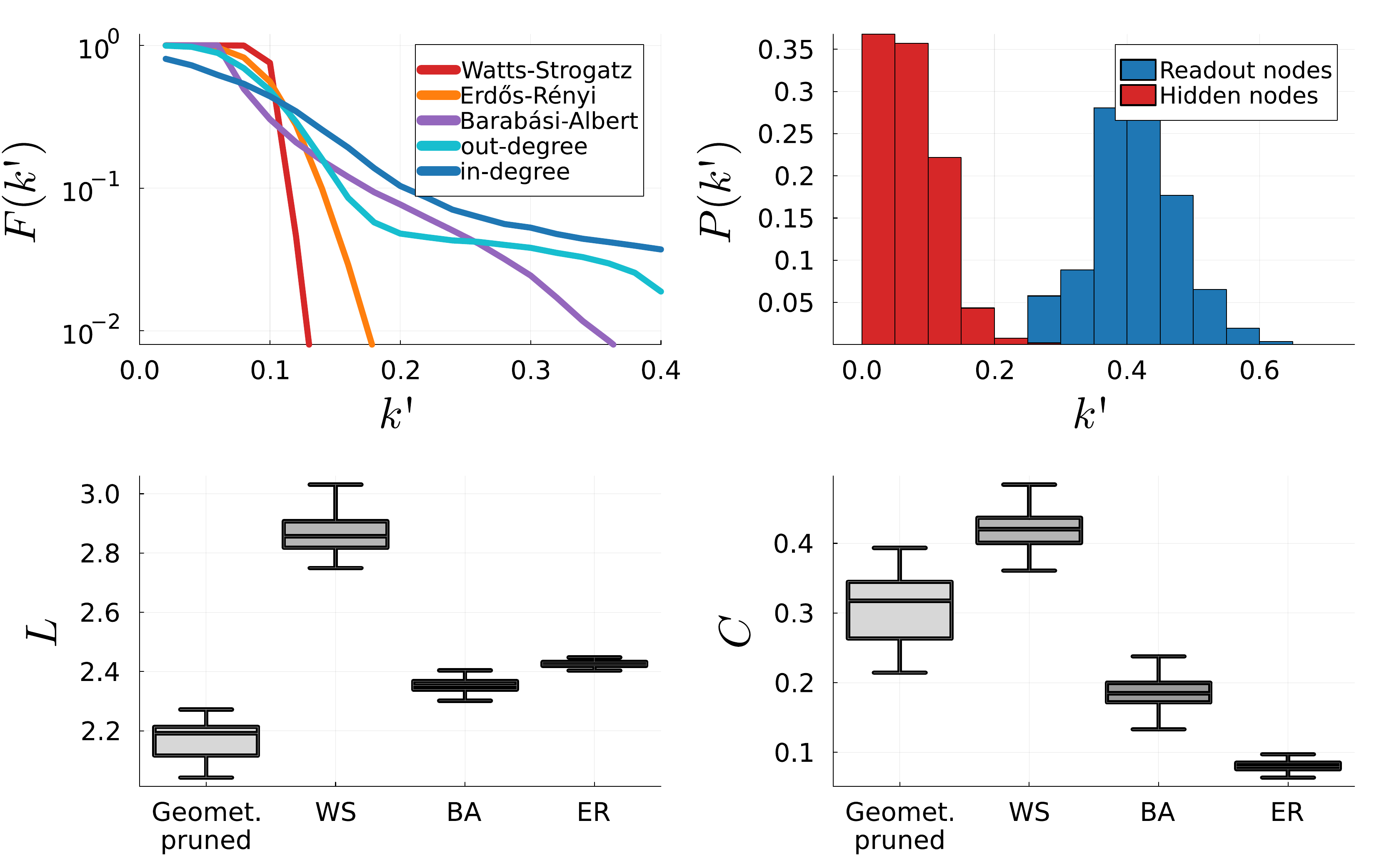}
    \caption{Same as Fig. \ref{fig:network_graph_properties} for network size $M=100$.}
    \label{fig:network_graph_properties_M100}
\end{figure}

\begin{figure}[ht]
    \begin{center}
    \includegraphics[width=1.0\textwidth]{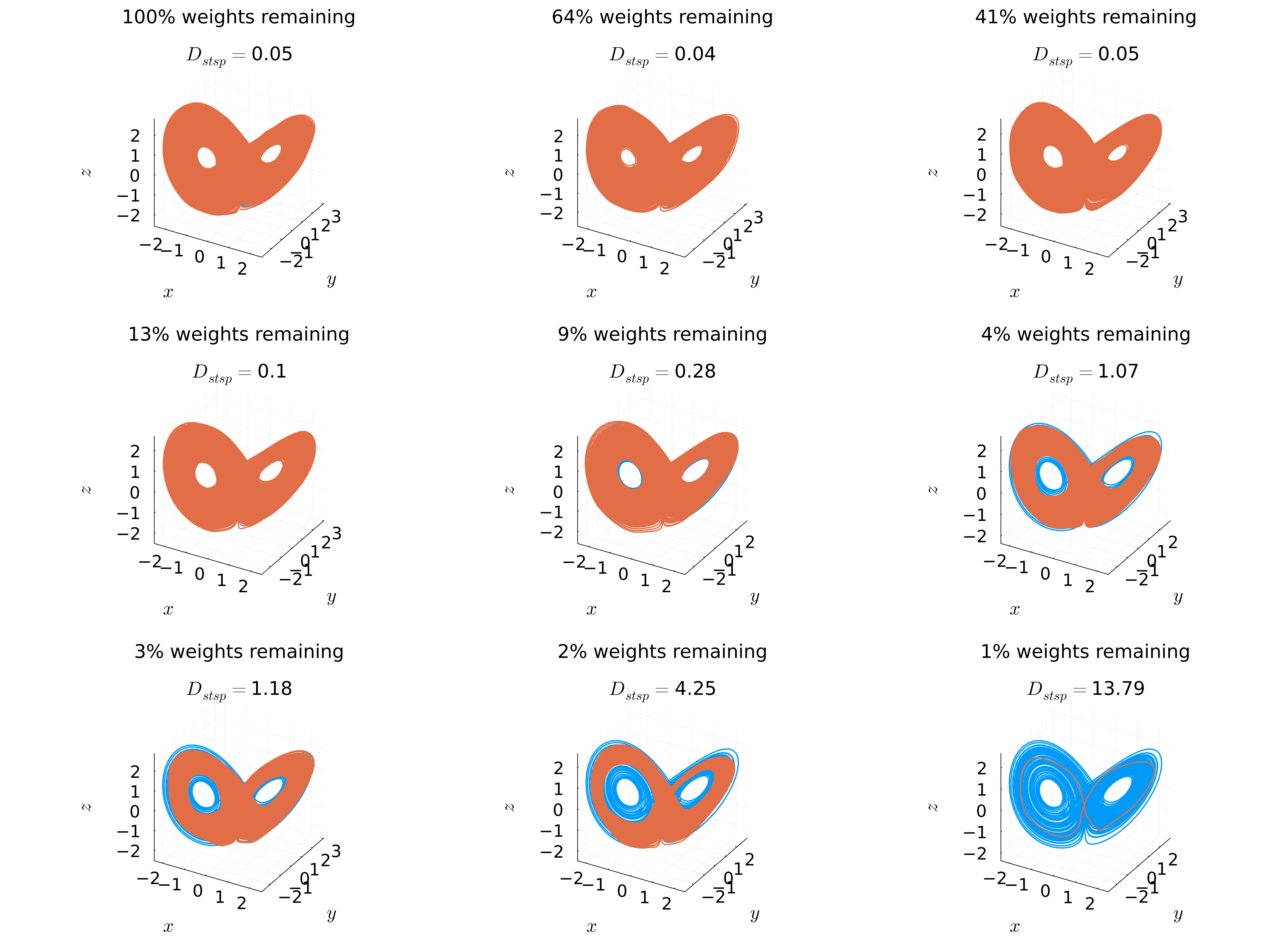}
    \caption{Example reconstructions of the Lorenz-63 at different levels of geometry-based parameter pruning, with an indication of reconstruction quality as measured by $D_{\text{stsp}}$ on top.}
    \label{fig:reconstructions_geometry_based}
    \end{center}
\end{figure}

\clearpage
\subsection{Graph Algorithms} \label{ch:graph_models}

\subsubsection{Erdős-Rényi Model}
Algorithm \ref{alg:Erdős-Rényi} produces an undirected Erdős-Rényi random graph, where edges are chosen uniformly at random. Required inputs are the number of nodes $n$ and the requested number of edges $k$.
\begin{figure}[ht]
\centering
\begin{minipage}{.9\linewidth}
\begin{algorithm}[H]
\caption{Erdős–Rényi model}
\SetKwInOut{Input}{Input}\SetKwInOut{Output}{Output}
\Input{Number of nodes $n$, number of edges $k$}
\Output{Erdős-Rényi graph $G(V,E)$}
\BlankLine
$G \gets$ empty graph with $|V|=n$ and $|E|=0$\;

\While{$|E|<k$}{
    Choose a random pair of nodes $v_i$ and $v_j$ from $G$\;
    
    \If{$e_{ij}\notin E$}{
        \BlankLine
        $E=E\cup \{e_{ij}\}$\;
    }
}
\label{alg:Erdős-Rényi}
\end{algorithm}
\end{minipage}
\end{figure}

\subsubsection{Watts-Strogatz Model}
The Watts-Strogatz model yields undirected small-world graphs. For constructing such a graph the number of nodes $n$, the desired degree $k$, and the rewiring probability $p$ need to be specified. 
\begin{figure}[ht]
\centering
\begin{minipage}{.9\linewidth}
\begin{algorithm}[H]
\SetKwInOut{Input}{Input}\SetKwInOut{Output}{Output}
\Input{Number of nodes $n$, number of edges
$k$, rewiring probability $p$}
\Output{Watts-Strogatz graph $G(V,E)$}
\BlankLine
$G \gets$ ring lattice with $n$ nodes, each connected to its $k$ nearest neighbors\;

\For{$i \gets 1$ \KwTo $n$}{
  \For{$j \gets i+1$ \KwTo $i+k/2$}{
    $G \gets$ rewire edge $e_{ij}$ with probability $p$\;
  }
}
\caption{Watts-Strogatz}
\label{alg:Watts-Strogatz}
\end{algorithm}
\end{minipage}
\end{figure}

\newpage
\subsubsection{Barabási-Albert Model}
The Barabási-Albert algorithm generates undirected graphs with a preferential attachment mechanism. The probability $p_i$ of adding an undirected edge to a node $v_i$ is given by
\begin{equation} \label{eq:BA_pereferential_prob}
    p_i=\frac{k_i}{\sum_{j=1}^n k_j}\;,
\end{equation}
which leads to hub-like structures with scale-free degree distribution. The algorithm starts with a fully connected graph with $n$ nodes, and then iteratively adds $n-k$ nodes which are attached to the $k$ existing nodes with probability Eqn. \ref{eq:BA_pereferential_prob}. 
\begin{figure}[ht]
\centering
\begin{minipage}{.9\linewidth}
\begin{algorithm}[H]
\SetKwInOut{Input}{Input}\SetKwInOut{Output}{Output}
\Input{Number of nodes $n$, number of edges $k$}
\Output{Barabási-Albert graph $G(V,E)$}
\BlankLine
$G \gets$ graph with $k$ nodes fully connected\;

\For{$i \gets k+1$ \KwTo $n$}{
Add node $v_i$

  \For{$l \gets 1$ \KwTo $k$}{
  Select node $v_j$ with probability $p_j = \frac{k_j}{\sum_h k_h}$\;
  
    $E\gets E\cup \{e_{ij}\}$\;
  }
}
\caption{Barabási-Albert}
\label{alg:Barabási-Albert}
\end{algorithm}
\end{minipage}
\end{figure}

\begin{figure}[ht]
    \begin{center}
    \includegraphics[width=1.0\textwidth]{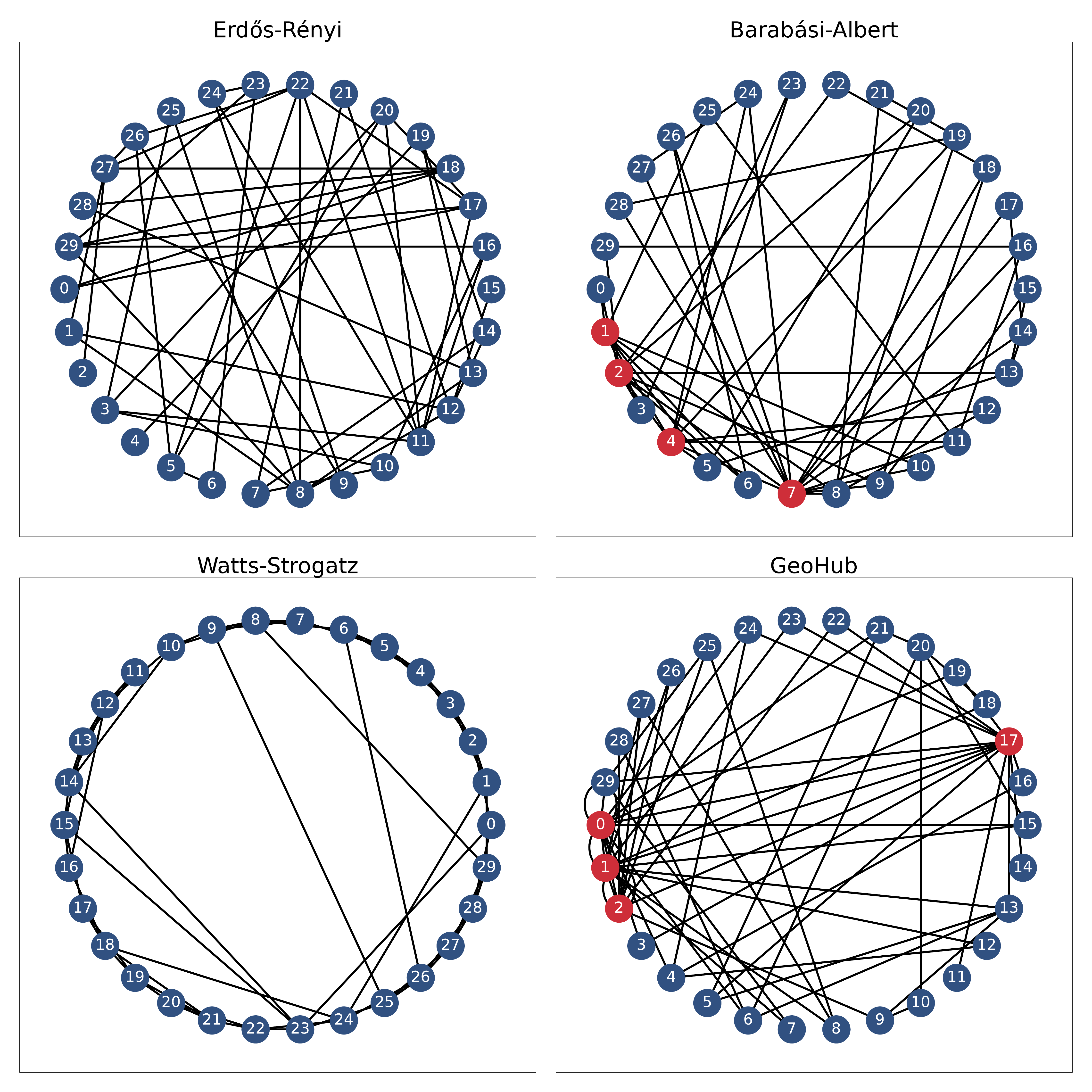}
    \caption{Example graph topologies with network sparsity of $85\%$. Hubs with $\geq 6$ connections are marked in red.}
    \label{fig:graph_examples}
    \end{center}
\end{figure}


\end{document}